\documentclass[11pt]{article}

\usepackage[preprint]{acl}

\usepackage{times}
\usepackage{latexsym}
\usepackage{amsmath}
\usepackage{amssymb}

\usepackage{adjustbox}
\usepackage{booktabs}
\usepackage[T1]{fontenc}

\usepackage[utf8]{inputenc}

\usepackage{graphicx}

\title{What You See Is What You Get: Observation-Aligned Supervision for Chart-to-Code Generation}

\author{
Tianhao Niu
\quad
\textbf{Qingfu Zhu}
\quad
\textbf{Wanxiang Che} \\
 Research Center for Social Computing and Interactive Robotics \\
 Harbin Institute of Technology, China \\
}

\usepackage{hyperref}
\usepackage{cleveref}

\usepackage{ragged2e}

\usepackage{listings}
\lstnewenvironment{promptbox}[1][]{%
  \lstset{basicstyle=\rmfamily\scriptsize,breaklines=true,
    breakatwhitespace=false,columns=fullflexible,keepspaces=true,
    showstringspaces=false,frame=single,#1}%
}{}
\crefname{lstlisting}{Box}{Boxes}
\Crefname{lstlisting}{Box}{Boxes}
\usepackage{array}
\usepackage{tabularx}
\usepackage{makecell}
\usepackage{multirow}
\usepackage{xcolor}

\lstdefinestyle{aclpython}{
  language=Python,
  basicstyle=\ttfamily\scriptsize,
  keywordstyle=\bfseries,
  commentstyle=\itshape,
  numbers=left,
  numberstyle=\tiny,
  stepnumber=1,
  numbersep=4pt,
  frame=single,
  breaklines=true,
  breakatwhitespace=true,
  columns=fullflexible,
  keepspaces=true,
  showstringspaces=false,
  tabsize=2,
  xleftmargin=1.2em,
  framexleftmargin=1.2em,
  captionpos=b
}

\begin{document}
\maketitle
\begin{abstract}
Chart-to-code generation is commonly trained through supervised fine-tuning on reference plotting scripts, implicitly treating the gold code as a fully observable target. However, many chart programs contain latent variables that cannot be uniquely recovered from the rendered image. We identify this latent–observation mismatch in four forms across five chart types: aggregation-induced mismatch, where raw samples are reduced to box statistics or histogram bin masses; normalization-induced mismatch, where absolute scale is removed in pie charts; projection-induced mismatch, where 3D information is lost through 2D rendering; and level-set-induced mismatch, where a scalar field is observable only through selected contour lines. These mismatches introduce target ambiguity and require models to generate information unsupported by the image. We propose Observation-Aligned Supervision, which replaces latent variables with visually constrained quantities. And we study fixed-view 3D scatter and contour charts through controlled experiments. Across VLMs, observation-aligned supervision generally improves observable recovery end-to-end and under both-executable evaluation. Against compact analytic contour formulas, end-to-end performance declines despite better conditional boundary recovery in both-exec setting, suggesting a trade-off with representational compactness.

\end{abstract}

\section{Introduction}
\label{sec:intro}

\begin{figure}[t]
  \includegraphics[width=\columnwidth]{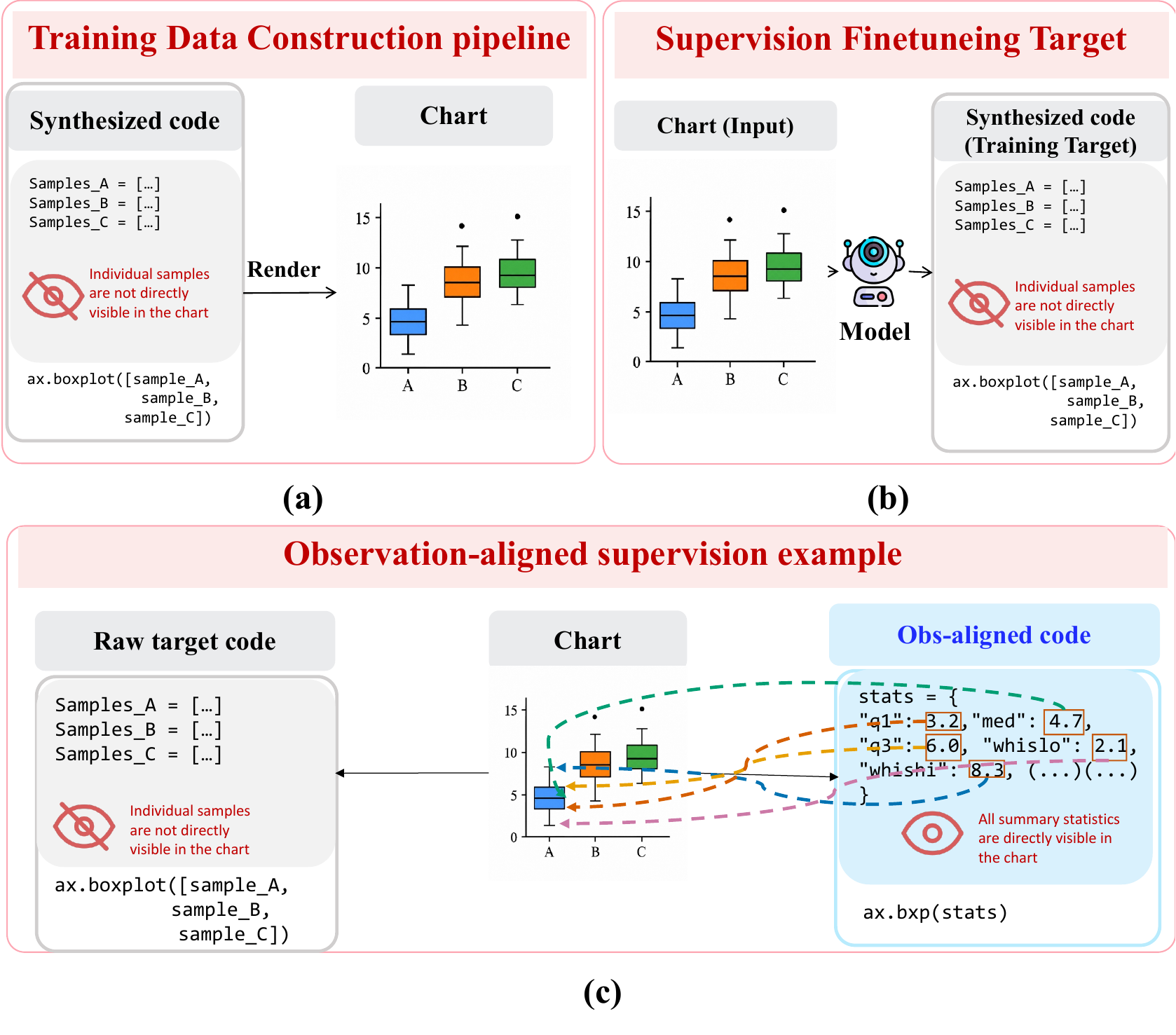}
  \caption{(a) Existing chart-to-code data are constructed by synthesizing code and rendering it into charts. (b) Rendering is a many-to-one process with information loss: e.g., the raw samples underlying a boxplot cannot be recovered from the rendered chart, yet are used as supervision targets, causing a latent-observation mismatch. (c) We reconstruct the target using observable statistics (e.g., quartiles and whiskers), aligning supervision with the visual input.}
  \label{fig:intro}
\end{figure}

Chart-to-code generation~\citep{wu-etal-2025-plot2code,yang2025chartmimic} recovers executable plotting programs from chart images, providing structured, reproducible representations of visualizations. Unlike textual descriptions, code preserves fine-grained layout, style, and data-related information, making it an increasingly important target for data visualization~\citep{shen2026recodereasoningcodegeneration,zhao2025chartcoderadvancingmultimodallarge}. Following existing benchmarks, we focus on \emph{visual production}: faithfully reproducing the input chart's visible content.\footnote{Our study focuses on visual production. Editing generated code to modify visualizations is another practical application of chart-to-code generation. Observation-aligned representations may change the raw-data editing operations; we discuss these in Section~\ref{sec:discussion}.}

Recent work has made substantial progress by scaling chart-code datasets~\citep{zhao2025chartcoderadvancingmultimodallarge,niu-etal-2025-chart2code53,tan2025chartmasteradvancingcharttocodegeneration} and designing stronger post-training objectives~\citep{chen2026breaking,tan2025chartmasteradvancingcharttocodegeneration,tang2026mmrecoderadvancingcharttocodegeneration,he2026chartspecificationstructuralrepresentations}. However, most supervised pipelines render synthesized or collected programs into images and retain those programs as training targets (Figure~\ref{fig:intro}). This raises a question: \textbf{is a visually valid reference program necessarily a suitable supervision target for learning visual production?}

The answer is not always true. Rendering is many-to-one and loses information: different programs and latent data configurations can produce visually indistinguishable charts. Some reference variables are therefore not identifiable from the image. Their values may be valid for authoring the chart, yet supervising their specific values requires predictions unsupported by the visual input.

We identify four forms of non-identifiability across five chart types. \emph{\textbf{Aggregation-induced mismatch}} arises when plotting APIs reduce raw samples to statistics. Boxplots reveal quartiles, medians, and whiskers, while histograms reveal bin-level counts or densities rather than individual samples. Infinitely many raw arrays can thus produce the same boxplot or histogram. \emph{\textbf{Normalization-induced mismatch}} arises in pie charts, whose wedge proportions remain unchanged under positive rescaling of the raw values, leaving absolute magnitudes unidentifiable. \emph{\textbf{Projection-induced mismatch}} arises in fixed-view 3D scatter plots: images constrain projected marker locations, colors, and visible ordering, but not unique 3D coordinates. Even with known camera parameters, points along a viewing line can share a projected location. \emph{\textbf{Level-set-induced mismatch}} arises in contour charts, which constrain a scalar field through its displayed level sets but not its values elsewhere. Infinitely many fields can therefore produce the same visible contour geometry.

We term these problems \emph{\textbf{latent--observation mismatch}}. Unlike annotation noise, this mismatch persists even when reference programs are executable and visually valid. Supervising latent degrees of freedom as unique gold targets makes the inverse problem ill-posed, with two consequences. \emph{\textbf{First}}, target ambiguity can penalize observationally equivalent programs whose latent values differ from the reference. \emph{\textbf{Second}}, \emph{latent completion} requires models to generate unobserved quantities that must nevertheless reproduce the image, increasing learning and inference difficulty.

We propose \emph{observation-aligned supervision}, replacing non-identifiable variables with quantities constrained by rendered observations. Boxplots use explicit statistics, histograms use bin edges and weights, and pie charts use normalized proportions while preserving displayed percentages. Across vision-language models on ChartMimic and ChartX, this generally improves observable-value recovery in end-to-end and paired both-executable evaluations. We further study projection- and level-set-induced mismatch through controlled 3D scatter and contour experiments. Observation-aligned supervision improves fixed-view 3D scatter reconstruction; for contours, it improves boundary and topology recovery over dense numerical-field supervision. Against compact analytic-formula contour targets, however, end-to-end performance declines despite better boundary recovery when both predictions execute.

Our contributions are threefold. \emph{\textbf{First}}, we formulate latent--observation mismatch as a systematic supervision problem and distinguish four sources across five chart types: aggregation, normalization, projection, and level-set reduction. \emph{\textbf{Second}}, we develop executable observation-aligned code rewriting that replaces latent samples, scale, depth, and scalar fields with visually constrained quantities and geometries. \emph{\textbf{Third}}, data audits and experiments across models, benchmarks, leakage-controlled 3D scatter, and controlled contour settings demonstrate improved observable recovery in most tested settings while revealing the limitation against compact analytic-formula targets.

\section{Related Work}
\paragraph{Chart-to-Code Generation.}\footnote{Full related work is provided in appendix~\ref{sec:full_related_work}}
Existing work on chart-to-code generation broadly studies evaluation, data scaling, and improved training objectives. Benchmarks such as ChartMimic~\citep{yang2025chartmimic}, RealChart2Code~\citep{zhang2026realchart2codeadvancingcharttocodegeneration}, Chart2Code~\citep{tang2026chartscodehierarchicalbenchmark}, Plot2Code~\citep{wu-etal-2025-plot2code}, and Plot-Gen~\citep{zhao2025plotgenbenchevaluatingvlmsgenerating} formulate the task as visually grounded code generation and show that current multimodal models still struggle to reproduce complex charts. Other work scales supervised training data through synthetic or rewritten plotting programs, including ChartCoder~\citep{zhao2025chartcoderadvancingmultimodallarge}, Chart2Code53~\citep{niu-etal-2025-chart2code53}, VisCodex~\citep{jiang2025viscodexunifiedmultimodalcode}, and ChartMaster~\citep{tan2025chartmasteradvancingcharttocodegeneration}. More recent methods improve generation through reinforcement learning, structured intermediate representations, and visual feedback~\citep{chen2026breaking,tang2026mmrecoderadvancingcharttocodegeneration,zheng2026chartidedatacentriccharttocodegeneration,he2026chartspecificationstructuralrepresentations}. However, most existing pipelines use a code-first construction process: plotting code is executed to produce an image, and the same code is retained as the supervision target. This assumes that variables in the reference program are recoverable from the rendered chart. We instead study cases where the original script contains latent quantities that cannot be identified from the image. Our work is complementary to multi-view supervision across plotting languages~\citep{zhang2026alignedmultiviewscriptsuniversal}: we focus on non-uniqueness within a single language and align supervision with visually recoverable quantities.

\paragraph{Chart Parsing.}
Chart parsing aims to extract structured content from chart images~\citep{liu2023deplotoneshotvisuallanguage,xia2025chartxchartvlmversatile,xu2025chartmoemixturediverselyaligned,li2026visualselfrefinepixelguidedparadigm} whose output is typically a structured representation of a chart, rather than an executable program that reproduces the figure. Our problem is different: chart-to-code models are supervised with plotting scripts, and these scripts may contain raw data or latent variables that cannot be uniquely recovered from the image. We identify and correct a mismatch between the visual input and the code supervision target.

\begin{figure*}[t]
  \includegraphics[width=\textwidth]{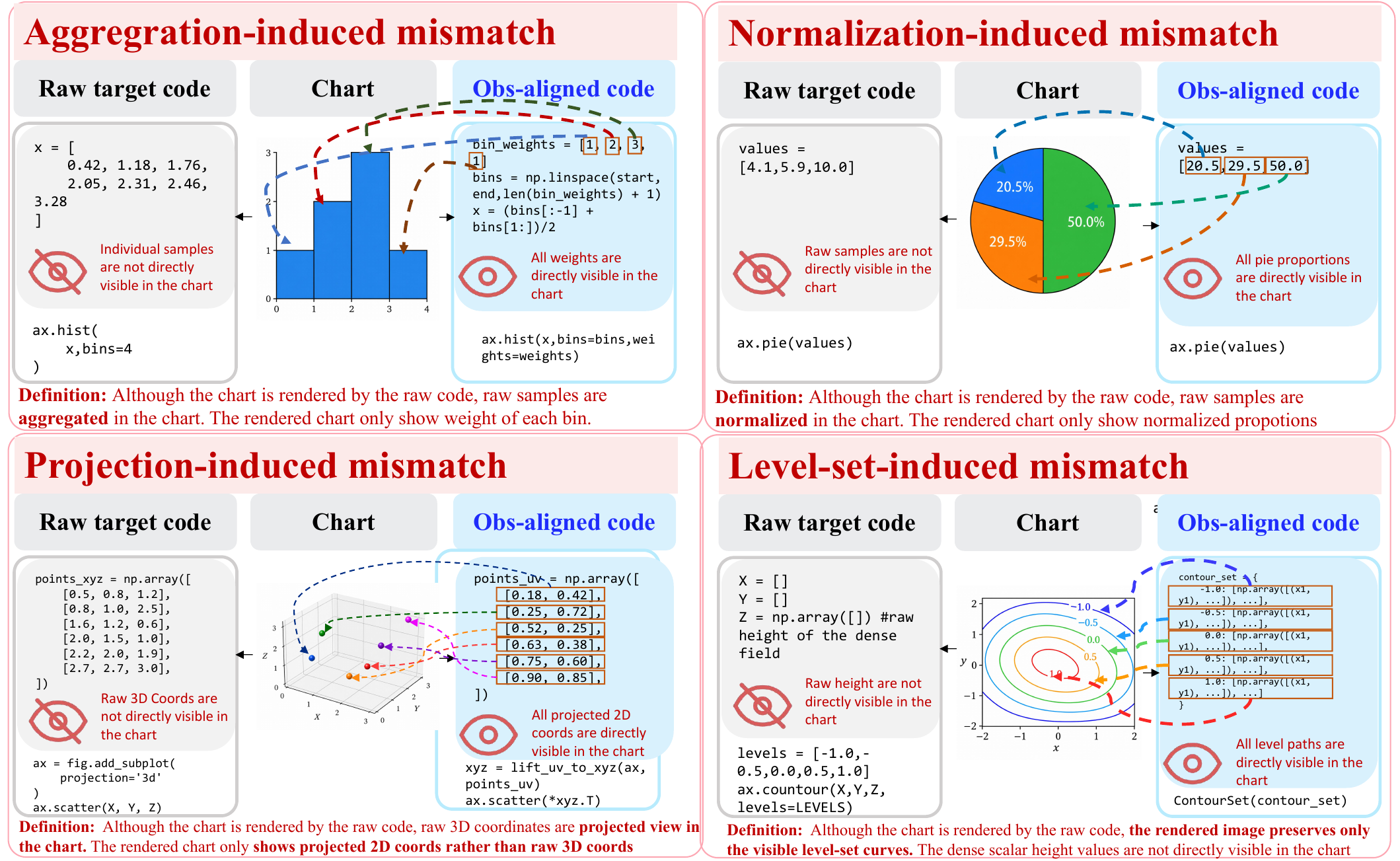}
    \caption{Taxonomy of latent–observation mismatches and their corresponding observation-aligned targets.}
  
\end{figure*}

\section{Method}
\subsection{Formulation}
\label{sec:formalization}

\subsubsection{Definitions}
We provide some formal definitions of observable variables, latent variables, latent-observation mismatch, and its taxonomy in appendix~\ref{sec:more_definitions_and_tax}.

\subsubsection{Observation-Aligned Supervision}

Let \(I=R(c)\) be the
rendered image of a program \(c=(o,z)\), where \(o\) denotes visually
observable chart quantities and \(z\) denotes latent variables used by the
original script. 

We therefore rewrite the original target into an observation-aligned
program \(\bar{c}=T(c)\) that satisfies two properties:
\[
d(R(\bar{c}),R(c))\leq \epsilon, \qquad \bar{c}=g(\hat{o}).
\]
The first property preserves the rendered chart up to small rendering
differences, while the second ensures that the supervised target depends
on visually observable evidence rather than arbitrary latent variables. This reformulation changes the learning problem from recovering
``what the script happened to use'' to generating code for ``what the
chart actually shows.''\footnote{The same observability principle also has implications for evaluating visual-production code. We discuss this more in appendix~\ref{sec:app_receval}.}

This reformulation can help for two reasons.
First, it \textbf{reduces target ambiguity}: different latent
configurations can produce the same chart, yet supervising one
arbitrary configuration treats it as the unique gold target.
Second, it \textbf{avoids latent completion}: with raw targets,
the model must recover the visible chart quantities and construct
unobserved data that reproduce them through the plotting process.
These data are not uniquely determined by the image, but must
nevertheless yield the observed statistics, proportions,
projections, or level sets. Observation-aligned targets directly
encode visually constrained quantities, avoiding this additional
construction requirement and potentially simplifying generation.
Table~\ref{tab:why_obs_alignment_works} details these mechanisms for each chart type.

\paragraph{Prevalence of the Latent--Observation Mismatch}
Auditing ChartCoder and ReChartPrompt, we sample 200 pie charts and 200 histograms. We classify pies without direct percentage inputs and histograms without explicit bin-and-weight representations as mismatched: 183 pies (91.5\%) and 174 histograms (87.0\%). Nearly all boxplots use raw samples; all fixed-view 3D scatter and contour targets use raw 3D data and scalar fields, respectively, indicating widespread mismatch.

\subsection{Aggregation and Normalization: Rewriting and Auditing}
\subsubsection{Data Rewriting}
We rewrite targets with DeepSeek-V4~\citep{deepseekai2026deepseekv4highlyefficientmilliontoken}; details, examples, and prompts appear in appendix~\ref{sec:aggr_normalization_rewriting_auditing_detail}, Figure~\ref{fig:rewriting_pipe_box_pie_hist}, and appendix~\ref{sec:prompts}, respectively. We discard non-executable outputs (less than 1\%) and manually correct no programs.

\paragraph{Box plots.}
Executing each \texttt{boxplot} call yields medians, quartiles, whiskers, and outliers. We replace raw arrays with statistic dictionaries and render them through \texttt{bxp}.

\paragraph{Histograms.}
We recover effective bin edges and counts or masses, replacing raw samples with a compact weighted representation passed to \texttt{hist} with explicit \texttt{bins} and \texttt{weights}.

\paragraph{Pie charts.}
We normalize \texttt{pie} inputs into proportions or percentages while preserving percentage formatting, labels, and visual settings. Each nested or donut ring is normalized independently.

\subsubsection{Training Data Auditing}
\label{sec:training_dataset_auditng_aggr_normal}
We retain samples whose main plotting API is \texttt{boxplot}, \texttt{hist}, or \texttt{pie} from ChartCoder's Chart2Code data~\citep{zhao2025chartcoderadvancingmultimodallarge} and ReChartPrompt-240K~\citep{tan2025chartmasteradvancingcharttocodegeneration}; sizes appear in appendix Table~\ref{tab:training-data-audit}. There are total 19,451 pairs.

Non-identical renderings generally have high SSIM. Inspecting 500 non-identical samples per type attributes most changes to normalization, rounding, and explicit observation-aligned conversion rather than substantial content changes. For boxplots and histograms, training on only 40\% of rewritten data already surpasses raw-code baselines on data recovery, suggesting that rewriting-induced non-equivalence is less likely to primarily explain the gains. Full audits and categories appear in Appendix~\ref{sec:training_dataset_auditng}.

\subsection{Controlled Synthesis and Auditing for 3D Scatter Charts}
We sample visible 2D marker positions and add random displacements along viewing lines, changing raw XYZ coordinates without changing projected positions. Paired targets predict either raw XYZ or normalized marker-center pixel coordinates with canonical zero depth; a deterministic compiler maps the latter to valid XYZ. This removes arbitrary depth targets and latent completion. Details appear in Appendix~\ref{app:3d_scatter_pipeline}.\label{sec:3d_scatter_data_auditing}
We generate $15{,}000$ training pairs, $5{,}000$ per depth scale in $\{0.10, 0.25, 0.45\}$ (appendix~\ref{app:3d_scatter_pipeline}). We independently execute both targets and compare rasterized outputs pixel by pixel, confirming visual equivalence. The example for raw and obs-aligned target are shown in Figure~\ref{fig:3d-raw-code} and Figure~\ref{fig:3d-obs-code}.

\subsection{Controlled Synthesis and Auditing for Contour Charts}
\label{sec:csaafcc}
Each image has raw and observation-aligned targets. In the \emph{dense-field setting}, raw code serializes randomly generated grid values for contouring; aligned code directly renders visible level-set polylines. 
The example for raw and obs-aligned target are shown in Figure~\ref{fig:contour-dense-raw-code} and Figure~\ref{fig:contour-dense-obs-code}.
In the \emph{analytic-formula setting}, raw code uses a compact expression and parameters, while aligned code again uses polylines.
The example for raw and obs-aligned target are shown in Figure~\ref{fig:contour-formula-raw-code} and Figure~\ref{fig:contour-formula-obs-code}.
\emph{Note that a compact analytic representation cannot be assumed for an arbitrary scalar field.} We therefore evaluate both the analytic-formula setting, where such a representation is available, and the dense-field setting, which specifies grid values directly without requiring an analytic formula. The charts of the two settings are different.  Details appear in appendix~\ref{sec:app_countour_data_synthesize}.
Each setting contains $10{,}000$ training and $200$ test pairs, with both targets derived from the same chart; details appear in the appendix~\ref{sec:app_countour_data_synthesize}.

\begin{table*}[t]
\centering
\scriptsize
\setlength{\tabcolsep}{3.8pt}
\renewcommand{\arraystretch}{1.05}
\begin{adjustbox}{max width=\textwidth,center}
\begin{tabular}{ll*{12}{c}}
\toprule
& & \multicolumn{3}{c}{Box}
& \multicolumn{3}{c}{Pie}
& \multicolumn{3}{c}{Hist}
& \multicolumn{3}{c}{Hist-ori} \\
\cmidrule(lr){3-5}
\cmidrule(lr){6-8}
\cmidrule(lr){9-11}
\cmidrule(lr){12-14}
Model & Supervision
& Ex. & Value & TC Avg.
& Ex. & F1 & TC Avg.
& Ex. & Value & TC Avg.
& Ex. & Value & TC Avg. \\
\midrule

\multicolumn{14}{c}{\textit{ChartMimic-ori}} \\
\midrule
InternVL3-8B & No-modified
& 80 & 54.8 & 64.2
& 90 & 53.9 & 70.8
& \textbf{85} & \textbf{40.1} & \textbf{78.6}
& -- & -- & -- \\
InternVL3-8B & Modified
& \textbf{88} & \textbf{57.5}$^{*}$ & \textbf{70.2}
& \textbf{100} & \textbf{84.5}$^{*}$ & \textbf{90.8}
& 75 & 38.4 & 70.9
& -- & -- & -- \\

InternVL3-14B & No-modified
& \textbf{92} & 53.4 & \textbf{75.4}
& 90 & 60.3 & 72.8
& \textbf{80} & 35.5 & \textbf{74.7}
& -- & -- & -- \\
InternVL3-14B & Modified
& 88 & \textbf{66.4}$^{*}$ & 69.8
& \textbf{100} & \textbf{79.1}$^{*}$ & \textbf{87.8}
& 75 & \textbf{51.6}$^{*}$ & 71.4
& -- & -- & -- \\

Qwen2.5-VL-3B & No-modified
& 68 & 45.2 & 50.6
& 95 & 53.2 & 72.7
& 85 & 33.8 & \textbf{78.3}
& -- & -- & -- \\
Qwen2.5-VL-3B & Modified
& \textbf{88} & \textbf{63.2}$^{*}$ & \textbf{69.8}
& 95 & \textbf{84.8}$^{*}$ & \textbf{82.9}
& 85 & \textbf{40.4}$^{*}$ & 76.3
& -- & -- & -- \\

Qwen2.5-VL-7B & No-modified
& 88 & 61.6 & 69.8
& \textbf{100} & 65.5 & \textbf{80.9}
& 75 & 34.0 & 69.1
& -- & -- & -- \\
Qwen2.5-VL-7B & Modified
& \textbf{92} & \textbf{70.6}$^{*}$ & \textbf{74.5}
& 90 & \textbf{76.7}$^{*}$ & 79.7
& \textbf{95} & \textbf{50.1}$^{*}$ & \textbf{87.6}
& -- & -- & -- \\

\midrule
\multicolumn{14}{c}{\textit{ChartMimic-Both executable}} \\
\midrule
InternVL3-8B & No-modified
& 100 & \textbf{67.8} & \textbf{79.6}
& 100 & 59.9 & 78.6
& 100 & 49.7 & 94.0
& -- & -- & -- \\
InternVL3-8B & Modified
& 100 & 65.2 & 78.7
& 100 & \textbf{85.6} & \textbf{91.7}
& 100 & \textbf{50.0} & \textbf{96.0}
& -- & -- & -- \\

InternVL3-14B & No-modified
& 100 & 58.9 & \textbf{82.1}
& 100 & 67.0 & 80.9
& 100 & 50.9 & 92.7
& -- & -- & -- \\
InternVL3-14B & Modified
& 100 & \textbf{76.4} & 79.7
& 100 & \textbf{81.0} & \textbf{88.3}
& 100 & \textbf{70.4} & \textbf{95.0}
& -- & -- & -- \\

Qwen2.5-VL-3B & No-modified
& 100 & 65.1 & 76.4
& 100 & 53.6 & 76.8
& 100 & 40.4 & \textbf{91.9}
& -- & -- & -- \\
Qwen2.5-VL-3B & Modified
& 100 & \textbf{77.9} & \textbf{80.0}
& 100 & \textbf{88.6} & \textbf{87.0}
& 100 & \textbf{49.4} & 91.6
& -- & -- & -- \\

Qwen2.5-VL-7B & No-modified
& 100 & 69.0 & 79.1
& 100 & 68.6 & 81.9
& 100 & 48.6 & \textbf{91.8}
& -- & -- & -- \\
Qwen2.5-VL-7B & Modified
& 100 & \textbf{74.9} & \textbf{81.6}
& 100 & \textbf{85.2} & \textbf{88.6}
& 100 & \textbf{54.1} & 91.5
& -- & -- & -- \\

\midrule
\multicolumn{14}{c}{\textit{ChartX-ori}} \\
\midrule
InternVL3-8B & No-modified
& \textbf{96} & \textbf{93.3} & 82.8
& 98 & 88.5 & 92.7
& 90 & 43.5 & 82.7
& 50 & 16.0 & 33.4 \\
InternVL3-8B & Modified
& 92 & 91.0 & \textbf{83.1}
& \textbf{99} & \textbf{93.9}$^{*}$ & \textbf{95.8}
& \textbf{94} & \textbf{80.8}$^{*}$ & \textbf{89.4}
& \textbf{60} & \textbf{31.7}$^{*}$ & \textbf{40.8} \\

InternVL3-14B & No-modified
& \textbf{94} & 88.9 & 79.1
& 99 & 92.6 & 94.5
& 98 & 51.8 & 94.1
& 68 & 14.8 & 44.3 \\
InternVL3-14B & Modified
& 90 & \textbf{89.0} & \textbf{81.7}
& \textbf{100} & \textbf{96.0}$^{*}$ & \textbf{96.1}
& \textbf{100} & \textbf{88.4}$^{*}$ & \textbf{95.5}
& \textbf{70} & \textbf{26.9}$^{*}$ & \textbf{46.6} \\

Qwen2.5-VL-3B & No-modified
& 88 & 84.8 & 74.4
& \textbf{99} & 84.1 & 92.9
& \textbf{92} & 52.0 & \textbf{88.0}
& \textbf{70} & 17.7 & \textbf{46.5} \\
Qwen2.5-VL-3B & Modified
& \textbf{94} & \textbf{93.8}$^{*}$ & \textbf{85.7}
& 97 & \textbf{92.6}$^{*}$ & \textbf{93.9}
& 90 & \textbf{64.8}$^{*}$ & 86.0
& 56 & \textbf{21.2}$^{*}$ & 37.0 \\

Qwen2.5-VL-7B & No-modified
& \textbf{88} & 85.0 & 77.9
& 99 & 88.7 & 94.1
& 96 & 55.5 & 92.3
& 72 & 13.9 & 44.0 \\
Qwen2.5-VL-7B & Modified
& 86 & \textbf{85.9}$^{*}$ & \textbf{81.0}
& 99 & \textbf{94.6}$^{*}$ & \textbf{96.0}
& \textbf{100} & \textbf{93.4}$^{*}$ & \textbf{96.8}
& 72 & \textbf{23.1}$^{*}$ & \textbf{50.0} \\

\midrule
\multicolumn{14}{c}{\textit{ChartX-Both executable}} \\
\midrule
InternVL3-8B & No-modified
& 100 & 97.1 & 86.5
& 100 & 91.3 & 95.3
& 100 & 49.4 & 93.3
& 100 & 41.1 & \textbf{78.7} \\
InternVL3-8B & Modified
& 100 & \textbf{98.8} & \textbf{90.1}
& 100 & \textbf{95.8} & \textbf{96.8}
& 100 & \textbf{85.6} & \textbf{95.1}
& 100 & \textbf{65.1} & 78.3 \\

InternVL3-14B & No-modified
& 100 & 94.3 & 83.3
& 100 & 93.5 & 95.5
& 100 & 52.9 & \textbf{96.0}
& 100 & 21.3 & 64.0 \\
InternVL3-14B & Modified
& 100 & \textbf{98.8} & \textbf{90.6}
& 100 & \textbf{97.0} & \textbf{96.8}
& 100 & \textbf{89.0} & 95.4
& 100 & \textbf{37.6} & \textbf{68.8} \\

Qwen2.5-VL-3B & No-modified
& 100 & 96.1 & 85.2
& 100 & 85.5 & 93.9
& 100 & 56.7 & \textbf{95.6}
& 100 & 37.3 & \textbf{80.0} \\
Qwen2.5-VL-3B & Modified
& 100 & \textbf{99.8} & \textbf{90.9}
& 100 & \textbf{95.5} & \textbf{96.8}
& 100 & \textbf{72.8} & 95.4
& 100 & \textbf{50.5} & 73.5 \\

Qwen2.5-VL-7B & No-modified
& 100 & 96.7 & 88.4
& 100 & 89.5 & 95.0
& 100 & 57.8 & 96.1
& 100 & 21.8 & 64.3 \\
Qwen2.5-VL-7B & Modified
& 100 & \textbf{99.9} & \textbf{95.0}
& 100 & \textbf{95.5} & \textbf{96.9}
& 100 & \textbf{94.4} & \textbf{96.6}
& 100 & \textbf{35.1} & \textbf{68.9} \\

\bottomrule
\end{tabular}
\end{adjustbox}
\caption{Results on ChartMimic and ChartX under all-sample (ori)
and paired both-executable evaluation. Ex. is execution rate;
TC Avg. averages text and color scores; Hist-ori uses the original
ChartX histogram subset. No-modified and Modified denote raw-code
and observation-aligned supervision. Each executable prediction
reproduces the specified chart type. Bold marks the higher paired
score. In the all-sample blocks, $^{*}$ indicates that the paired
95\% bootstrap confidence interval for Modified minus Raw
excludes zero, with chart examples as the resampling units.
Appendix~\ref{app:additional_results} reports layout scores for
selected models and three-seed training results for Qwen2.5-VL-3B.}
\label{tab:main_results_compact}
\end{table*}

\section{Experiment}

\subsection{Experiments on Aggregation- and Normalization-Induced Mismatch}
\subsubsection{Experimental Setup}
\paragraph{Dataset and Models}
We use the dataset defined in Sec~\ref{sec:training_dataset_auditng_aggr_normal}
for training, total 19451 samples for Non-Modified setting (trained with raw supervision) and Modified setting (trained with observation-aligned supervision) respectively. We use these data to fine-tune Qwen2.5-VL~\citep{bai2025qwen25vltechnicalreport} and InternVL3~\citep{zhu2025internvl3exploringadvancedtraining}. 

\paragraph{Evaluation Benchmarks} We use ChartX~\citep{xia2025chartxchartvlmversatile} pie, hist and boxplot subsets. The bin boundaries of the original histogram in ChartX are discontinuous. We construct a canonicalized ChartX-Hist  by converting histogram-like bar charts with categorical or discontinuous bin labels into continuous-bin histogram representations. The numerical bin heights are unchanged. We use the modified Hist subset in our main experiment. The modification details , example and rationales are shown in the appendix~\ref{sec:ChartX_hist_modification_details}. We also report the performance of the Hist values on the original ChartX, where the results also show improvement.
And we also evaluate our model on ChartMimic~\citep{yang2025chartmimic} Hist, Boxplot and pie subsets.

\paragraph{Evaluation Metrics for Data-Value Recovery} 

\textbf{(1) Hist-Value:} For histograms, we compare
bin heights extracted from the executed reference and predicted
code.  A predicted bin is matched to at most one reference bin, and the
matching cost is the absolute height difference normalized by the value-axis
scale of the reference subplot.  Missing or extra bins receive a unit
penalty.  \textbf{(2) BoxPlot-Value:} For boxplots, we represent each box by the five statistics that
define its shape, namely the lower whisker, first quartile, median, third
quartile, and upper whisker.  We then match reference and predicted boxes
using the normalized $\ell_1$ distance between their five-number vectors,
again penalizing unmatched boxes.  The normalization axis follows the plot
orientation: vertical boxplots use the reference $y$-axis scale, while
horizontal boxplots use the reference $x$-axis scale.  \textbf{(3) Pie-F1:} For pie charts, we
extract wedge percentages and compute an F1 score under a percentage
tolerance of 0.1.  
These metrics directly evaluate whether the predicted code
preserves the observable numerical structure behind the chart. 
More details are in appendix~\ref{sec:aggr_normalization_eval_metric_detail}.
\textbf{Note that our data-value metrics are execution-based, not code-similarity-based, two programs receive the same score whenever they produce the same artist-level quantities, regardless of whether they use raw data or an observation-aligned representation. }. We also conduct $\tau$-sensitive analysis of the pie-F1 metric and the results are shown in Appendix~\ref{sec:app_tausensitiveanalysis}.

\paragraph{Evaluation Metrics for Chart Style}
We also report ChartMimic low-level color and text metrics to show the other Chart2Code evaluation dimensions.\footnote{Note that the gold colors and pred colors of the boxplot are directly extracted from the box artist instead of the .boxplot API.} We report the text and color average score in Table~\ref{tab:main_results_compact}
and show the full evaluation results in appendix Table~\ref{tab:full_results_grouped_with_hist_ori}.
\paragraph{Evaluation Settings}
We report results under two evaluation settings. In the \textit{Ori} setting, all metrics are assigned zero scores when the generated code fails to execute. In the \textit{Both-executable} setting, we restrict evaluation to paired samples where both the Non-modified and Modified outputs from the same model are executable. This paired setting removes the confounding effect of execution failures, enabling a more focused comparison of data-value fidelity and style reconstruction. \footnote{The both-executable setting is intended only as a diagnostic analysis. It is not intended to replace the standard all-sample evaluation. We report the proportion of samples included in the both-executable subset for each model, chart type, and dataset in appendix~\ref{sec:both_exec_eval_sample_coverage}. The coverage ranges from 60\% to 99\%, therefore the diagnostic analysis still covers most evaluation samples. }

\subsubsection{Main Results}
\paragraph{Observation-Aligned Supervision Generally Improves Data-Recovery Performance}
Table~\ref{tab:main_results_compact} shows that observation-aligned supervision improves data recovery on both ChartMimic and ChartX. In the original setting, the gains are sometimes confounded by changes in execution rate. This is important for boxplots, whose gains look small in some original results as several modified outputs fail to execute. To reduce this effect, we also report the both-executable setting, where both compared programs run successfully. In this setting, the improvement becomes clearer and more stable.  These results show that making latent observations explicit helps. The gains are not tied to one model family. Full results are show in appendix~\ref{sec:full_main_exp_res}.

\begin{table}[t]
    \centering
    \small
    \setlength{\tabcolsep}{3.5pt}
    \resizebox{\columnwidth}{!}{%
        \begin{tabular}{llccc}
            \toprule
            Supervision & Subset
            & Hist-Value
            & Box-Value
            & Pie-F1 \\
            \midrule
            Raw & All
            & 73.35 & 78.72 & 82.65 \\
            Obs.-Aligned & All
            & \textbf{96.41} & \textbf{99.82} & \textbf{92.12} \\
            \midrule
            Raw & Both-exec.
            & 79.18 & 96.18 & 82.65 \\
            Obs.-Aligned & Both-exec.
            & \textbf{99.34} & \textbf{99.83} & \textbf{92.12} \\
            \bottomrule
        \end{tabular}%
    }
    \caption{
        Observable value recovery results (\%).
        ``Both-exec.'' includes only examples for which both the Raw and
        Observation-Aligned predictions execute successfully.
    }
    \label{tab:observable_value_results}
\end{table}

\subsubsection{Controlled Synthetic Experiment}
\label{sec:cse}
\textbf{To further rule out effects from code canonicalization introduced by observation-aligned supervision}, we added a controlled synthetic experiment. For each chart type, we  generated 3K training samples for each type using the code template shown in appendix~\ref{sec:controlled_exp_for_aggr_normal}. The Raw and Observation-Aligned settings use the same images, code templates, chart styles, model configuration, and training budget; the only difference is whether the supervision target contains arbitrary raw data or the corresponding observable chart values. The training and test sets are generated independently and there are no pixel-identical images. \textbf{The paired charts are pixel-equivalent.} The results in Table~\ref{tab:observable_value_results} show that code canonicalization alone does not remove the latent–observation mismatch, while observation-aligned supervision still substantially improves numerical recovery.

\begin{table*}[t]
    \centering
    \small
    \setlength{\tabcolsep}{7pt}
    \begin{tabular}{llccccc}
        \toprule
        Model & Supervision
        & Exec. $\uparrow$
        & F1@5px $\uparrow$
        & OSPA@20px $\downarrow$
        & SSIM $\uparrow$
        & Markers (M/P/G) \\
        \midrule
        Qwen2.5-VL-7B
        & Raw XYZ
        & 100.0
        & 99.926
        & 0.788
        & 0.9974
        & 3036/3038/3040 \\
        & Observation-aligned
        & 100.0
        & \textbf{99.964}
        & \textbf{0.382}
        & \textbf{0.9991}
        & \textbf{3038}/3039/3040 \\
        \midrule
        InternVL3-14B
        & Raw XYZ
        & 100.0
        & 99.196
        & 0.808
        & 0.9975
        & 3005/3033/3040 \\
        & Observation-aligned
        & 100.0
        & \textbf{99.709}
        & \textbf{0.616}
        & \textbf{0.9981}
        & \textbf{3026}/3036/3040 \\
        \bottomrule
    \end{tabular}
    \caption{
    Results on the controlled fixed-view 3D scatter evaluation set.
    Marker F1 is reported as a percentage, and OSPA is measured in pixels.
    In M/P/G, P and G denote the total numbers of predicted and gold markers,
    respectively, while M denotes the number of one-to-one predicted--gold
    marker pairs whose center distance is within the 5-pixel matching
    threshold.
    Observation-aligned supervision reduces OSPA by $0.407$ pixels for
    Qwen2.5-VL-7B (95\% bootstrap CI: $[0.380,0.435]$; better on
    200/200 examples; paired Wilcoxon $p=1.4\times10^{-34}$) and by
    $0.192$ pixels for InternVL3-14B (95\% CI: $[0.148,0.239]$;
    better on 168/200 examples; $p=8.2\times10^{-23}$).
    The paired gain is defined as
    $\mathrm{OSPA}_{\mathrm{raw}}-\mathrm{OSPA}_{\mathrm{obs}}$.
    }
    \label{tab:3d_scatter_results}
\end{table*}

\subsubsection{Data Composition Analysis and Robustness analysis}
We show the effects of different training-data compositions in appendix~\ref{sec:data_composition_analysis}.  We conduct experiments on different histogram bin counts and on pie charts without percentage annotations. observation-aligned supervision still improves in these settings. More details are in Appendix~\ref{sec:robustness_analysis}.

\subsection{Projection-induced Mismatch Experiments}

\subsubsection{Experimental Setup}

\paragraph{Dataset and models}
We use the controlled 3D scatter dataset described above in Section~\ref{sec:3d_scatter_data_auditing}. For each backbone, we fine-tune one model using raw XYZ targets and another using observation-aligned targets. Both variants use the same input images, training size, and optimization budget. We evaluate Qwen2.5-VL-7B and InternVL3-14B-Instruct. We independently generate $200$ held-out charts using random seeds not used during training. The raw and observation-aligned models are evaluated on exactly the same images.  We check train--test overlap
using rendered-image hashes and visible marker positions.

\paragraph{Evaluation Metrics}
We execute every predicted and gold program before evaluation. For both
supervision settings, the evaluator extracts markers only from valid
Matplotlib 3D scatter collections and projects their 3D centers onto the
canvas using the chart's actual camera and coordinate transforms. All metrics
are therefore computed from the same projected marker sets and rendered
images, rather than from the predicted XYZ or UV values. A Raw XYZ prediction
and an observation-aligned prediction receive the same score if they produce
the same fixed-view observation. 

We report execution rate, Marker F1 at a 5-pixel threshold, OSPA-Center with a
20-pixel cutoff, and Crop-SSIM. Marker F1 evaluates one-to-one marker recovery
within the matching threshold.  More details are in appendix~\ref{sec:3d_metric_details}.

\subsubsection{Main Results}

The results are shown in Table~\ref{tab:3d_scatter_results}. Observation-aligned supervision improves fixed-view
reconstruction across both model families while preserving a 100\%
execution rate.

\subsection{Level-set-induced Mismatch Experiments}
\label{sec:contour_experiments}
\begin{table*}[t]
\centering
\tiny
\setlength{\tabcolsep}{2pt}
\renewcommand{\arraystretch}{1.08}
\resizebox{\textwidth}{!}{%
\begin{tabular}{@{}llrrrrrrrrrrrr@{}}
\toprule
Eval. & Target & Ex.$\uparrow$ & MTO
& mBF1$\uparrow$ & BF1@.001$\uparrow$ & BF1@.0025$\uparrow$
& BF1@.005$\uparrow$ & SSIM$\uparrow$ & CompF1$\uparrow$
& Closure$\uparrow$ & CntErr$\downarrow$ & Chamfer$\downarrow$ & HD95$\downarrow$ \\
\midrule
\multicolumn{14}{@{}l}{\textbf{Dense fields: Qwen2.5-VL-7B ($n=200/122$)}} \\
E2E & Raw & \textbf{0.7500} & 0 & 0.0662 & 0.0207 & 0.0586 & 0.1192 & 0.4427 & 0.0003 & 0.0100 & 0.4577 & \textbf{0.2665} & \textbf{0.2957} \\
E2E & OA & 0.7250 & 0 & \textbf{0.1833}\textsuperscript{*} & \textbf{0.0657}\textsuperscript{*} & \textbf{0.1734}\textsuperscript{*} & \textbf{0.3108}\textsuperscript{*} & \textbf{0.5236}\textsuperscript{*} & \textbf{0.0239}\textsuperscript{*} & \textbf{0.3050}\textsuperscript{*} & \textbf{0.3312}\textsuperscript{*} & 0.2865 & 0.3007 \\
Both & Raw & -- & 0 & 0.0840 & 0.0262 & 0.0742 & 0.1516 & 0.6035 & 0.0003 & 0.0082 & 0.2808 & 0.0228 & 0.0636 \\
Both & OA & -- & 0 & \textbf{0.2572}\textsuperscript{*} & \textbf{0.0928}\textsuperscript{*} & \textbf{0.2441}\textsuperscript{*} & \textbf{0.4347}\textsuperscript{*} & \textbf{0.7270}\textsuperscript{*} & \textbf{0.0360}\textsuperscript{*} & \textbf{0.4344}\textsuperscript{*} & \textbf{0.0813}\textsuperscript{*} & \textbf{0.0170} & \textbf{0.0363}\textsuperscript{*} \\
\midrule
\multicolumn{14}{@{}l}{\textbf{Dense fields: InternVL3-8B ($n=200/130$)}} \\
E2E & Raw & 0.6850 & 0 & 0.0554 & 0.0173 & 0.0490 & 0.1000 & 0.4080 & 0.0000 & 0.0000 & 0.4883 & 0.3406 & 0.3672 \\
E2E & OA & \textbf{0.9000}\textsuperscript{*} & 29 & \textbf{0.5604}\textsuperscript{*} & \textbf{0.3605}\textsuperscript{*} & \textbf{0.6217}\textsuperscript{*} & \textbf{0.6991}\textsuperscript{*} & \textbf{0.6733}\textsuperscript{*} & \textbf{0.5454}\textsuperscript{*} & \textbf{0.7476}\textsuperscript{*} & \textbf{0.2715}\textsuperscript{*} & \textbf{0.2466}\textsuperscript{*} & \textbf{0.2511}\textsuperscript{*} \\
Both & Raw & -- & 0 & 0.0804 & 0.0251 & 0.0711 & 0.1451 & 0.5973 & 0.0000 & 0.0000 & 0.2559 & \textbf{0.0382}\textsuperscript{*} & \textbf{0.0770}\textsuperscript{*} \\
Both & OA & -- & 23 & \textbf{0.6165}\textsuperscript{*} & \textbf{0.4053}\textsuperscript{*} & \textbf{0.6818}\textsuperscript{*} & \textbf{0.7624}\textsuperscript{*} & \textbf{0.7384}\textsuperscript{*} & \textbf{0.6028}\textsuperscript{*} & \textbf{0.8140}\textsuperscript{*} & \textbf{0.2027} & 0.1787 & 0.1839 \\
\midrule
\multicolumn{14}{@{}l}{\textbf{Analytic formulas: Qwen2.5-VL-7B ($n=200/122$)}} \\
E2E & Raw & \textbf{1.0000}\textsuperscript{*} & -- & \textbf{0.07687} & 0.02444 & \textbf{0.06808} & \textbf{0.13807} & \textbf{0.68687}\textsuperscript{*} & 0.00000 & 0.00000 & \textbf{0.11778} & \textbf{0.02941}\textsuperscript{*} & \textbf{0.08187} \\
E2E & OA & 0.6100 & -- & 0.07252 & \textbf{0.02644} & 0.06762 & 0.12350 & 0.45571 & \textbf{0.00711} & \textbf{0.04500} & 0.44937 & 0.40338 & 0.43140 \\
Both & Raw & 1.0000 & -- & 0.06458 & 0.02043 & 0.05735 & 0.11596 & 0.74555 & 0.00000 & 0.00000 & 0.11197 & 0.03345 & 0.09103 \\
Both & OA & 1.0000 & -- & \textbf{0.11888}\textsuperscript{*} & \textbf{0.04334}\textsuperscript{*} & \textbf{0.11085}\textsuperscript{*} & \textbf{0.20245}\textsuperscript{*} & \textbf{0.74707} & \textbf{0.01166}\textsuperscript{*} & \textbf{0.07377}\textsuperscript{*} & \textbf{0.09733} & \textbf{0.02193}\textsuperscript{*} & \textbf{0.06787}\textsuperscript{*} \\
\bottomrule
\end{tabular}%
}
\caption{Contour results. Raw/OA: raw/observation-aligned targets; E2E/Both: end-to-end/both-executable evaluation. Each panel lists E2E/Both sample counts. Ex.: execution rate; MTO: metric timeouts; CntErr: component-count error; SSIM: Crop-SSIM. Timeouts use worst-case scores (similarity lower bounds; error/distance upper bounds). Bold: better paired mean; $^*$: paired-bootstrap 95\% CI excludes zero (10k/20k replicates for dense/formula). Dashes: unreported/inapplicable.}
\label{tab:contour_combined}
\label{tab:contour_dense_e2e}
\label{tab:contour_dense_both}
\label{tab:analytic_formula_combined}
\end{table*}

\paragraph{Experimental setup}
We use the two contour datasets in Section~\ref{sec:csaafcc}, each with
$10{,}000$ paired training examples and $200$ held-out test charts.
Qwen2.5-VL-7B is evaluated in both settings and InternVL3-8B on dense
fields. Within each comparison, raw and observation-aligned models receive
identical images. We check train--test overlap
using rendered-image hashes and contour geometry.

\paragraph{Evaluation}
We independently execute predicted and gold programs, scoring only
fixed-view contours and rendered images. Identical visible contours
receive identical scores regardless of scalar-field, formula, or
polyline representations. Table~\ref{tab:contour_combined}
reports execution, boundary F1, image similarity, topology, and geometric
distances; metric definitions are in
Appendix~\ref{sec:app_countour_exp_details}. End-to-end evaluation includes
all charts and gives failed/missing predictions worst-case scores;
both-executable evaluation retains pairs where both programs execute. InternVL's metric calculation \textbf{\emph{timeouts count as executable but receive worst-case
scores, making similarity/F1 conservative lower bounds and
error/distance upper bounds.}}\footnote{These timeouts occur during downstream metric computation after successful program execution, rather than during model generation or execution; we therefore retain execution credit and assign worst-case metric scores for conservative evaluation.}. We use $10{,}000$/$20{,}000$
paired-bootstrap replicates for dense-field/analytic-formula comparisons;
$^*$ denotes a paired $95\%$ confidence interval excluding zero.

\paragraph{Results}
Observation alignment improves dense-field boundary and topology
recovery across both backbones and evaluation settings, although
InternVL's timeout-penalized conditional distances remain conservative
upper bounds. For analytic formulas, compact targets perform better
end-to-end because of higher execution ($1.00$ vs.\ $0.61$), whereas
observation alignment improves conditional contour recovery
(mBF1: $0.06458\!\rightarrow\!0.11888$). Thus, Analytic Contour results suggest a trade-off with representational compactness, with compact formulas retaining an end-to-end advantage.

\subsection{Case Study}
We show some qualitative cases in Appendix~\ref{sec:case_study}.

\section{Discussion}
\label{sec:discussion}

\paragraph{Scope and practical value of visual production}
Following existing benchmarks, we focus on \emph{visual production}: reproducing observable chart information. Visual production code supports higher-resolution re-rendering, figure resizing, and executable reconstruction when source code is unavailable,

\paragraph{Discussion on raw sample editing}
One practical application of chart-to-code generation is editing the generated code to quickly modify a visualization. Observation-aligned code supports edits to values, labels, colors, and layout, but summary statistics make sample-level editing less direct. One possible solution is a deterministic compiler that synthesizes editable samples matching predicted box statistics or histogram counts for \texttt{boxplot} or \texttt{hist}. The model thus learns only observable quantities, while synthetic samples enable subsequent editing without recovering the original data. Examples appear in appendix~\ref{app:obs-editable-programs}.

\section{Conclusion}
We studied latent–observation mismatch in chart-to-code generation: visually valid reference programs may contain quantities not directly identifiable from the image. We propose observation-aligned supervision and investigate aggregation, normalization, projection, and level-set induced mismatch across five chart types. Experiments show better recovery of observable chart information in most end-to-end and both-executable settings. Analytic Contour results suggest a trade-off with representational compactness, with compact formulas retaining an end-to-end advantage.

\section*{Limitations}
Our current study only covers five chart types: box plots, histograms, pie charts, 3D scatter Charts and Countour Charts. 
However, many other plotting APIs also contain latent computation inside the library. 
For example, violin plots and density plots often depend on kernel density estimation rather than directly visible raw samples; \texttt{hist2d} and \texttt{hexbin} aggregate points into two-dimensional bins.
In these cases, the model may again be asked to recover hidden data or intermediate computations that are not uniquely determined by the final image. 
We have not yet studied how to rewrite such charts into observation-aligned forms. 
Extending our method to these aggregation-based APIs is an important direction for future work.

\bibliographystyle{acl_natbib}
\bibliography{custom}

\appendix

\section{Appendix}
\label{sec:appendix}

\subsection{Full Related Work}
\label{sec:full_related_work}
\paragraph{Chart2Code Generation.}

One line of works focus on evaluation. ChartMimic~\citep{yang2025chartmimic}, RealChart2Code~\citep{zhang2026realchart2codeadvancingcharttocodegeneration}, Chart2Code~\citep{tang2026chartscodehierarchicalbenchmark}, Plot2Code~\citep{wu-etal-2025-plot2code}, Plot-Gen~\citep{zhao2025plotgenbenchevaluatingvlmsgenerating} formulate chart-to-code as a visually grounded code generation problem and introduce scientific chart-code pairs with multi-level evaluation metrics, highlighting that even strong multimodal models struggle to do the task.

A second line of works improve chart-to-code generation by scaling synthetic training data and applying SFT. ChartCoder~\citep{zhao2025chartcoderadvancingmultimodallarge} introduces Chart2Code-160K and Snippet-of-Thought supervision, while Chart2Code53~\citep{niu-etal-2025-chart2code53} and VisCodex~\citep{jiang2025viscodexunifiedmultimodalcode} expand chart-code data to more diverse chart types through mining online plotting code. ChartMaster~\citep{tan2025chartmasteradvancingcharttocodegeneration} further constructs ReChartPrompt-240K from real-world arXiv charts to improve data diversity and visual realism. However, these data synthesize pipelines largely follow a code-first recipe: obtain or synthesize plotting code, execute it to render a chart, and use the same code as the supervision target. They do not explicitly examine whether the target code contains variables that are not identifiable from the rendered image, or whether the original script is the most suitable target for an image-conditioned model. Aligned Multi-View Scripts~\citep{zhang2026alignedmultiviewscriptsuniversal} introducing multi-language supervision target for the same chart and show benefits. Our work is complementary: rather than studying equivalence across plotting languages, we focus on non-uniqueness within a single language, where the reference script may contain latent raw quantities that are not recoverable from the rendered chart.

A third line of works improve chart2code through new algorithm or training objectives.  Breaking the SFT Plateau~\citep{chen2026breaking}, Chartmaster~\citep{tan2025chartmasteradvancingcharttocodegeneration}, MMRecoder~\citep{tang2026mmrecoderadvancingcharttocodegeneration},CharTide~\citep{zheng2026chartidedatacentriccharttocodegeneration} improving chart2code with RL. ChartSpec~\citep{he2026chartspecificationstructuralrepresentations} introduces chart specifications as structure-aware intermediate representations and uses specification to compute rewards to provide denser feedback. These methods address important limitations of direct SFT, but focus primarily on reward design and feedback rather than the identifiability of variables contained in the target code used in the supervision finetuning. Furthermore , these works still treats 3D scatter coordinates as recoverable targets. In contrast, our projection-induced setting explicitly shows that the fixed 3d scatter rendered image only identifies projected marker centers, not source 3D coordinates, and therefore supervision and evaluation should be defined in the projected observation space.

\paragraph{Chart Parsing.}
A closely related line of work studies chart parsing, where the goal is to recover structured representations or answer questions from chart images rather than generate executable plotting programs. DePlot ~\citep{liu2023deplotoneshotvisuallanguage} converts chart images into linearized tables and then relies on language models to perform downstream reasoning over the recovered tabular content. ChartX~\citep{xia2025chartxchartvlmversatile} further expands the evaluation scope of chart-domain multimodal models by covering diverse chart types, tasks, and disciplinary topics, highlighting the difficulty of robust chart reasoning beyond simple value extraction. More recent systems improve chart-specific perception and alignment in different ways: ChartMoE~\citep{xu2025chartmoemixturediverselyaligned} introduces expert-based visual-language alignment for chart understanding using multiple chart-centered supervision formats, while Visual Self-Refine and ChartVSR~\citep{li2026visualselfrefinepixelguidedparadigm} emphasize pixel-level localization and iterative visual feedback to reduce parsing errors such as omission, misalignment, and hallucination. These works share the goal of extracting faithful observable structures from chart images. In contrast, we revisit the supervision targets used in chart-to-code generation and show that reference plotting scripts may contain latent authoring variables that are not identifiable from the rendered image. By rewriting such targets into observation-aligned target, our method brings chart-to-code supervision closer to the observable, while still preserving executable code as the final output.

\subsection{More Formulations}
\subsubsection{Definitions}
\label{sec:more_definitions_and_tax}
\paragraph{Observable variables}
Observable variables are chart quantities that are directly constrained by the
rendered image and can therefore be recovered, at least approximately, from
visual evidence. They may be explicit in the original code, such as axis limits,
or implicit in the rendering, such as histogram bin weights, box statistics,
pie proportions, and projected marker locations.

\paragraph{Latent variables}
Latent variables are chart quantities used by the original program but not uniquely
determined by the rendered chart. Different latent values may produce the same
or visually indistinguishable image, such as raw samples in boxplots and
histograms, the absolute scale of pie-chart values, or the original 3D
coordinates behind a projected scatter plot.

\paragraph{Latent-observation mismatch}
Latent--observation mismatch occurs when such non-identifiable latent variables
are treated as supervision targets.

\paragraph{Taxonomy}
We distinguish four common forms of latent--observation mismatch across five chart types.
\emph{\textbf{Aggregation-induced mismatch}} arises when raw data
are summarized before rendering, as in boxplots and histograms.
\emph{\textbf{Normalization-induced mismatch}} arises when rendering
removes absolute scale, as in pie charts.
\emph{\textbf{Projection-induced mismatch}} arises when
higher-dimensional data are mapped to a lower-dimensional image,
as in 3D scatter charts.
\emph{\textbf{Level-set-induced mismatch}} arises when rendering
retains only selected level sets of an underlying scalar field,
as in contour charts.

\subsection{Chart-Specific Rationale for Observation-Aligned Supervision}
\label{app:why_obs_alignment_works}

Table~\ref{tab:why_obs_alignment_works} summarizes how each rewriting
addresses target ambiguity and latent completion. These rationales
motivate the supervision design rather than guarantee performance gains.

\begin{table*}[t]
\centering
\small
\setlength{\tabcolsep}{3pt}
\renewcommand{\arraystretch}{1.15}
\begin{tabular}{
    @{}
    p{0.09\textwidth}
    p{0.16\textwidth}
    p{0.22\textwidth}
    p{0.24\textwidth}
    p{0.23\textwidth}
    @{}
}
\toprule
\textbf{Chart type} &
\textbf{Non-identifiable raw target} &
\textbf{Observation-aligned target} &
\textbf{Reducing target ambiguity} &
\textbf{Avoiding latent completion} \\
\midrule

Boxplot &
Raw sample arrays. &
Explicit medians, quartiles, whiskers, and displayed outliers,
rendered through \texttt{bxp}. &
Different sample arrays can yield the same displayed statistics.
Supervising these statistics removes arbitrary sample-level
choices. &
Raw targets require constructing unobserved samples whose
computed box statistics match the visible chart.
Aligned targets directly encode those statistics, avoiding
sample construction and its consistency constraints. \\

\addlinespace

Histogram &
Individual samples underlying each bin. &
Effective bin edges and bin-level counts or masses, encoded
through explicit \texttt{bins} and \texttt{weights}. &
Different sample sets can produce identical bin heights.
Explicit bin geometry and values remove within-bin sample
ambiguity. &
Raw targets require constructing unobserved samples whose
binning reproduces the visible counts or densities.
Aligned targets encode bin edges and values through explicit
bins and weights, avoiding sample construction. \\

\addlinespace

Pie chart &
Absolute input values before normalization. &
Normalized wedge proportions or percentages, with displayed
percentage formatting preserved. &
Positive rescaling leaves wedge proportions unchanged.
Normalization removes this arbitrary scale factor. &
Raw targets require choosing absolute values whose normalization
reproduces the visible wedge proportions, although their scale
is not identifiable.
Aligned targets directly encode the proportions, fixing this
otherwise arbitrary scale. \\

\addlinespace

3D scatter &
Raw XYZ coordinates, including unobservable depth along the
viewing direction. &
Normalized 2D marker-center coordinates with canonical depth,
converted to valid XYZ by a deterministic compiler. &
Different points along a viewing line can share a projected
location. Canonical depth removes arbitrary depth assignments
in the controlled fixed-view setting. &
Raw targets require constructing XYZ coordinates whose
projection matches the visible marker positions, although
depth is not uniquely determined.
Aligned targets encode projected positions; a deterministic
compiler supplies canonical XYZ coordinates. \\

\addlinespace

Contour &
Scalar-field values beyond those constrained by displayed
level sets. &
Visible level-set polylines rendered directly. &
Different scalar fields can share the same displayed contours.
Supervising level-set geometry avoids treating one such field
as the unique target. &
Raw targets require constructing a scalar field whose selected
level sets reproduce the visible curves, although values away
from those curves are not uniquely determined.
Aligned targets directly encode the level-set geometry. \\

\bottomrule
\end{tabular}
\caption{
Chart-specific mechanisms by which observation-aligned
supervision reduces target ambiguity and avoids latent completion.
Latent completion requires constructing unobserved authoring data
that are not uniquely determined by the chart but must nevertheless
reproduce its visible content through the plotting process.
}
\label{tab:why_obs_alignment_works}
\end{table*}

\subsection{More Aggregation-induced and Normalization-induced Mismatch Rewriting and Auditing Details}
\label{sec:aggr_normalization_rewriting_auditing_detail}
\subsubsection{Data Rewriting Details}
\paragraph{Rewriting Examples}
\begin{figure*}[t]
  \includegraphics[width=\textwidth]{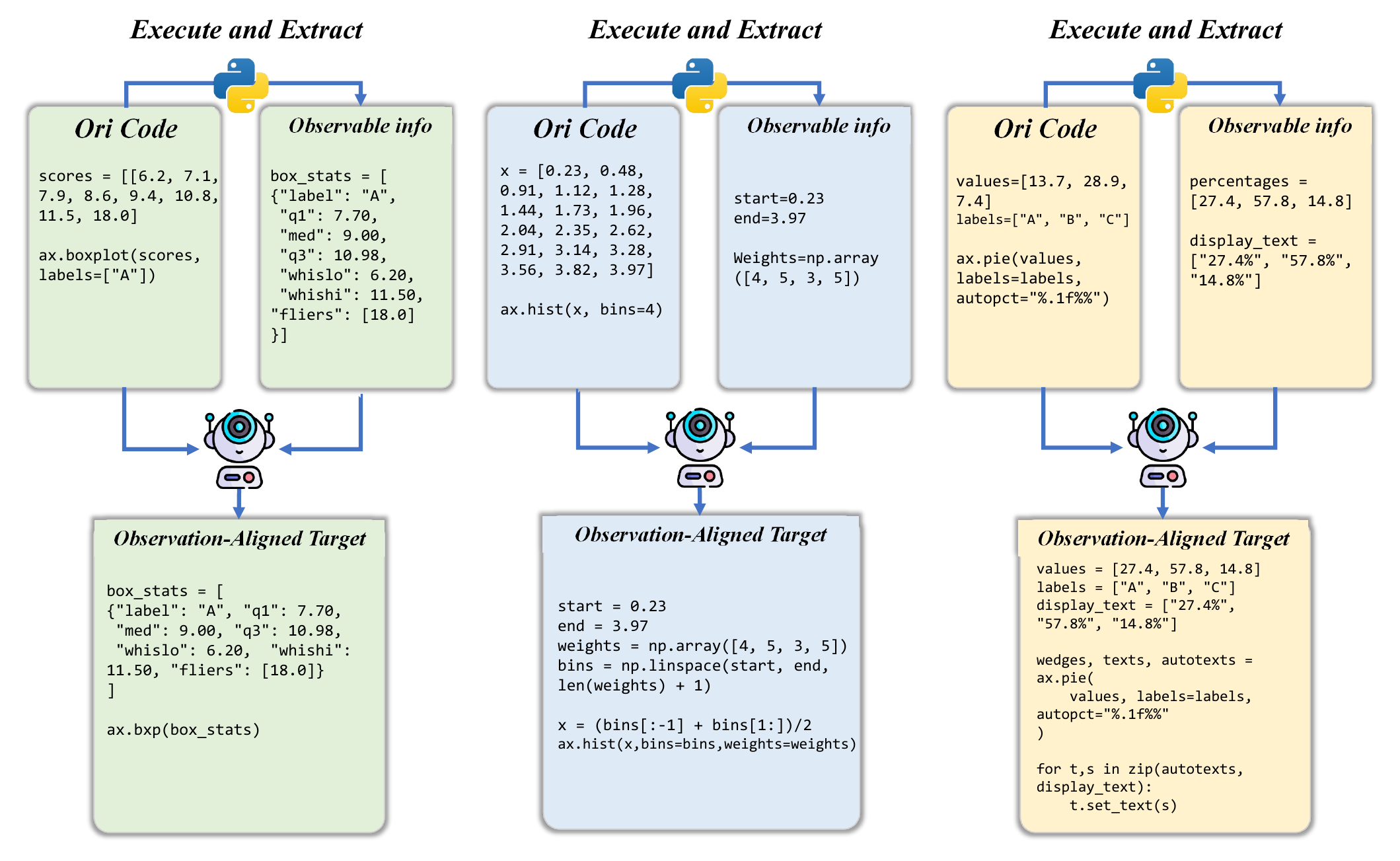}
    \caption{Rewriting example for box, pie and hist charts.}
  \label{fig:rewriting_pipe_box_pie_hist}
\end{figure*}
We show the rewriting example in Figure~\ref{fig:rewriting_pipe_box_pie_hist}.

\paragraph{Rewriting For Box Charts}
For each boxplot call, we execute the original script and compute the statistics required by the rendered boxplot. For each box group, we construct a statistic dictionary containing \texttt{med}, \texttt{q1}, \texttt{q3}, \texttt{whislo}, \texttt{whishi}, and \texttt{fliers}. When the original chart uses notches, confidence intervals, mean markers, or mean lines, we additionally store the corresponding fields such as \texttt{cilo}, \texttt{cihi}, and mean-related values when they are required by the rendering API. The raw sample arrays are then replaced with the list of statistic dictionaries, and the plotting call is rewritten from a sample-level interface such as \texttt{boxplot} to a statistic-level interface such as \texttt{bxp}. We preserve visual arguments including positions, widths, labels, orientation, patch settings, color properties, line styles, whisker settings, cap settings, flier styles, median styles, and axis configuration. Full rewriting prompts are shown in the appendix~\ref{sec:prompts} and the prompt ICL example also shows some rewritten examples.

\paragraph{Rewriting For Hist Charts}
For each histogram, we execute the original plotting call and recover the effective bin edges and bin-level values used for rendering. We then replace the raw sample array with a compact weighted representation. Specifically, for each bin, we assign it a weight equal to the corresponding bin count or bin mass. The rewritten call explicitly passes the recovered \texttt{bins} and the constructed \texttt{weights} to \texttt{hist}. For histograms with \texttt{density=True}, we compute weights according to the normalized density values and bin widths so that the rewritten program follows the same density convention. Visual options such as color, alpha, edge style, line width, orientation, stacking mode, histogram type, labels, legends, log scaling, and axis limits are preserved. Full rewriting prompts are shown in the appendix~\ref{sec:prompts} and the prompt ICL example also shows some rewritten examples.

\paragraph{Rewriting For Pie Charts}
For each pie chart, we first identify the value array passed to \texttt{pie} and execute the original script to obtain the rendered wedge proportions. We then normalize the original values into percentage-style quantities and use the normalized array as the rewritten data input. If the original code contains \texttt{autopct}, we preserve its formatting behavior when possible and rewrite the data so that the displayed percentage text remains consistent with the rendered wedges. Labels, colors, explode offsets, start angle, radius, shadow settings, wedge properties, text properties, legend calls, and axis settings are copied from the original program. When the pie chart is nested or donut-shaped, the normalization is applied independently to each ring. The rewritten program keeps the same pie-chart structure while replacing raw values with normalized wedge-level targets. Full rewriting prompts are shown in the appendix~\ref{sec:prompts} and the prompt ICL example also shows some rewritten examples.

\subsubsection{Data Auditing Details}
\label{sec:training_dataset_auditng}

\begin{figure*}[t]
    \centering
    \includegraphics[width=0.98\textwidth]{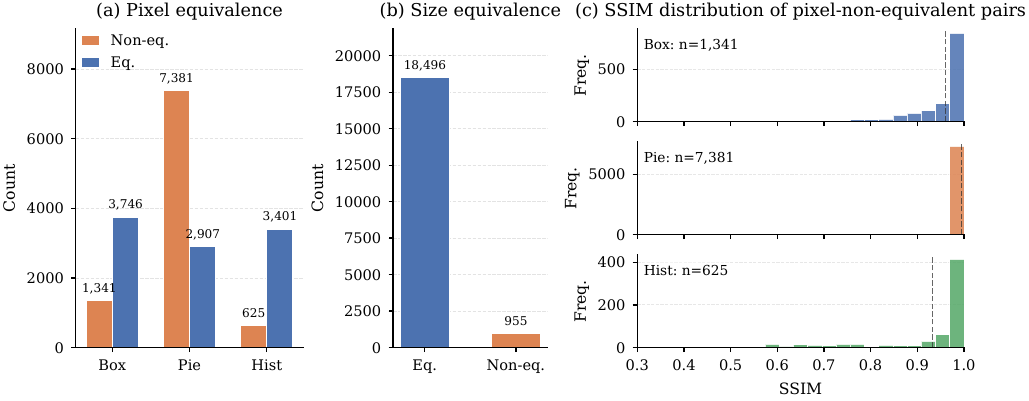}
    \caption{Audit results of visual equivalence before and after rewriting.}
    \label{fig:pixel_size_ssim_audit}
\end{figure*}

To audit the rewritten observation-aligned training data, we compare the original and rewritten samples from both pixel-level and code-semantics perspectives. For pixel-level equivalence, Figure~\ref{fig:pixel_size_ssim_audit} reports the distribution of rendering differences before and after rewriting. Approximately one quarter of the boxplot samples, one seventh of the histogram samples, and most of the pie-chart samples are not pixel-identical after rewriting. However, the SSIM distributions in the right subfigure are concentrated at high values, especially for pie charts, suggesting that most non-identical cases remain visually close to the original renderings.

We further conduct a code-semantics audit on non-pixel-equivalent samples. For each chart type, we randomly sample 500 non-equivalent cases and manually categorize the semantic changes introduced by rewriting. The taxonomy and distribution of these categories are reported. We find that most differences are small semantic shifts caused by normalization, rounding, or explicit conversion from raw values to observation-aligned quantities, rather than substantial changes to the chart content. This supports that the rewriting process mainly introduces controlled rendering-level or numeric perturbations.

In addition, Figure~\ref{fig:latent_obs_ratio} shows that for histograms and boxplots, models trained with only 40\% of the rewritten data already outperform the raw-code baseline on data-recovery metrics. This partially rules out the possibility that the improvements are simply driven by noisier rewritten data, and suggests that the gains for these two chart types are not primarily caused by non-equivalent samples introduced during rewriting.

\begin{table}[t]
\centering
\small
\setlength{\tabcolsep}{3.5pt}
\renewcommand{\arraystretch}{1.08}
\begin{tabular}{llrrr}
\toprule
Dataset & API group & API-count & Main-only & Final \\
\midrule
\multicolumn{5}{l}{\textit{ChartCoder}} \\
Pie & Pie & 10,486 & 9,570 & 9,328 \\
Box & Box & 6,938 & 3,441 & 3,246 \\
Hist-all & Hist & 1,262 & 948 & -- \\
Hist-wo-rand & Hist & -- & 221 & 119 \\
\midrule
\multicolumn{5}{l}{\textit{ReChartPrompt}} \\
Pie & Pie & 1,197 & 1,080 & 960 \\
Box & Box & 3,380 & 2,044 & 1,891 \\
Hist-all & Hist & 11,335 & 7,675 & -- \\
Hist-wo-rand & Hist & -- & 4,246 & 3,907 \\
\bottomrule
\end{tabular}
\caption{
Training-data audit for latent-observation rewriting. \textit{API-count} denotes the number of code files containing the corresponding plotting API before filtering. \textit{Main-only} keeps only examples whose main plotting API belongs to the target group and excludes examples containing other main plotting APIs. \textit{Final} denotes the image-deduplicated training subset used in our main experiments. ``Hist-all'' includes both randomly generated original data and non-random histogram data, while ``Hist-wo-rand'' keeps only histograms whose original data are not expressed by random number generation. We do not use the Hist-all subset for training after deduplication.
}
\label{tab:training-data-audit}
\end{table}

\paragraph{Image-pixel Based Auditing Result} The results are shown in Figure~\ref{fig:pixel_size_ssim_audit}.

\paragraph{Code-Semantic Based Auditing Result} 
\begin{figure*}[t]
    \centering
    \includegraphics[width=0.98\textwidth]{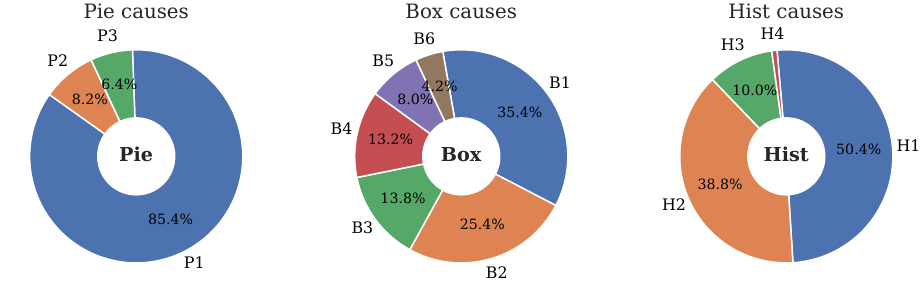}
    \caption{Audit results of code-semantic before and after rewriting.}
    \label{fig:code_semantic_audit}
\end{figure*}
\begin{figure*}[t]
    \centering
    \includegraphics[width=0.98\textwidth]{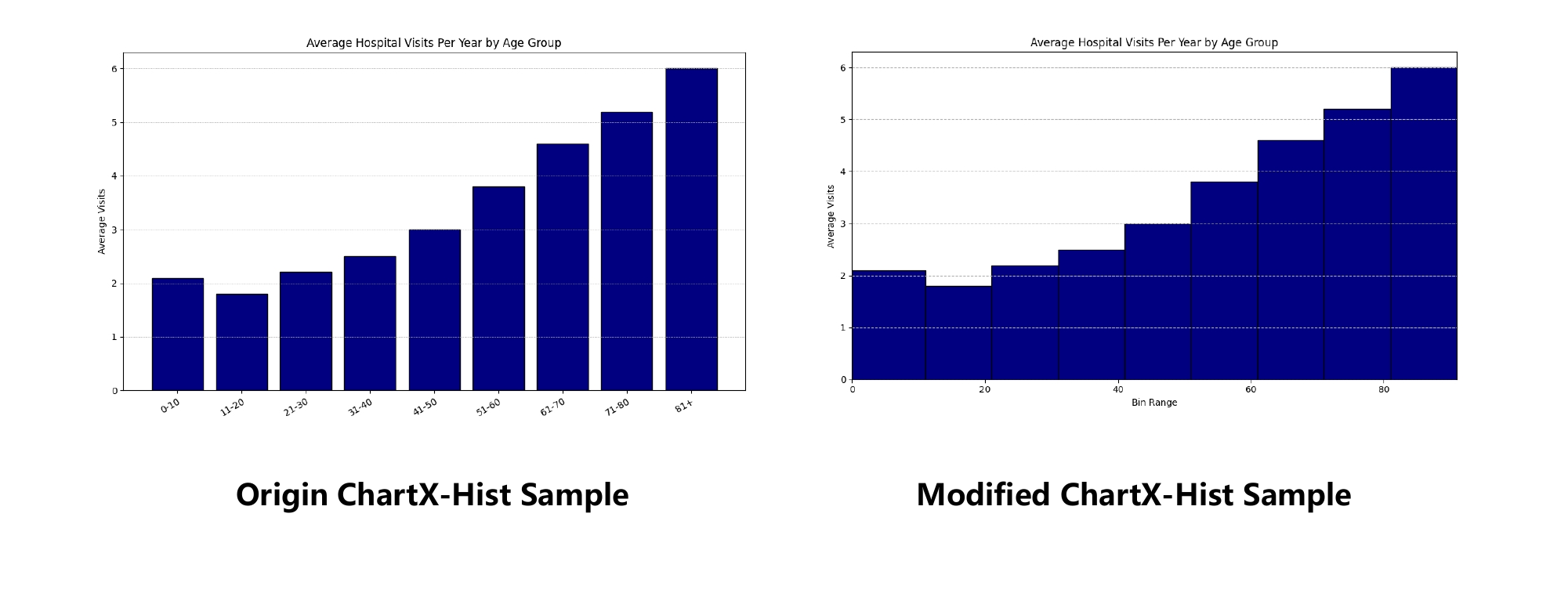}
    \caption{ChartX Hist modification examples}
    \label{fig:chartxhistmodify}
\end{figure*}

We show the Code-semantic Based auditing result in Figure~\ref{fig:code_semantic_audit}. The meaning of each wedge is as follows.
\begin{itemize}
    \item \textbf{Pie.}
    \textbf{P1} denotes cases where the original raw values are replaced with rounded percentages.
    \textbf{P2} denotes cases where the original code does not use \texttt{autopct}, but the wedge sizes are still replaced by rounded percentage-like values.
    \textbf{P3} denotes cases where rounded percentages alter the angular geometry of sectors in donut or nested pie charts.

    \item \textbf{Box.}
    \textbf{B1} denotes cases where the original code contains \texttt{np.random}, making the regenerated before-rewrite figure non-reproducible.
    \textbf{B2} denotes cases where a notched boxplot is rewritten with explicit \texttt{cilo}/\texttt{cihi} values or \texttt{shownotches}, leading to slight geometric drift in the confidence interval notch.
    \textbf{B3} denotes cases where a standard boxplot is rewritten as \texttt{bxp}, and rounding of the five-number statistics or fliers introduces minor visual changes.
    \textbf{B4} denotes cases where style-related keyword arguments also change, such as \texttt{boxprops} \texttt{color} being rewritten as \texttt{edgecolor} or \texttt{linestyle}.
    \textbf{B5} denotes cases where multi-group boxplots written with loops and \texttt{positions=[...]} exhibit differences due to position or style propagation after rewriting.
    \textbf{B6} denotes other causes.

    \item \textbf{Hist.}
    \textbf{H1} denotes cases where \texttt{density=True} is rewritten into explicit weights or bins, introducing differences from density conversion and decimal rounding.
    \textbf{H2} denotes cases where a standard frequency histogram is rewritten as weighted bins, causing subtle drift in bin edges or representative points due to truncation errors.
    \textbf{H3} denotes cases where small count or weight differences are visually amplified under a log-scale histogram.
    \textbf{H4} denotes other causes.
\end{itemize}

\subsection{Controlled Synthesis for 3D scatter Charts and Auditing Details}
\label{app:3d_scatter_pipeline}

\paragraph{Separation of supervision and compilation.}
The complete plotting templates before and after
rewriting are shown in Figures~\ref{fig:3d-raw-code}
and~\ref{fig:3d-obs-code}, respectively.
Each template should be read from the left column
to the right column.

In the raw setting, the model is supervised to
generate three-dimensional XYZ coordinates,
including randomly sampled depth values that
cannot be uniquely inferred from the rendered
image. In the observation-aligned setting, the
model instead predicts normalized two-dimensional
marker centers $(u,v)$ on the full image canvas.
The conversion from UV to XYZ is delegated to
a deterministic lifting compiler, exposed through
the external API \texttt{lift\_uv\_to\_xyz}.
The compiler is a fixed runtime dependency,
rather than a function that the model must learn
to implement or generate.

\paragraph{Role of the lifting compiler.}
For a fixed rendering configuration $\theta$,
including the camera, axis limits, aspect ratio,
and canvas geometry, the lifting compiler maps
each observable marker center $(u,v)$ to one
representative three-dimensional point:
\begin{equation}
\mathbf{q}
=
\operatorname{Lift}_{\theta}(u,v).
\end{equation}
Because the projection from 3D to 2D is
non-injective, this operation does not recover
the original latent depth. Instead, it resolves
the depth ambiguity by selecting a canonical
depth corresponding to the projected depth
of the world-space origin.

Let $\pi_{\theta}$ denote the full-canvas
normalized projection and $d_{\theta}$ the
projected depth coordinate. Under the fixed
orthographic camera, the compiler constructs
$\mathbf{q}$ such that
\begin{equation}
\pi_{\theta}(\mathbf{q}) = (u,v),
\qquad
d_{\theta}(\mathbf{q})
=
d_{\theta}(\mathbf{0}).
\end{equation}
Thus, the reconstructed point lies on the
canonical-depth plane through the world-space
origin, perpendicular to the viewing direction.
The canonical-depth convention does not imply
that the world-space Z coordinate is zero.

The implementation uses the runtime Matplotlib
projection matrix and coordinate transforms.
It is supplied independently of the supervised
plotting scripts, as detailed in
Figure~\ref{fig:3d-lift-api}.
Consequently, the observation-aligned model
only needs to predict visually constrained UV
coordinates and invoke a predefined API.
It does not need to infer arbitrary latent
XYZ coordinates or implement inverse projection.

\begin{figure*}[t]
\centering

\begin{minipage}[t]{0.48\textwidth}
\vspace{0pt}
\scriptsize
\begin{verbatim}
from pathlib import Path

import matplotlib
matplotlib.use("Agg")
import matplotlib.pyplot as plt
import numpy as np


# Raw supervision target.
# The model predicts XYZ, including
# unobservable random depth.
POINTS_XYZ = np.array(
    [
        [-0.729, -0.149,  0.061],
        [-0.132, -0.830, -0.032],
        [-0.237,  0.039,  0.073],
        [-0.403,  0.297,  0.220],
        [-0.243,  0.165, -0.354],
        [ 0.051,  0.002, -0.047],
    ],
    dtype=float,
)


# Create the 3D plotting canvas.
fig = plt.figure(
    figsize=(6, 6),
    dpi=100,
)
ax = fig.add_subplot(
    111,
    projection="3d",
)
\end{verbatim}
\end{minipage}
\hfill
\begin{minipage}[t]{0.48\textwidth}
\vspace{0pt}
\scriptsize
\begin{verbatim}
# Fixed camera and axis settings.
ax.set_proj_type("ortho")
ax.view_init(
    elev=25,
    azim=-55,
)

ax.set_xlim(-1.0, 1.0)
ax.set_ylim(-1.0, 1.0)
ax.set_zlim(-1.0, 1.0)
ax.set_box_aspect(
    (1.0, 1.0, 1.0)
)

ax.set_xlabel("X")
ax.set_ylabel("Y")
ax.set_zlabel("Z")

# Directly render predicted XYZ.
# No lifting compiler is used.
ax.scatter(
    POINTS_XYZ[:, 0],
    POINTS_XYZ[:, 1],
    POINTS_XYZ[:, 2],
    s=36,
    color="tab:blue",
    alpha=1.0,
    depthshade=False,
)

fig.savefig(
    Path(__file__).with_suffix(".png")
)
plt.close(fig)
\end{verbatim}
\end{minipage}

\caption{
Raw XYZ supervision template for the controlled
3D scatter experiment.
The model directly predicts all three world-space
coordinates for each marker, including randomly
assigned latent depth that is not identifiable
from the fixed-view image.
The predicted coordinates are passed directly
to Matplotlib for rendering.
The code continues from the left column
to the right column.
}
\label{fig:3d-raw-code}
\end{figure*}

\begin{figure*}[t]
\centering

\begin{minipage}[t]{0.48\textwidth}
\vspace{0pt}
\scriptsize
\begin{verbatim}
from pathlib import Path

import matplotlib
matplotlib.use("Agg")
import matplotlib.pyplot as plt
import numpy as np

# Fixed external compiler API.
# Supplied by the runtime, not
# implemented by the model.
from uv_lift import lift_uv_to_xyz


# Observation-aligned target.
# Model predicts normalized 2D
# marker centers on the full canvas.
# No latent 3D depth is predicted.
POINTS_UV01 = np.array(
    [
        [0.367, 0.453],
        [0.389, 0.560],
        [0.484, 0.463],
        [0.486, 0.403],
        [0.499, 0.541],
        [0.533, 0.507],
    ],
    dtype=float,
)


# Create the same 3D canvas.
fig = plt.figure(
    figsize=(6, 6),
    dpi=100,
)
ax = fig.add_subplot(
    111,
    projection="3d",
)
\end{verbatim}
\end{minipage}
\hfill
\begin{minipage}[t]{0.48\textwidth}
\vspace{0pt}
\scriptsize
\begin{verbatim}
# Same fixed camera and axes.
ax.set_proj_type("ortho")
ax.view_init(
    elev=25,
    azim=-55,
)

ax.set_xlim(-1.0, 1.0)
ax.set_ylim(-1.0, 1.0)
ax.set_zlim(-1.0, 1.0)
ax.set_box_aspect(
    (1.0, 1.0, 1.0)
)

ax.set_xlabel("X")
ax.set_ylabel("Y")
ax.set_zlabel("Z")

# External compiler/API call:
# UV -> canonical representative XYZ.
# No latent depth inference.
points_xyz = lift_uv_to_xyz(
    ax,
    POINTS_UV01,
)

# Standard Matplotlib rendering.
ax.scatter(
    points_xyz[:, 0],
    points_xyz[:, 1],
    points_xyz[:, 2],
    s=36,
    color="tab:blue",
    alpha=1.0,
    depthshade=False,
)

fig.savefig(
    Path(__file__).with_suffix(".png")
)
plt.close(fig)
\end{verbatim}
\end{minipage}

\caption{
Observation-aligned plotting template for the
controlled 3D scatter experiment.
Unlike raw XYZ supervision, the model predicts
only normalized 2D marker centers.
The \texttt{lift\_uv\_to\_xyz} call invokes an
externally provided deterministic compiler,
which selects representative XYZ coordinates
under the fixed Matplotlib projection.
The compiler implementation is not part of
the model-generated code.
The resulting XYZ coordinates are rendered
using the same camera and plotting settings
as the raw baseline.
The code continues from the left column
to the right column.
}
\label{fig:3d-obs-code}
\end{figure*}

\begin{figure*}[t]
\centering

\begin{minipage}[t]{0.48\textwidth}
\vspace{0pt}
\scriptsize
\begin{verbatim}
# uv_lift.py
# Shared external compiler module.
# NOT a model-generated plotting script.

import numpy as np


def lift_uv_to_xyz(ax, points_uv01):
    """Compile UV into canonical XYZ.

    Inputs:
      ax: configured Matplotlib 3D axes
      points_uv01: N x 2 marker centers

    Output:
      N x 3 representative XYZ points

    This selects canonical depth;
    it does not recover true depth.
    """
    uv = np.asarray(
        points_uv01, dtype=float
    )
    if uv.ndim != 2 or uv.shape[1] != 2:
        raise ValueError("UV must be N x 2")

    # Resolve the runtime canvas.
    fig = ax.figure
    fig.canvas.draw()
    width, height = (
        fig.canvas.get_width_height()
    )

    # Normalized UV (top-left origin)
    # -> display pixels (bottom-left).
    display_xy = np.column_stack(
        [
            uv[:, 0] * width,
            (1.0 - uv[:, 1]) * height,
        ]
    )

    # Display pixels -> projected XY.
    projected_xy = (
        ax.transData.inverted()
        .transform(display_xy)
    )
\end{verbatim}
\end{minipage}
\hfill
\begin{minipage}[t]{0.48\textwidth}
\vspace{0pt}
\scriptsize
\begin{verbatim}
    # Read the actual projection.
    projection = ax.get_proj()

    # Canonical depth is the projected
    # depth of the world-space origin.
    origin_clip = projection @ np.array(
        [0.0, 0.0, 0.0, 1.0]
    )
    canonical_z = (
        origin_clip[2] / origin_clip[3]
    )

    # Complete projected XY with the
    # fixed canonical depth.
    projected_h = np.column_stack(
        [
            projected_xy,
            np.full(
                len(uv),
                canonical_z,
            ),
            np.ones(len(uv)),
        ]
    )

    # Invert the homogeneous projection.
    world_h = (
        projected_h
        @ np.linalg.inv(projection).T
    )

    # Homogeneous -> Cartesian XYZ.
    points_xyz = (
        world_h[:, :3]
        / world_h[:, 3, None]
    )
    return points_xyz
\end{verbatim}
\end{minipage}

\caption{
Implementation of the shared external
\texttt{lift\_uv\_to\_xyz} compiler API.
Given normalized 2D marker centers and an
already configured Matplotlib 3D axis,
the compiler converts full-canvas coordinates
into projected coordinates, assigns the projected
depth of the world-space origin, and applies
the inverse projection to obtain a canonical
3D representative for each marker.
Under the fixed orthographic view, the
representatives preserve the input 2D marker
positions without recovering the original
latent depths.
The implementation continues from the
left column to the right column.
}
\label{fig:3d-lift-api}
\end{figure*}

We render each chart on a $600\times600$ canvas using a fixed orthographic camera with elevation $25^\circ$ and azimuth $-55^\circ$. Axis limits and aspect ratios are fixed across all examples. We use opaque markers, disable depth shading, and reject layouts with overlapping markers.

Let $\mathbf{n}$ denote the viewing direction. Each sampled point is constructed as
\begin{equation}
    \mathbf{p}_i=\mathbf{q}_i+d_i\mathbf{n},
\end{equation}
where $\mathbf{q}_i$ is a canonical point and $d_i$ is a randomly sampled depth. Since displacement along $\mathbf{n}$ does not change the fixed-view projection,
\begin{equation}
    \pi(\mathbf{q}_i+d_i\mathbf{n})=\pi(\mathbf{q}_i).
\end{equation}

We use depth scales of $0.10$, $0.25$, and $0.45$, with $5{,}000$ training examples for each scale. The resulting training set contains $15{,}000$ examples. The raw target serializes the sampled coordinates
\begin{equation}
    Y_{\mathrm{raw}}=\{(x_i,y_i,z_i)\}_{i=1}^{N}.
\end{equation}
The observation-aligned target instead serializes normalized full-canvas marker centers,
\begin{equation}
    Y_{\mathrm{obs}}=\{(u_i,v_i,0)\}_{i=1}^{N}.
\end{equation}
At execution time, the compiler uses the Matplotlib projection matrix and coordinate transforms to lift each $(u_i,v_i,0)$ into a 3D point whose rendered center remains at $(u_i,v_i)$. The code template for the raw target are shown in Figure~\ref{fig:3d-raw-code} and the code template for the obs-aligned target code template is shown in Figure~\ref{fig:3d-obs-code}.

\subsection{Controlled Synthesis for Contour Charts and Auditing Details}

\label{sec:app_countour_data_synthesize}

\subsubsection{Dense-Field Generation and Example}
\label{sec:contour_dense_generation_details}

We generate $10{,}000$ training charts and $200$ independent test charts,

(1)\textbf{Create a grid.} We choose a grid size from
    $\{16,24,32\}$ and sample increasing, non-uniform $x$ and $y$
    coordinates. Their spans are $[4,7]$ and $[3,6]$, respectively,
    and their centers lie in $[-1,1]$.
(2)\textbf{Fill the grid with smooth values.} We smooth two random
    Gaussian-noise grids and add them with weights 1 and $0.30$.
    The first smoothing scale is chosen from $\{1.2,2.0,3.0\}$;
    the second is half that scale, with a minimum of $0.6$.
    We standardize the result, multiply it by a value in $[0.75,1.50]$,
    add an offset in $[-0.50,0.50]$, and round to four decimals.
    These steps run only during data generation.
(3)\textbf{Draw the raw chart.} We choose 3, 5, or 7 contour levels
    from evenly spaced field quantiles between $0.12$ and $0.88$.
    Levels must remain increasing after rounding.

\paragraph{Creating the observation-aligned target.}
Before adding labels, we read the paths from \texttt{ContourSet.allsegs}.
We remove redundant vertices with Ramer--Douglas--Peucker (RDP)
simplification, using tolerance $0.0015D_{xy}$, where $D_{xy}$ is the axes
diagonal in data coordinates. Closed paths remain closed, and coordinates
are rounded to four decimals. We store the paths in \texttt{ALLSEGS}, grouped
by their \texttt{LEVELS}, and draw them directly with \texttt{ContourSet}.
The grid values are no longer needed.

\paragraph{Checking the pair.}
Dense charts must have 3--45 contour components and at least 20 vertices.
We compare the raw and simplified paths at tolerances
$\{0.001,0.0025,0.005\}D_{xy}$. Their mean bidirectional coverage F1 must be
at least $0.990$. If this check fails, we halve the simplification tolerance
and retry, allowing at most six attempts.

We also render both programs independently. We keep a dense pair only if
the mean absolute pixel difference is at most $0.75$ and the fraction of
changed pixels is at most $0.03$. After both checks pass, the same raw PNG
is used as input for both targets, with the same sample ID. This construction
check uses coverage F1, rather than the evaluator's one-to-one point matching.

Figures~\ref{fig:contour-dense-raw-code}
and~\ref{fig:contour-dense-obs-code} show a smaller $5\times5$ example
with two levels.

\begin{figure*}[t]
\centering
\begin{minipage}[t]{0.48\textwidth}
\vspace{0pt}
\scriptsize
\begin{verbatim}
from pathlib import Path

import matplotlib
matplotlib.use("Agg")
import matplotlib.pyplot as plt
import numpy as np

LEVELS = np.array([0.5, 1.0])
LABEL_TEXT = {0.5: "0.5", 1.0: "1.0"}
LABEL_POSITIONS = [(-1.45, 0.), (-.72, 0.)]

# Materialized dense raw field.
X_VALUES = np.array([-2., -1., 0., 1., 2.])
YZ_ROWS = [
    (-2., [0., .1, .2, .1, 0.]),
    (-1., [.1, .5, .9, .5, .1]),
    ( 0., [.2, .9, 1.4, .9, .2]),
    ( 1., [.1, .5, .9, .5, .1]),
    ( 2., [0., .1, .2, .1, 0.]),
]
Y_VALUES = np.array([y for y, _ in YZ_ROWS])
Z_VALUES = np.array([z for _, z in YZ_ROWS])
if Z_VALUES.shape != (
    Y_VALUES.size, X_VALUES.size
):
    raise ValueError("invalid grid")
\end{verbatim}
\end{minipage}
\hfill
\begin{minipage}[t]{0.48\textwidth}
\vspace{0pt}
\scriptsize
\begin{verbatim}
fig, ax = plt.subplots(
    figsize=(6, 5), dpi=100
)
# Set the final transform before clabel.
ax.set_xlim(-2., 2.)
ax.set_ylim(-2., 2.)
ax.set_xlabel("X")
ax.set_ylabel("Y")
ax.set_title("Labeled Contour Plot")
ax.set_aspect("auto")
ax.grid(False)
fig.subplots_adjust(
    left=.13, right=.97,
    bottom=.14, top=.90,
)

# Render the raw field at LEVELS.
contour_set = ax.contour(
    X_VALUES, Y_VALUES, Z_VALUES,
    levels=LEVELS,
    colors="#202020",
    linewidths=1.25,
    linestyles="solid",
)

# Same visible labels on both targets.
fig.canvas.draw()
ax.clabel(
    contour_set,
    levels=LEVELS,
    fmt=LABEL_TEXT,
    manual=LABEL_POSITIONS,
    inline=False,
    fontsize=8,
    colors="#202020",
)
fig.savefig(
    Path(__file__).with_suffix(".png")
)
plt.close(fig)
\end{verbatim}
\end{minipage}
\caption{Compact dense-field raw target. The model predicts the materialized
$Z$ rows, while the visible image constrains the resulting contour paths.  Code continues from left to right.}
\label{fig:contour-dense-raw-code}
\end{figure*}

\begin{figure*}[t]
\centering
\begin{minipage}[t]{0.48\textwidth}
\vspace{0pt}
\scriptsize
\begin{verbatim}
from pathlib import Path

import matplotlib
matplotlib.use("Agg")
import matplotlib.pyplot as plt
import numpy as np
from matplotlib.contour import ContourSet

LEVELS = np.array([0.5, 1.0])
LABEL_TEXT = {0.5: "0.5", 1.0: "1.0"}
LABEL_POSITIONS = [(-1.45, 0.), (-.72, 0.)]

# ALLSEGS[k] stores LEVELS[k] components.
ALLSEGS = [
    [
        [
            [ 0.0000, -1.5714],
            [ 1.0000, -1.0000],
            [ 1.0000, -1.0000],
            [ 1.5714,  0.0000],
            [ 1.0000,  1.0000],
            [ 1.0000,  1.0000],
            [ 0.0000,  1.5714],
            [-1.0000,  1.0000],
            [-1.0000,  1.0000],
            [-1.5714,  0.0000],
            [-1.0000, -1.0000],
            [-1.0000, -1.0000],
            [ 0.0000, -1.5714],
        ],
    ],
    [
        [
            [ 0.0000, -0.8000],
            [ 0.8000,  0.0000],
            [ 0.0000,  0.8000],
            [-0.8000,  0.0000],
            [ 0.0000, -0.8000],
        ],
    ],
]
\end{verbatim}
\end{minipage}
\hfill
\begin{minipage}[t]{0.48\textwidth}
\vspace{0pt}
\scriptsize
\begin{verbatim}
fig, ax = plt.subplots(
    figsize=(6, 5), dpi=100
)
# Set the final transform before clabel.
ax.set_xlim(-2., 2.)
ax.set_ylim(-2., 2.)
ax.set_xlabel("X")
ax.set_ylabel("Y")
ax.set_title("Labeled Contour Plot")
ax.set_aspect("auto")
ax.grid(False)
fig.subplots_adjust(
    left=.13, right=.97,
    bottom=.14, top=.90,
)

# Directly render level-grouped paths.
contour_set = ContourSet(
    ax, LEVELS, ALLSEGS,
    filled=False,
    colors="#202020",
    linewidths=1.25,
    linestyles="solid",
)

# Same visible labels on both targets.
fig.canvas.draw()
ax.clabel(
    contour_set,
    levels=LEVELS,
    fmt=LABEL_TEXT,
    manual=LABEL_POSITIONS,
    inline=False,
    fontsize=8,
    colors="#202020",
)
fig.savefig(
    Path(__file__).with_suffix(".png")
)
plt.close(fig)
\end{verbatim}
\end{minipage}
\caption{Observation-aligned counterpart of Figure~\ref{fig:contour-dense-raw-code}.
The serialized
paths are extracted from the raw example rather than drawn independently.}
\label{fig:contour-dense-obs-code}
\end{figure*}

\subsubsection{Analytic-Formula Generation and Example}
\label{sec:contour_formula_generation_details}

This setting uses $10{,}000$ training charts and $200$ independent test
charts. Training and testing share the same formula families and parameter
ranges, but use different seeds and formula instances. The families are:
\begin{itemize}
    \item \textbf{Gaussian mixtures}: a sum of 2--4 signed Gaussian peaks,
    with widths of $0.14$--$0.32$ of the axis span.
    \item \textbf{Separable waves}: an $x$-direction sine and a
    $y$-direction cosine, with $0.65$--$2.10$ cycles per axis span.
    \item \textbf{Rotated quadratics}: a quadratic in shifted and rotated
    coordinates, with axis scales of $0.35$--$0.70$ of the span.
\end{itemize}
Biases and linear terms follow the corresponding formula template.

For each chart, we sample $x$ and $y$ spans from $[5,7]$ and $[4,6]$,
centers from $[-0.75,0.75]$, and a grid size of 64 or 96. We round the
formula parameters to six decimals, evaluate the field, and select 3, 5,
or 7 levels from its $0.12$--$0.88$ quantiles. Levels are also stored to six
decimals. The raw program stores this fixed formula rather than the full
grid or the parameter sampler.

The observation-aligned target uses the same path extraction,
simplification, \texttt{ContourSet}, and \texttt{clabel} procedure as above.
Formula charts must have 1--100 components and at least 20 vertices.
The geometry threshold remains $0.990$; the pixel thresholds are $2.0$
for mean absolute difference and $0.10$ for the fraction of changed pixels.
Only accepted pairs share the same input PNG.

We record formula, contour-geometry, and image hashes to exclude repeated
instances across training and testing.
Figures~\ref{fig:contour-formula-raw-code}
and~\ref{fig:contour-formula-obs-code} use a $9\times9$ grid for readability;
the generated data use grids of size 64 or 96. The observation-aligned
paths are extracted from the corresponding raw example.

\begin{figure*}[t]
\centering
\begin{minipage}[t]{0.48\textwidth}
\vspace{0pt}
\scriptsize
\begin{verbatim}
from pathlib import Path

import matplotlib
matplotlib.use("Agg")
import matplotlib.pyplot as plt
import numpy as np

LEVELS = np.array([0.5, 1.0])
LABEL_TEXT = {0.5: "0.5", 1.0: "1.0"}
LABEL_POSITIONS = [(-.80, 0.), (-1.15, 0.)]

# Compact raw formula illustration.
GRID_SIZE = 9
X_VALUES = np.linspace(-1.5, 1.5, GRID_SIZE)
Y_VALUES = np.linspace(-1.5, 1.5, GRID_SIZE)
X_GRID, Y_GRID = np.meshgrid(
    X_VALUES, Y_VALUES
)
Z_VALUES = (
    0.100000
    + 0.800000 * X_GRID ** 2
    + 1.200000 * Y_GRID ** 2
    + 0.200000 * X_GRID
)
\end{verbatim}
\end{minipage}
\hfill
\begin{minipage}[t]{0.48\textwidth}
\vspace{0pt}
\scriptsize
\begin{verbatim}
fig, ax = plt.subplots(
    figsize=(6, 5), dpi=100
)
# Set the final transform before clabel.
ax.set_xlim(-1.5, 1.5)
ax.set_ylim(-1.5, 1.5)
ax.set_xlabel("X")
ax.set_ylabel("Y")
ax.set_title("Labeled Contour Plot")
ax.set_aspect("auto")
ax.grid(False)
fig.subplots_adjust(
    left=.13, right=.97,
    bottom=.14, top=.90,
)

# Render the raw field at LEVELS.
contour_set = ax.contour(
    X_GRID, Y_GRID, Z_VALUES,
    levels=LEVELS,
    colors="#202020",
    linewidths=1.25,
    linestyles="solid",
)

# Same visible labels on both targets.
fig.canvas.draw()
ax.clabel(
    contour_set,
    levels=LEVELS,
    fmt=LABEL_TEXT,
    manual=LABEL_POSITIONS,
    inline=False,
    fontsize=8,
    colors="#202020",
)
fig.savefig(
    Path(__file__).with_suffix(".png")
)
plt.close(fig)
\end{verbatim}
\end{minipage}
\caption{Compact analytic raw target. Code continues from left to right.}
\label{fig:contour-formula-raw-code}
\end{figure*}

\begin{figure*}[t]
\centering
\begin{minipage}[t]{0.48\textwidth}
\vspace{0pt}
\scriptsize
\begin{verbatim}
from pathlib import Path

import matplotlib
matplotlib.use("Agg")
import matplotlib.pyplot as plt
import numpy as np
from matplotlib.contour import ContourSet

LEVELS = np.array([0.5, 1.0])
LABEL_TEXT = {0.5: "0.5", 1.0: "1.0"}
LABEL_POSITIONS = [(-.80, 0.), (-1.15, 0.)]

# ALLSEGS[k] stores LEVELS[k] components.
ALLSEGS = [
    [
        [
            [-0.3750, -0.5185],
            [-0.6518, -0.3750],
            [-0.7500, -0.2222],
            [-0.8269,  0.0000],
            [-0.7500,  0.2222],
            [-0.6518,  0.3750],
            [-0.3750,  0.5185],
            [ 0.0000,  0.5463],
            [ 0.3750,  0.4074],
            [ 0.4148,  0.3750],
            [ 0.5682,  0.0000],
            [ 0.4148, -0.3750],
            [ 0.3750, -0.4074],
            [ 0.0000, -0.5463],
            [-0.3750, -0.5185],
        ],
    ],
    [
        [
            [-0.3750, -0.8333],
            [-0.6429, -0.7500],
            [-0.7500, -0.6944],
            [-1.0817, -0.3750],
            [-1.1250, -0.2500],
            [-1.1842,  0.0000],
            [-1.1250,  0.2500],
            [-1.0817,  0.3750],
            [-0.7500,  0.6944],
            [-0.6429,  0.7500],
            [-0.3750,  0.8333],
            [ 0.0000,  0.8500],
            [ 0.3750,  0.7667],
            [ 0.4091,  0.7500],
            [ 0.7500,  0.4722],
            [ 0.8272,  0.3750],
            [ 0.9265,  0.0000],
            [ 0.8272, -0.3750],
            [ 0.7500, -0.4722],
            [ 0.4091, -0.7500],
            [ 0.3750, -0.7667],
            [ 0.0000, -0.8500],
            [-0.3750, -0.8333],
        ],
    ],
]
\end{verbatim}
\end{minipage}
\hfill
\begin{minipage}[t]{0.48\textwidth}
\vspace{0pt}
\scriptsize
\begin{verbatim}
fig, ax = plt.subplots(
    figsize=(6, 5), dpi=100
)
# Set the final transform before clabel.
ax.set_xlim(-1.5, 1.5)
ax.set_ylim(-1.5, 1.5)
ax.set_xlabel("X")
ax.set_ylabel("Y")
ax.set_title("Labeled Contour Plot")
ax.set_aspect("auto")
ax.grid(False)
fig.subplots_adjust(
    left=.13, right=.97,
    bottom=.14, top=.90,
)

# Directly render level-grouped paths.
contour_set = ContourSet(
    ax, LEVELS, ALLSEGS,
    filled=False,
    colors="#202020",
    linewidths=1.25,
    linestyles="solid",
)

# Same visible labels on both targets.
fig.canvas.draw()
ax.clabel(
    contour_set,
    levels=LEVELS,
    fmt=LABEL_TEXT,
    manual=LABEL_POSITIONS,
    inline=False,
    fontsize=8,
    colors="#202020",
)
fig.savefig(
    Path(__file__).with_suffix(".png")
)
plt.close(fig)
\end{verbatim}
\end{minipage}
\caption{Observation-aligned counterpart of Figure~\ref{fig:contour-formula-raw-code}.
Its nested \texttt{ALLSEGS} is extracted from the same raw-grid contours.
The analytic expression and sampled field are replaced by level-grouped
geometry}
\label{fig:contour-formula-obs-code}
\end{figure*}

\subsection{More Experimental Details For Aggregation-induced and Normalization-induced Mismatch}
\subsubsection{Implementation Details}
For Qwen models, we finetune them with Llama-Factory~\citep{zheng2024llamafactoryunifiedefficientfinetuning} with the learning rate of 2.0e-4, global batch size of 16 and train the models for 4 epochs. For Internvl models, we finetune them using the official code and using the same hyperparameters. 

\subsubsection{More Evaluation Metric Details}
\label{sec:aggr_normalization_eval_metric_detail}
We execute both the reference and generated programs and extract numerical quantities from the resulting Matplotlib artists which are the exactly the  observation in the target chart or rendered chart objects. Thus, the evaluation does not reward matching source-code tokens, variable names, raw-data literals. \textbf{\emph{It instead measures whether the executed prediction recovers the observable chart-level observations in the target chart: histogram bin heights, boxplot five-number geometry, and pie wedge percentage and makes the evaluation on assessing the true visually identifiable informations from the chart, which is aligned with the visual reproduction goal of Chart2Code task. Two programs receive the same score whenever they produce the same artist-level quantities, regardless of whether they use raw data or an observation-aligned representation. }}

\paragraph{Hist-Value.}
For each histogram subplot $a$, we extract the  bin heights from the
reference and predicted charts, denoted as
$G^h_a=\{h^g_i\}_{i=1}^{m}$ and
$P^h_a=\{h^p_j\}_{j=1}^{n}$.  We compare the two sets by a one-to-one
matching.  The cost of matching a reference bin $i$ to a predicted bin $j$
is
\begin{equation}
\label{eq:hist-cost}
d^h_{ij}
=
\min\left(
\frac{|h^g_i-h^p_j|}{s^g_a+\epsilon},
1
\right),
\end{equation}
where $s^g_a$ is the value-axis scale of the reference subplot.  For
vertical histograms, this is the $y$-axis scale; for horizontal histograms,
this is the $x$-axis scale.  Let $\mathcal{M}^h_a$ be the minimum-cost
one-to-one matching between $G^h_a$ and $P^h_a$.  The unmatched-bin penalty is
\begin{equation}
\label{eq:hist-unmatched}
U^h_a=\max(m,n)-|\mathcal{M}^h_a|.
\end{equation}
The subplot-level histogram error is
\begin{equation}
\label{eq:hist-error}
E^h_a
=
\frac{
\sum_{(i,j)\in\mathcal{M}^h_a} d^h_{ij}
+
U^h_a
}{
\max(m,n,1)
}.
\end{equation}
We aggregate over subplots by weighting each subplot by the number of
compared bins:
\begin{equation}
\label{eq:hist-weight}
w^h_a=\max(|G^h_a|,|P^h_a|,1),
\end{equation}
\begin{equation}
\label{eq:hist-avg-error}
\bar{E}^h
=
\frac{\sum_a w^h_a E^h_a}
{\sum_a w^h_a}.
\end{equation}
The final histogram value score is
\begin{equation}
\label{eq:hist-value}
\mathrm{Hist\text{-}Value}
=
100\cdot(1-\bar{E}^h).
\end{equation}

\paragraph{Box-Value.}
For each boxplot, we represent every box by the five statistics that define
its geometry:
\begin{equation}
\label{eq:box-vector}
\mathbf{b}
=
[\mathrm{whislo},q_1,\mathrm{med},q_3,\mathrm{whishi}].
\end{equation}
For subplot $a$, let
$G^b_a=\{\mathbf{b}^g_i\}_{i=1}^{m}$ and
$P^b_a=\{\mathbf{b}^p_j\}_{j=1}^{n}$ be the reference and predicted box
sets.  The cost of matching two boxes is the average normalized absolute
error over the five statistics:
\begin{equation}
\label{eq:box-cost}
d^b_{ij}
=
\min\left(
\frac{\|\mathbf{b}^g_i-\mathbf{b}^p_j\|_1}
{5(s^g_{a,o}+\epsilon)},
1
\right).
\end{equation}
Here $o$ is the value axis of the boxplot: $o=y$ for vertical boxplots and
$o=x$ for horizontal boxplots.  The scale $s^g_{a,o}$ is always taken from
the corresponding reference subplot axis.  Let $\mathcal{M}^b_a$ be the
minimum-cost one-to-one matching between $G^b_a$ and $P^b_a$.  The unmatched
penalty is
\begin{equation}
\label{eq:box-unmatched}
U^b_a=\max(m,n)-|\mathcal{M}^b_a|.
\end{equation}
The subplot-level boxplot error is
\begin{equation}
\label{eq:box-error}
E^b_a
=
\frac{
\sum_{(i,j)\in\mathcal{M}^b_a} d^b_{ij}
+
U^b_a
}{
\max(m,n,1)
}.
\end{equation}
We average the error over subplots using
\begin{equation}
\label{eq:box-weight}
w^b_a=\max(|G^b_a|,|P^b_a|,1),
\end{equation}
\begin{equation}
\label{eq:box-avg-error}
\bar{E}^b
=
\frac{\sum_a w^b_a E^b_a}
{\sum_a w^b_a}.
\end{equation}
The final boxplot value score is
\begin{equation}
\label{eq:box-value}
\mathrm{Box\text{-}Value}
=
100\cdot(1-\bar{E}^b).
\end{equation}

\paragraph{Pie-F1.}
For pie charts, we extract the percentage of each wedge.  Let
$G^p=\{r^g_i\}_{i=1}^{m}$ and
$P^p=\{r^p_j\}_{j=1}^{n}$ denote the reference and predicted wedge
percentage sets.  A reference wedge and a predicted wedge can be matched if
their percentages differ by no more than a tolerance $\tau$, here we set $\tau$ to 0.1:
\begin{equation}
\label{eq:pie-match}
|r^g_i-r^p_j|\leq \tau.
\end{equation}
We find the maximum one-to-one matching under this criterion and denote the
number of matched wedges as $M$.  Precision and recall are
\begin{equation}
\label{eq:pie-precision}
P_{\mathrm{pie}}=\frac{M}{\max(n,1)},
\end{equation}
\begin{equation}
\label{eq:pie-recall}
R_{\mathrm{pie}}=\frac{M}{\max(m,1)}.
\end{equation}
The final pie score is the F1 score:
\begin{equation}
\label{eq:pie-f1}
\mathrm{Pie\text{-}F1}
=
100\cdot
\frac{2P_{\mathrm{pie}}R_{\mathrm{pie}}}
{P_{\mathrm{pie}}+R_{\mathrm{pie}}+\epsilon}.
\end{equation}

\begin{table*}[t]
\centering
\scriptsize
\setlength{\tabcolsep}{1.1pt}
\renewcommand{\arraystretch}{1.05}
\begin{adjustbox}{max size={\textwidth}{0.94\textheight},center}
\begin{tabular}{ll*{20}{c}}
\toprule
 &  & \multicolumn{5}{c}{Box} & \multicolumn{5}{c}{Pie} & \multicolumn{5}{c}{Hist} & \multicolumn{5}{c}{Hist-ori} \\
\cmidrule(lr){3-7}\cmidrule(lr){8-12}\cmidrule(lr){13-17}\cmidrule(lr){18-22}
Model & Supervision
& Ex. & Value & Text & Color & TC Avg.
& Ex. & F1 & Text & Color & TC Avg.
& Ex. & Value & Text & Color & TC Avg.
& Ex. & Value & Text & Color & TC Avg. \\
\midrule

\multicolumn{22}{c}{\textit{ChartMimic-ori}} \\
\midrule
InternVL3-8B & No-modified
& 80 & 54.8 & 60.6 & 67.8 & 64.2
& 90 & 53.9 & 62.7 & 78.8 & 70.8
& \textbf{85} & \textbf{40.1} & \textbf{80.9} & \textbf{76.3} & \textbf{78.6}
& -- & -- & -- & -- & -- \\
InternVL3-8B & Modified
& \textbf{88} & \textbf{57.5} & \textbf{65.5} & \textbf{74.9} & \textbf{70.2}
& \textbf{100} & \textbf{84.5} & \textbf{93.9} & \textbf{87.8} & \textbf{90.8}
& 75 & 38.4 & 73.8 & 68.0 & 70.9
& -- & -- & -- & -- & -- \\
InternVL3-14B & No-modified
& \textbf{92} & 53.4 & \textbf{73.6} & \textbf{77.2} & \textbf{75.4}
& 90 & 60.3 & 65.7 & 79.9 & 72.8
& \textbf{80} & 35.5 & \textbf{77.8} & \textbf{71.7} & \textbf{74.7}
& -- & -- & -- & -- & -- \\
InternVL3-14B & Modified
& 88 & \textbf{66.4} & 66.6 & 72.9 & 69.8
& \textbf{100} & \textbf{79.1} & \textbf{86.7} & \textbf{88.9} & \textbf{87.8}
& 75 & \textbf{51.6} & 74.6 & 68.3 & 71.4
& -- & -- & -- & -- & -- \\
Qwen2.5-VL-3B & No-modified
& 68 & 45.2 & 45.5 & 55.8 & 50.6
& 95 & 53.2 & 63.8 & 81.7 & 72.7
& 85 & 33.8 & \textbf{81.2} & \textbf{75.5} & \textbf{78.3}
& -- & -- & -- & -- & -- \\
Qwen2.5-VL-3B & Modified
& \textbf{88} & \textbf{63.2} & \textbf{67.0} & \textbf{72.6} & \textbf{69.8}
& 95 & \textbf{84.8} & \textbf{83.2} & \textbf{82.7} & \textbf{82.9}
& 85 & \textbf{40.4} & 80.2 & 72.3 & 76.3
& -- & -- & -- & -- & -- \\
Qwen2.5-VL-7B & No-modified
& 88 & 61.6 & 71.3 & 68.2 & 69.8
& \textbf{100} & 65.5 & 75.1 & \textbf{86.7} & \textbf{80.9}
& 75 & 34.0 & 71.8 & 66.4 & 69.1
& -- & -- & -- & -- & -- \\
Qwen2.5-VL-7B & Modified
& \textbf{92} & \textbf{70.6} & \textbf{74.3} & \textbf{74.6} & \textbf{74.5}
& 90 & \textbf{76.7} & \textbf{80.0} & 79.4 & 79.7
& \textbf{95} & \textbf{50.1} & \textbf{93.1} & \textbf{82.0} & \textbf{87.6}
& -- & -- & -- & -- & -- \\

\midrule
\multicolumn{22}{c}{\textit{ChartMimic-Both executable}} \\
\midrule
InternVL3-8B & No-modified
& 100 & \textbf{67.8} & \textbf{74.5} & \textbf{84.7} & \textbf{79.6}
& 100 & 59.9 & 69.7 & 87.6 & 78.6
& 100 & 49.7 & 95.3 & 92.8 & 94.0
& -- & -- & -- & -- & -- \\
InternVL3-8B & Modified
& 100 & 65.2 & 73.0 & 84.5 & 78.7
& 100 & \textbf{85.6} & \textbf{94.3} & \textbf{89.1} & \textbf{91.7}
& 100 & \textbf{50.0} & \textbf{98.7} & \textbf{93.3} & \textbf{96.0}
& -- & -- & -- & -- & -- \\
InternVL3-14B & No-modified
& 100 & 58.9 & \textbf{80.5} & 83.6 & \textbf{82.1}
& 100 & 67.0 & 73.0 & 88.8 & 80.9
& 100 & 50.9 & 96.2 & 89.1 & 92.7
& -- & -- & -- & -- & -- \\
InternVL3-14B & Modified
& 100 & \textbf{76.4} & 75.2 & \textbf{84.3} & 79.7
& 100 & \textbf{81.0} & \textbf{87.4} & \textbf{89.2} & \textbf{88.3}
& 100 & \textbf{70.4} & \textbf{100} & \textbf{90.0} & \textbf{95.0}
& -- & -- & -- & -- & -- \\
Qwen2.5-VL-3B & No-modified
& 100 & 65.1 & 69.8 & 83.1 & 76.4
& 100 & 53.6 & 67.8 & 85.8 & 76.8
& 100 & 40.4 & 95.6 & \textbf{88.3} & \textbf{91.9}
& -- & -- & -- & -- & -- \\
Qwen2.5-VL-3B & Modified
& 100 & \textbf{77.9} & \textbf{76.2} & \textbf{83.9} & \textbf{80.0}
& 100 & \textbf{88.6} & \textbf{86.8} & \textbf{87.1} & \textbf{87.0}
& 100 & \textbf{49.4} & 95.6 & 87.6 & 91.6
& -- & -- & -- & -- & -- \\
Qwen2.5-VL-7B & No-modified
& 100 & 69.0 & 80.8 & 77.4 & 79.1
& 100 & 68.6 & 75.6 & 88.2 & 81.9
& 100 & 48.6 & 95.4 & \textbf{88.3} & \textbf{91.8}
& -- & -- & -- & -- & -- \\
Qwen2.5-VL-7B & Modified
& 100 & \textbf{74.9} & \textbf{82.2} & \textbf{81.0} & \textbf{81.6}
& 100 & \textbf{85.2} & \textbf{88.8} & \textbf{88.3} & \textbf{88.6}
& 100 & \textbf{54.1} & \textbf{97.3} & 85.6 & 91.5
& -- & -- & -- & -- & -- \\

\midrule
\multicolumn{22}{c}{\textit{ChartX-ori}} \\
\midrule
InternVL3-8B & No-modified
& \textbf{96} & \textbf{93.3} & \textbf{90.9} & 74.7 & 82.8
& 98 & 88.5 & 90.5 & 95.0 & 92.7
& 90 & 43.5 & 86.9 & 78.4 & 82.7
& 50 & 16.0 & 46.0 & 20.8 & 33.4 \\
InternVL3-8B & Modified
& 92 & 91.0 & 86.7 & \textbf{79.4} & \textbf{83.1}
& \textbf{99} & \textbf{93.9} & \textbf{95.6} & \textbf{96.0} & \textbf{95.8}
& \textbf{94} & \textbf{80.8} & \textbf{92} & \textbf{86.8} & \textbf{89.4}
& \textbf{60} & \textbf{31.7} & \textbf{53.5} & \textbf{28.2} & \textbf{40.8} \\
InternVL3-14B & No-modified
& \textbf{94} & 88.9 & \textbf{87.4} & 70.8 & 79.1
& 99 & 92.6 & 92.6 & 96.4 & 94.5
& 98 & 51.8 & 95.6 & 92.6 & 94.1
& 68 & 14.8 & 62.9 & 25.8 & 44.3 \\
InternVL3-14B & Modified
& 90 & \textbf{89.0} & 85.4 & \textbf{78.1} & \textbf{81.7}
& \textbf{100} & \textbf{96} & \textbf{95.5} & \textbf{96.7} & \textbf{96.1}
& \textbf{100} & \textbf{88.4} & \textbf{96.7} & \textbf{94.4} & \textbf{95.5}
& \textbf{70} & \textbf{26.9} & \textbf{67.3} & \textbf{25.9} & \textbf{46.6} \\
Qwen2.5-VL-3B & No-modified
& 88 & 84.8 & 84.7 & 64.2 & 74.4
& \textbf{99} & 84.1 & 90.3 & \textbf{95.5} & 92.9
& \textbf{92} & 52.0 & \textbf{89.9} & \textbf{86.2} & \textbf{88.0}
& \textbf{70} & 17.7 & \textbf{62.1} & \textbf{30.9} & \textbf{46.5} \\
Qwen2.5-VL-3B & Modified
& \textbf{94} & \textbf{93.8} & \textbf{92.3} & \textbf{79.1} & \textbf{85.7}
& 97 & \textbf{92.6} & \textbf{93.6} & 94.1 & \textbf{93.9}
& 90 & \textbf{64.8} & 89.6 & 82.5 & 86.0
& 56 & \textbf{21.2} & 51.0 & 22.9 & 37.0 \\
Qwen2.5-VL-7B & No-modified
& \textbf{88} & 85.0 & 84.3 & 71.5 & 77.9
& 99 & 88.7 & 91.7 & \textbf{96.5} & 94.1
& 96 & 55.5 & 93.3 & 91.3 & 92.3
& \textbf{72} & 13.9 & 67.7 & 20.3 & 44.0 \\
Qwen2.5-VL-7B & Modified
& 86 & \textbf{85.9} & \textbf{85} & \textbf{76.9} & \textbf{81.0}
& 99 & \textbf{94.6} & \textbf{95.5} & 96.4 & \textbf{96.0}
& \textbf{100} & \textbf{93.4} & \textbf{99.3} & \textbf{94.2} & \textbf{96.8}
& \textbf{72} & \textbf{23.1} & \textbf{69.5} & \textbf{30.5} & \textbf{50.0} \\

\midrule
\multicolumn{22}{c}{\textit{ChartX-Both executable}} \\
\midrule
InternVL3-8B & No-modified
& 100 & 97.1 & \textbf{94.3} & 78.7 & 86.5
& 100 & 91.3 & 92.8 & \textbf{97.9} & 95.3
& 100 & 49.4 & 96.9 & 89.7 & 93.3
& \textbf{100} & 41.1 & \textbf{91.1} & 66.3 & \textbf{78.7} \\
InternVL3-8B & Modified
& 100 & \textbf{98.8} & 93.9 & \textbf{86.3} & \textbf{90.1}
& 100 & \textbf{95.8} & \textbf{96.7} & 97.0 & \textbf{96.8}
& 100 & \textbf{85.6} & \textbf{98.4} & \textbf{91.7} & \textbf{95.1}
& \textbf{100} & \textbf{65.1} & 88.6 & \textbf{68.0} & 78.3 \\
InternVL3-14B & No-modified
& 100 & 94.3 & 92.6 & 73.9 & 83.3
& 100 & 93.5 & 93.6 & 97.4 & 95.5
& 100 & 52.9 & \textbf{97.6} & \textbf{94.5} & \textbf{96.0}
& \textbf{100} & 21.3 & 90.9 & 37.0 & 64.0 \\
InternVL3-14B & Modified
& 100 & \textbf{98.8} & \textbf{94.5} & \textbf{86.7} & \textbf{90.6}
& 100 & \textbf{97.0} & \textbf{96.0} & \textbf{97.7} & \textbf{96.8}
& 100 & \textbf{89.0} & 96.6 & 94.2 & 95.4
& \textbf{100} & \textbf{37.6} & \textbf{97.3} & \textbf{40.4} & \textbf{68.8} \\
Qwen2.5-VL-3B & No-modified
& 100 & 96.1 & 97.2 & 73.3 & 85.2
& 100 & 85.5 & 91.2 & 96.7 & 93.9
& 100 & 56.7 & 97.4 & \textbf{93.7} & \textbf{95.6}
& \textbf{100} & 37.3 & \textbf{95.2} & \textbf{64.7} & \textbf{80.0} \\
Qwen2.5-VL-3B & Modified
& 100 & \textbf{99.8} & \textbf{98.0} & \textbf{83.8} & \textbf{90.9}
& 100 & \textbf{95.5} & \textbf{96.5} & \textbf{97.0} & \textbf{96.8}
& 100 & \textbf{72.8} & \textbf{99.5} & 91.3 & 95.4
& \textbf{100} & \textbf{50.5} & 92.4 & 54.5 & 73.5 \\
Qwen2.5-VL-7B & No-modified
& 100 & 96.7 & 95.6 & 81.2 & 88.4
& 100 & 89.5 & 92.6 & \textbf{97.5} & 95.0
& 100 & 57.8 & 97.2 & \textbf{95.1} & 96.1
& \textbf{100} & 21.8 & 95.0 & 33.5 & 64.3 \\
Qwen2.5-VL-7B & Modified
& 100 & \textbf{99.9} & \textbf{100} & \textbf{90.1} & \textbf{95.0}
& 100 & \textbf{95.5} & \textbf{96.4} & 97.4 & \textbf{96.9}
& 100 & \textbf{94.4} & \textbf{99.3} & 93.9 & \textbf{96.6}
& \textbf{100} & \textbf{35.1} & \textbf{95.8} & \textbf{42.1} & \textbf{68.9} \\

\bottomrule
\end{tabular}
\end{adjustbox}
\caption{Full results on ChartMimic and ChartX under the original evaluation setting and the both-executable subset. No-modified denotes training on original raw-code supervision, while Modified denotes training on observation-aligned supervision. The better score between No-modified and Modified for each model and metric is highlighted in bold. Hist-ori reports the additional original ChartX histogram results; the corresponding columns are left as placeholders for ChartMimic.}
\label{tab:full_results_grouped_with_hist_ori}
\end{table*}

\subsubsection{$\tau$-sensitive analysis of Pie-F1 metrics}
\label{sec:app_tausensitiveanalysis}

We evaluate Pie-F1 on ChartX with matching tolerances $\tau$ from
0.1 to 2.0 percentage points. A predicted wedge matches a reference
wedge when their percentages differ by at most $\tau$.
Table~\ref{tab:chartx_pie_tau} shows that observation-aligned
supervision achieves higher Pie-F1 than raw-code supervision at every
tested tolerance for both models. The gap narrows for Qwen2.5-VL-7B
as the tolerance increases, but the advantage remains. Thus, the
improvement is not limited to the default tolerance of $\tau=0.1$.

\begin{table}[t]
    \centering
    \small
    \setlength{\tabcolsep}{3pt}
    \begin{tabular}{@{}c cc cc@{}}
        \toprule
        & \multicolumn{2}{c}{Qwen2.5-VL-7B}
        & \multicolumn{2}{c}{InternVL3-14B} \\
        \cmidrule(lr){2-3} \cmidrule(lr){4-5}
        $\tau$ & Raw & Obs. & Raw & Obs. \\
        \midrule
        0.1 & 88.7 & \textbf{94.6} & 92.6 & \textbf{96.0} \\
        0.3 & 92.2 & \textbf{95.6} & 94.0 & \textbf{97.3} \\
        0.5 & 94.0 & \textbf{96.0} & 94.7 & \textbf{97.3} \\
        0.7 & 95.0 & \textbf{96.6} & 95.1 & \textbf{97.7} \\
        1.0 & 95.6 & \textbf{97.0} & 95.8 & \textbf{98.1} \\
        2.0 & 97.7 & \textbf{97.9} & 96.8 & \textbf{99.2} \\
        \bottomrule
    \end{tabular}
    \caption{Pie-F1 (\%) on ChartX at different matching tolerances
    $\tau$ (percentage points). Obs. denotes observation-aligned
    supervision. Bold marks the higher score within each model pair.}
    \label{tab:chartx_pie_tau}
\end{table}

\subsection{Additional Validation}
\label{app:additional_results}

\paragraph{Three-seed training.}
We train Qwen2.5-VL-3B with three random seeds under each supervision
setting. Table~\ref{tab:qwen3b_multiseed} reports the mean and standard
deviation across seeds. Observation-aligned supervision achieves
higher mean recovery scores in all six benchmark--chart settings,
showing that the advantage remains when averaging over training seeds.

\begin{table}[t]
\centering
\small
\setlength{\tabcolsep}{4pt}
\begin{tabular}{@{}lcc@{}}
\toprule
Chart & Raw & Obs.-aligned \\
\midrule
\multicolumn{3}{c}{\textit{ChartMimic}} \\
\midrule
Box  & $47.2 \pm 2.1$ & $\mathbf{62.0 \pm 0.2}$ \\
Hist & $33.6 \pm 0.2$ & $\mathbf{39.9 \pm 0.5}$ \\
Pie  & $55.0 \pm 2.2$ & $\mathbf{88.0 \pm 2.9}$ \\
\midrule
\multicolumn{3}{c}{\textit{ChartX}} \\
\midrule
Box  & $85.2 \pm 0.4$ & $\mathbf{90.9 \pm 2.7}$ \\
Hist & $50.1 \pm 2.0$ & $\mathbf{68.4 \pm 3.2}$ \\
Pie  & $85.4 \pm 1.1$ & $\mathbf{92.5 \pm 1.2}$ \\
\bottomrule
\end{tabular}
\caption{Qwen2.5-VL-3B recovery scores (\%), reported as mean
$\pm$ standard deviation across three training seeds.
Box and Hist use Value; Pie uses Pie-F1.
Bold marks the higher mean.}
\label{tab:qwen3b_multiseed}
\end{table}

\paragraph{Layout recovery.}
Table~\ref{tab:chartmimic_layout} reports ChartMimic layout scores
for two models. Observation-aligned supervision increases the
Qwen2.5-VL-7B score from 87.69 to 90.76, while InternVL3-14B
remains at 87.69. 

\begin{table}[t]
\centering
\small
\setlength{\tabcolsep}{4pt}
\begin{tabular}{@{}lcc@{}}
\toprule
Model & Raw & Obs.-aligned \\
\midrule
Qwen2.5-VL-7B & 87.69 & \textbf{90.76} \\
InternVL3-14B & 87.69 & 87.69 \\
\bottomrule
\end{tabular}
\caption{ChartMimic layout scores for selected models.
Bold marks the higher score; tied scores are not highlighted.}
\label{tab:chartmimic_layout}
\end{table}

\subsubsection{Full Main Experiment Results}
\label{sec:full_main_exp_res}
We show the full experimental results in Table~\ref{tab:full_results_grouped_with_hist_ori}.

\subsubsection{ChartX Hist Modification Details}
\label{sec:ChartX_hist_modification_details}
We construct a canonicalized ChartX-Hist subset by converting histogram-like bar charts with categorical or discontinuous bin labels into continuous-bin histogram representations. The numerical bin heights are unchanged, but the x-axis bin geometry and tick representation are standardized. Figure~\ref{fig:chartxhistmodify} shows an example of our ChartX-Hist canonicalization. The original ChartX instance uses age-bin labels as categorical x ticks, while the canonicalized version represents the same heights over continuous bin intervals. This transformation preserves the y-values used for value recovery but changes the x-axis rendering.

\paragraph{Why Modification} The training histograms samples  use continuous numerical intervals, whereas some samples in  ChartX-Hist-ori use discrete or categorical bin labels, creating a  distribution shift. We therefore use the canonicalized subset  to evaluate continuous-interval histograms under the target setting,  while retaining Hist-ori as a complementary out-of-distribution  evaluation.      2. More importantly, on the unmodified Hist-ori subset, all four models improve value-recovery under both the all-sample and both-executable settings. 
\subsubsection{Both-exec setting Evaluation Sample Coverage}
\label{sec:both_exec_eval_sample_coverage}

\begin{table}[t]
    \centering
    \small
    \setlength{\tabcolsep}{3pt}
    \renewcommand{\arraystretch}{1.08}
    \begin{tabular}{@{}llcc@{}}
        \toprule
        Model & Type & ChartMimic & ChartX \\
        \midrule

        \multirow{3}{*}{\makecell[l]{InternVL3-8B-\\Instruct}}
        & Box  & 76.0 & 88.0 \\
        & Pie  & 90.0 & 97.0 \\
        & Hist & 65.0 & 86.0 \\

        \midrule
        \multirow{3}{*}{\makecell[l]{InternVL3-14B-\\Instruct}}
        & Box  & 84.0 & 84.0 \\
        & Pie  & 90.0 & 99.0 \\
        & Hist & 60.0 & 98.0 \\

        \midrule
        \multirow{3}{*}{\makecell[l]{Qwen2.5-VL-3B}}
        & Box  & 64.0 & 82.0 \\
        & Pie  & 90.0 & 97.0 \\
        & Hist & 70.0 & 82.0 \\

        \midrule
        \multirow{3}{*}{\makecell[l]{Qwen2.5-VL-7B}}
        & Box  & 84.0 & 74.0 \\
        & Pie  & 90.0 & 98.0 \\
        & Hist & 70.0 & 96.0 \\

        \bottomrule
    \end{tabular}
    \caption{
        Both-exec evaluation sample coverage rate
    }
    \label{tab:both_exec_sample_coverage}
\end{table}

The both-executable setting is intended only as a diagnostic analysis of whether observation-aligned supervision improves numerical recovery when both methods successfully execute. It is not intended to replace the standard all-sample evaluation.
To make the scope of this analysis explicit, we report the proportion of samples included in the both-executable subset for each model, chart type, and dataset in Table~\ref{tab:both_exec_sample_coverage}. The coverage ranges from 60\% to 99\%, therefore the diagnostic analysis still covers most evaluation samples. 

\subsubsection{Controlled Experiment Code Template}
\label{sec:controlled_exp_for_aggr_normal}

\begin{figure}[t]
\centering
\begin{minipage}{\columnwidth}
\scriptsize
\begin{verbatim}
import matplotlib.pyplot as plt

fig, ax = plt.subplots()

# Raw
data = [348.0, 316.0, 136.0]
ax.pie(
    data,
    autopct="%1.1f%%",
    normalize=True,
)

# Observation-aligned
data = [43.5, 39.5, 17.0]
ax.pie(
    data,
    autopct="%1.1f%%",
    normalize=True,
)
\end{verbatim}
\end{minipage}
\caption{Python code for generating the raw and observation-aligned pie charts.}
\label{fig:pie-code}
\end{figure}

\begin{figure}[t]
\centering
\begin{minipage}{\columnwidth}
\scriptsize
\begin{verbatim}
import matplotlib.pyplot as plt

fig, ax = plt.subplots()

# Raw
data = [...]
bin_edges = [...]
ax.hist(
    data,
    bins=bin_edges,
)

# Observation-aligned
positions = [...]
bin_edges = [...]
weights = [...]
ax.hist(
    positions,
    bins=bin_edges,
    weights=weights,
)
\end{verbatim}
\end{minipage}
\caption{Python code for generating the raw and observation-aligned histograms.}
\label{fig:hist-code—normalizatio}
\end{figure}

\begin{figure}[t]
\centering
\begin{minipage}{\columnwidth}
\scriptsize
\begin{verbatim}
import matplotlib.pyplot as plt

fig, ax = plt.subplots()

# Raw
data = [...]
ax.boxplot(data)

# Observation-aligned
data = [
    {
        ...
    }
]
ax.bxp(data)
\end{verbatim}
\end{minipage}
\caption{Python code for generating the raw and observation-aligned box plots.}
\label{fig:boxplot-code}
\end{figure}

The code template for the controlled experiment is shown in Figure~\ref{fig:boxplot-code}, Figure~\ref{fig:pie-code} and Figure~\ref{fig:hist-code—normalizatio}

\subsubsection{Data Composition Analysis Experimental Results}
\label{sec:data_composition_analysis}

\begin{table}[t]
\centering
\small
\setlength{\tabcolsep}{3.8pt}
\renewcommand{\arraystretch}{1.08}
\begin{adjustbox}{max width=\columnwidth,center}
\begin{tabular}{llccc}
\toprule
Training set & Variant & Pie-F1 & Hist-Value & Box-Value \\
\midrule
\multirow{3}{*}{Hist/Pie/Box}
& Original & 88.70 & 55.50 & 85.02 \\
& Modified & 94.63 & \textbf{93.43} & 85.92 \\
& $\Delta$ & +5.94 & +37.93 & +0.90 \\
\midrule
\multirow{3}{*}{Hist/Pie/Box + Other}
& Original & 87.36 & 49.61 & 88.77 \\
& Modified & \textbf{95.11} & 90.40 & \textbf{89.93} \\
& $\Delta$ & +7.75 & +40.79 & +1.16 \\
\bottomrule
\end{tabular}
\end{adjustbox}
\caption{Effect of latent-observation rewriting with and without additional chart types. $\Delta$ denotes the absolute improvement from the original to the modified training data.}
\label{tab:latent_obs_with_other_types}
\end{table}

\begin{figure}[t]
    \centering
    \includegraphics[width=\columnwidth]{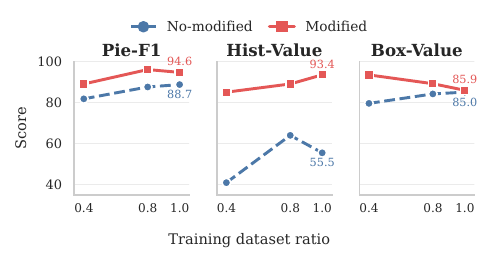}
    \caption{Performance trends of Qwen2.5-VL-7B under different training data ratios. Modified denotes the latent observation rewritten setting.}
    \label{fig:latent_obs_ratio}
\end{figure}
\begin{figure}[t]
    \centering
    \includegraphics[width=\columnwidth]{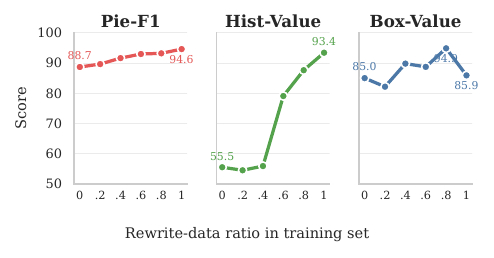}
    \caption{Performance trends of Qwen2.5-VL-7B under different ratios of rewritten training data.}
    \label{fig:rewrite_ratio}
\end{figure}
\paragraph{Performance gaps exists when trained with other Chart Types data}
To examine whether additional chart types reduce the effect of observation-aligned supervision, we mix the target training data with non-boxplot, non-histogram, and non-pie samples. Specifically, we randomly sample 50K training examples from ReChartPrompt-240K and 50K from ChartCoder, and add them to both the original and rewritten training sets. As shown in Table~\ref{tab:latent_obs_with_other_types}, the performance gap does not disappear after introducing other chart types. Models trained with observation-aligned supervision still achieve consistent gains, suggesting that the latent-observation mismatch cannot be fully mitigated by simply adding more general chart-to-code data.
\paragraph{Performance gaps exists under different training data amount}
We further vary the amount of original and rewritten training data to study how performance changes with data scale. The results are shown in Figure~\ref{fig:latent_obs_ratio}. Across different training sizes, observation-aligned supervision consistently outperforms raw-code supervision on data-recovery metrics. Notably, for boxplots and histograms, models trained with a smaller amount of rewritten data can already surpass models trained with the full original data.
\paragraph{Observation-Aligned Supervision Works on medium-to-high rewriting training data ratios}
We vary the ratio of observation-aligned samples mixed into the original training data. Results are shown in Figure~\ref{fig:rewrite_ratio}. When the rewritten ratio is low, performance can drop slightly, suggesting that a small amount of rewritten data may not be enough to change the learned behavior. As the rewriting ratio increases, the overall performance improves steadily. This trend shows that observation-aligned supervision becomes more effective when it accounts for a medium-to-large portion of the training data.

\subsubsection{Robustness Analysis}
\label{sec:robustness_analysis}

\begin{figure}[t]
    \centering
    \includegraphics[width=\columnwidth]{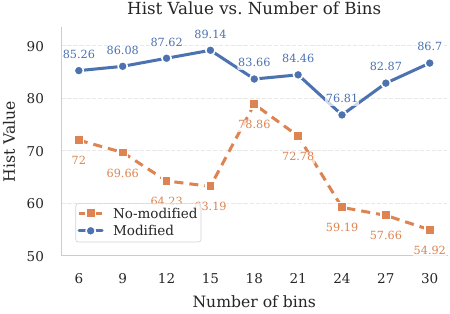}
    \caption{Hist Value trends under different histogram bin numbers.}
    \label{fig:hist_bins_trend}
\end{figure}

\paragraph{Observation-Aligned Supervision Improves Across Different Histogram Bin Counts}
To evaluate robustness to different histogram bins, we construct a set of simple histogram charts using a controlled template. The evaluation image-code samples are generated using the template code shown in Figure~\ref{fig:hist-code} filled with random data. For each bin count, we randomly generate 25 samples for evaluation.  As shown in Figure~\ref{fig:hist_bins_trend}, models trained with observation-aligned supervision consistently outperform the raw-code supervision baseline across all bin counts. In contrast, the baseline exhibits less stable performance as the bin count changes.

\paragraph{Observation-Aligned Supervision Improves on Pie Charts Without Percentage Annotations}

We remove percentage annotations from the pie charts constructed
using the controlled setup in Section~\ref{sec:cse} and compare raw-code
and observation-aligned supervision on the ChartX but remove percentage annotations.
Table~\ref{tab:unannotated_pie} reports Pie-F1 at different matching
tolerances $\tau$, measured in percentage points.
Observation-aligned supervision achieves higher scores at every
tested tolerance: Pie-F1 increases from 2.57 to 3.52 at $\tau=0.1$
and from 44.20 to 57.00 at $\tau=2.0$.
The benefit therefore persists without explicit percentage text,

\label{sec:histbincountexpdetail}

\begin{table}[t]
    \centering
    \small
    \begin{tabular}{@{}crr@{}}
        \toprule
        $\tau$ & Raw & Obs.-aligned \\
        \midrule
        0.1 & 2.57  & \textbf{3.52}  \\
        0.3 & 8.29  & \textbf{10.80} \\
        0.5 & 13.12 & \textbf{18.30} \\
        0.7 & 20.05 & \textbf{23.59} \\
        1.0 & 26.00 & \textbf{33.40} \\
        2.0 & 44.20 & \textbf{57.00} \\
        \bottomrule
    \end{tabular}
    \caption{Pie-F1 (\%) on controlled synthetic pie charts without
    percentage annotations. $\tau$ is the matching tolerance in
    percentage points. Bold marks the higher score at each tolerance.}
    \label{tab:unannotated_pie}
\end{table}

\begin{figure}[t]
\centering
\begin{minipage}{\columnwidth}
\scriptsize
\begin{verbatim}
from pathlib import Path
import matplotlib
matplotlib.use("Agg")
import matplotlib.pyplot as plt
import numpy as np

bin_count = 3
positions = (
    np.arange(bin_count, dtype=float)
    + 0.5
)
bin_edges = np.arange(
    bin_count + 1, dtype=float
)
heights = np.array(
    [4.724, 1.571, 0.236],
    dtype=float,
)

plt.figure(figsize=(6, 4), dpi=100)
plt.hist(
    positions,
    bins=bin_edges,
    weights=heights,
    color="lightblue",
)
plt.savefig(
    Path(__file__).with_suffix(".png"),
    bbox_inches="tight",
)
\end{verbatim}
\end{minipage}
\caption{Python code for generating the weighted histogram.}
\label{fig:hist-code}
\end{figure}

\subsection{More Experimental Details For Projection-induced Mismatch}
\label{sec:3d_metric_details}
We execute each predicted and gold program independently. Marker centers are
extracted from Matplotlib 3D scatter collections and projected to full-canvas
pixel coordinates using the chart's camera transform. Thus, all metrics use
the rendered observation and do not directly compare predicted XYZ or UV
values.

Let
$G=\{\mathbf{g}_i\}_{i=1}^{m}$ and
$P=\{\mathbf{p}_j\}_{j=1}^{n}$
denote the gold and predicted marker-center sets.

\paragraph{Marker F1.}
We perform one-to-one matching using Euclidean center distance. A pair is
correct if its distance is at most $\tau=5$ pixels. Let $T$ be the number of
correctly matched pairs. We compute
\begin{equation}
\begin{aligned}
    \mathrm{Precision} &= \frac{T}{n},
    &\mathrm{Recall} &= \frac{T}{m}, \\
    \mathrm{F1} &=
    \frac{2\,\mathrm{Precision}\,\mathrm{Recall}}
    {\mathrm{Precision}+\mathrm{Recall}}.
\end{aligned}
\label{eq:3d_marker_f1}
\end{equation}

\paragraph{OSPA-Center.}
We use OSPA with order $p=1$ and cutoff
$c=20$ pixels. Define the truncated center distance as
\begin{equation}
    \bar d_c(\mathbf{g},\mathbf{p})
    =
    \min\!\left(c,\lVert\mathbf{g}-\mathbf{p}\rVert_2\right).
\end{equation}
Assuming $m\leq n$, OSPA is
\begin{equation}
\begin{aligned}
d_{\mathrm{OSPA}}(G,P)
=
\frac{1}{n}\Bigg[
    &\min_{\pi\in\Pi_n}
    \sum_{i=1}^{m}
    \bar d_c
    \left(\mathbf{g}_i,\mathbf{p}_{\pi(i)}\right) \\
    &+c(n-m)
\Bigg],
\end{aligned}
\label{eq:3d_ospa}
\end{equation}
where $\Pi_n$ is the set of one-to-one assignments. If $m>n$, the two sets
are exchanged. The first term measures marker-center error, while the second
penalizes missing or extra markers. Lower values are better.

\paragraph{Crop-SSIM.}
We render both programs at the same resolution and compute SSIM between the
gold and predicted images using the same crop. The crop is defined by the gold
3D-axis region with four pixels of padding.

\paragraph{Failure handling and statistics.}
A failed prediction receives F1 $=0$, OSPA $=20$, and Crop-SSIM $=0$.
Metrics are first computed per chart and then averaged over the evaluation
set.

\subsection{More Experimental Details for Level-Set-Induced Mismatch}

\label{sec:app_countour_exp_details}

\subsubsection{Evaluation Metric Details}
\label{sec:contour_metric_details}

\paragraph{Execution and geometry extraction.}
Each gold and predicted program is executed independently in a subprocess
with the Matplotlib \texttt{Agg} backend. A prediction is executable if it
terminates successfully and leaves a renderable ordinary two-dimensional
figure. Execution measures program validity rather than contour accuracy:
an executable figure with no contour path receives execution credit but is
penalized by the geometry metrics.

Both raw and observation-aligned targets render a \texttt{ContourSet}. Raw targets obtain the artist
from \texttt{ax.contour}; observation-aligned targets construct it directly
from the ordered levels and their serialized polyline components. Geometry
is compared in display coordinates, independently of whether the program
contains a dense field, an analytic formula, or explicit contour paths.

Each contour artist's runtime transform maps its vertices into full-canvas
display pixels. Coordinates are converted to a common top-left origin and
clipped to the axes rectangle. 

Let $G=\{G_a\}_{a=1}^{A}$ and $P=\{P_b\}_{b=1}^{B}$ be the extracted gold
and predicted polyline components. Let $D$ be the diagonal length of the gold
axes rectangle. Unless stated otherwise, each polyline is resampled at
one-pixel arclength spacing. The resulting point sets are denoted by
$\widetilde{G}$ and $\widetilde{P}$.

\paragraph{Contour Boundary F1.}
For a normalized tolerance $\tau$, the matching threshold is $\tau D$, where
\begin{equation}
    \mathcal{T}=\{0.001,\,0.0025,\,0.005\},
    \qquad \tau\in\mathcal{T}.
\end{equation}
A bipartite graph joins gold and predicted points no farther apart than
this threshold. Maximum-cardinality matching credits each sampled point
at most once. If $M_\tau$ pairs are matched, then
\begin{equation}
\begin{aligned}
    \operatorname{Prec}_\tau
        &= \frac{M_\tau}{|\widetilde{P}|}, \\
    \operatorname{Rec}_\tau
        &= \frac{M_\tau}{|\widetilde{G}|}, \\
    \operatorname{BF1}_\tau
        &= \frac{2M_\tau}
        {|\widetilde{P}|+|\widetilde{G}|}.
\end{aligned}
\label{eq:contour_bf1}
\end{equation}
The last expression is the harmonic mean of precision and recall whenever
both point sets are non-empty. Shifted, missing, extra, and duplicated curves
are penalized. If exactly one set is empty, BF1 is zero; if both are empty,
the implementation returns one. Gold evaluation charts contain non-empty
geometry. The per-chart primary metric is
\begin{equation}
    \operatorname{mBF1}
    =\frac{1}{|\mathcal{T}|}
      \sum_{\tau\in\mathcal{T}}\operatorname{BF1}_\tau.
    \label{eq:mean_contour_bf1}
\end{equation}
Dataset values are macro averages of per-chart scores; boundary points from
different charts are never pooled.

\paragraph{Normalized geometric distances.}
For a point $p$ and a non-empty point set $S$, define
$d(p,S)=\min_{s\in S}\lVert p-s\rVert_2$. Write the directed mean distances
separately to keep the symmetric metric within one column:
\begin{equation}
\begin{aligned}
    \delta_{P\to G}
        &=\frac{1}{|\widetilde{P}|}
          \sum_{p\in\widetilde{P}}d(p,\widetilde{G}), \\
    \delta_{G\to P}
        &=\frac{1}{|\widetilde{G}|}
          \sum_{g\in\widetilde{G}}d(g,\widetilde{P}).
\end{aligned}
\end{equation}
The normalized symmetric Chamfer distance is
\begin{equation}
    \operatorname{Chamfer}(G,P)
    =\min\!\left\{
      \frac{\delta_{P\to G}+\delta_{G\to P}}{2D},\,1
    \right\}.
    \label{eq:contour_chamfer}
\end{equation}
For HD95, let $\mathcal{D}_{P\to G}$ and $\mathcal{D}_{G\to P}$ be the
directed nearest-neighbor distance sets:
\begin{equation}
\begin{aligned}
    \mathcal{D}_{P\to G}
        &=\{d(p,\widetilde{G}):p\in\widetilde{P}\}, \\
    \mathcal{D}_{G\to P}
        &=\{d(g,\widetilde{P}):g\in\widetilde{G}\}.
\end{aligned}
\end{equation}
Their union is denoted by $\mathcal{D}$:
\begin{equation}
    \mathcal{D}=\mathcal{D}_{P\to G}\cup\mathcal{D}_{G\to P}.
\end{equation}
With $Q_{0.95}$ denoting the empirical 95th percentile,
\begin{equation}
    \operatorname{HD95}(G,P)
    =\min\!\left\{\frac{Q_{0.95}(\mathcal{D})}{D},\,1\right\}.
    \label{eq:contour_hd95}
\end{equation}
Both metrics are lower-is-better. An empty prediction against non-empty gold
receives one. Chamfer summarizes average displacement, whereas HD95 captures
locally severe errors without being dominated by a single outlier.

\paragraph{Component and topology metrics.}
Each continuous polyline is one component. For each gold--predicted
component pair, the evaluator computes bidirectional tolerance-coverage F1
at $0.0025D$. A pair is eligible when its overlap is at least $0.80$.
Eligible pairs are assigned one-to-one, maximizing the number of valid
matches first and their total overlap second. If $C$ components are matched,
\begin{equation}
\begin{aligned}
    \operatorname{CompPrec} &=\frac{C}{B}, \\
    \operatorname{CompRec}  &=\frac{C}{A}, \\
    \operatorname{CompF1}   &=\frac{2C}{A+B}, \\
    \operatorname{CountErr}&=\frac{|B-A|}{\max(A,B,1)}.
\end{aligned}
\label{eq:contour_component_metrics}
\end{equation}
When one component set is empty and the other is not, component precision,
recall, and F1 are defined as zero rather than evaluated by division.

A component is closed if its first and last display-space vertices are
within $10^{-6}$ pixels. Let $c(H)$ be one for a closed component $H$ and zero
otherwise. For the matched pair set $\mathcal{A}$, closure accuracy is
\begin{equation}
    \operatorname{ClosureAcc}
    =\frac{1}{C}\sum_{(a,b)\in\mathcal{A}}
      \mathbf{1}\!\left[c(G_a)=c(P_b)\right].
    \label{eq:contour_closure}
\end{equation}
It is zero when no components match. Component F1 and closure accuracy are
higher-is-better; component-count error is lower-is-better. Keeping the same
non-inline label policy on both targets prevents label gaps from creating
artificial topology differences.

\paragraph{Crop-SSIM}
Gold and predicted programs are rendered at their native canvas resolution.
Both RGB renders are cropped at the gold axes rectangle with four pixels
of padding. Local SSIM uses a $7\times7$ reflected window independently over
the RGB channels and averages all local values. The constants are
$C_1=(0.01\cdot255)^2$ and $C_2=(0.03\cdot255)^2$.

\paragraph{Failure handling and paired aggregation.}
Missing or non-executable predictions receive zero for higher-is-better
metrics and one for normalized distances and error metrics. If metric
computation times out after successful execution, the same conservative
comparison penalties apply, while execution remains one. A gold execution
failure aborts evaluation rather than changing the reference silently.

The all-sample result averages over every gold chart and is the primary
end-to-end evaluation. Let $e_i^{\mathrm{raw}}$ and $e_i^{\mathrm{obs}}$ be
the paired execution indicators. The diagnostic both-executable subset is
\begin{equation}
    \mathcal{I}_{\mathrm{both}}
    =\{i:e_i^{\mathrm{raw}}=e_i^{\mathrm{obs}}=1\}.
    \label{eq:contour_both_exec}
\end{equation}
Both systems are recomputed on the same selected chart IDs. This conditional
comparison isolates geometric fidelity from executability, but does not
replace the all-sample result because selection occurs after generation.
Paired bootstrap resampling uses chart ID as the sampling unit and applies
the same indices to both systems. The comparison protocol uses $10{,}000$
replicates for the dense-field setting and $20{,}000$ for the analytic
setting, with percentile 95\% confidence intervals for paired mean differences.

\clearpage
\subsection{Editable Programs from Observation-Aligned Representations}

\label{app:obs-editable-programs}

Observation-aligned representations can also serve as inputs to a deterministic
compiler that produces editable representative data. The model still predicts
only observed quantities; the compiler runs afterward. The examples below
illustrate this optional extension, rather than an additional training experiment.
Both columns use the same observed inputs. Shared plot settings and long
coordinate lists are omitted;

\paragraph{Boxplots.}
The compiler constructs representative samples, then calls \texttt{boxplot}
instead of \texttt{bxp} (Figure~\ref{fig:app-box-editable}). It first checks
that the whiskers lie within the 1.5-IQR fences and the outliers lie outside.
Let $L$ and $U$ be the numbers of lower and upper outliers, and choose
$k=\max(L,U)+1$. Build a sorted array in this order: the $L$ lower outliers,
$k-L$ copies of the lower whisker, $k$ copies of the first quartile, $k$ copies
of the median, $k-U$ copies of the third quartile, one upper whisker, and the
$U$ upper outliers. This gives $4k+1$ samples. With linear quartiles, the
quartile indices are exactly $k$, $2k$, and $3k$ (counting from zero), so they
contain the requested values. The whiskers and outliers also match. Users may
then edit samples or change the whisker rule; mean markers and notches need
additional constraints.

\begin{figure*}[t]
\centering
\begin{minipage}[t]{0.48\textwidth}
\vspace{0pt}
\textbf{Observation-aligned}\par
\scriptsize
\begin{verbatim}
# Predict the visible summary statistics.
stats = {
    "label": "A",
    "q1": 7.70, "med": 9.00,
    "q3": 10.98,
    "whislo": 6.20,
    "whishi": 11.50,
    "fliers": [18.0],
}

fig, ax = plt.subplots()
ax.bxp(
    [stats], widths=0.5,
    showmeans=False,
    shownotches=False,
)
\end{verbatim}
\end{minipage}
\hfill
\begin{minipage}[t]{0.48\textwidth}
\vspace{0pt}
\textbf{Observation-aligned + editable}\par
\scriptsize
\begin{verbatim}
from compilers import compile_box

# Same predicted stats as on the left.
samples = compile_box(stats, whis=1.5)
# Example output:
# [6.20, 6.20, 7.70, 7.70, 9.00,
#  9.00, 10.98, 11.50, 18.00]

# Optional: edit samples before plotting.
fig, ax = plt.subplots()
ax.boxplot(
    samples, whis=1.5, widths=0.5,
    tick_labels=[stats["label"]],
    autorange=False,
    showmeans=False, notch=False,
)
\end{verbatim}
\end{minipage}
\caption{Boxplot code pair. The compiler fixes the quartile positions in a
sorted sample array and retains the specified whisker endpoints and outliers.
The initial displayed statistics are unchanged.}
\label{fig:app-box-editable}
\end{figure*}

\paragraph{Histograms.}
The paper already uses \texttt{hist} with explicit bins and weights. For a bin
with edges $b_i,b_{i+1}$ and integer count $n_i$, divide the interval into
$n_i$ equal parts and place one sample at each midpoint:
\[
    x_{ij}=b_i+\frac{j+1/2}{n_i}(b_{i+1}-b_i).
\]
Here $j=0,\ldots,n_i-1$; empty bins contribute no samples. Concatenate these
samples across bins and call \texttt{hist} without weights
(Figure~\ref{fig:app-hist-editable}). All samples lie strictly inside their bins,
so the initial counts match exactly. Users may edit samples or rebin them;
new bins reflect this chosen within-bin arrangement. Noninteger masses should
retain a weighted representation.

\begin{figure*}[t]
\centering
\begin{minipage}[t]{0.48\textwidth}
\vspace{0pt}
\textbf{Observation-aligned}\par
\scriptsize
\begin{verbatim}
# Predict bin-level counts and edges.
weights = np.array([4, 5, 3, 5])
bins = np.linspace(
    0.23, 3.97, len(weights) + 1
)
centers = (bins[:-1] + bins[1:]) / 2

fig, ax = plt.subplots()
ax.hist(
    centers, bins=bins,
    weights=weights,
)
\end{verbatim}
\end{minipage}
\hfill
\begin{minipage}[t]{0.48\textwidth}
\vspace{0pt}
\textbf{Observation-aligned + editable}\par
\scriptsize
\begin{verbatim}
from compilers import compile_hist

# Same predicted bins and weights.
samples = compile_hist(bins, weights)

# Optional: edit samples or change bins.
# bins = np.linspace(0.23, 3.97, 9)

fig, ax = plt.subplots()
ax.hist(
    samples, bins=bins,
)
\end{verbatim}
\end{minipage}
\caption{Histogram code pair. Weighted bin representatives are expanded into
unweighted representative samples. The original bin edges and counts are
preserved before editing.}
\label{fig:app-hist-editable}
\end{figure*}

\paragraph{Contours.}
We retain the paper's \texttt{LEVELS}/\texttt{ALLSEGS} representation,
\texttt{ContourSet}, and labels. The compiler checks that each level has one
simple closed curve, the curves are strictly nested, and levels increase
outward. It then fills a grid from these paths (Figure~\ref{fig:app-contour-editable}).
For each grid point $p$, compute its shortest distance to each curve's line
segments, and make the distance negative inside the curve and positive outside.
For adjacent inner and outer curves with levels $a<b$, let these signed
distances be $s_{\mathrm{in}}$ and $s_{\mathrm{out}}$. Set
\[
    Z(p)=a+(b-a)
    \frac{s_{\mathrm{in}}(p)}
         {s_{\mathrm{in}}(p)-s_{\mathrm{out}}(p)}.
\]
Between the curves, this is ordinary distance-weighted interpolation: it gives
$a$ on the inner curve, $b$ on the outer curve, and $(a+b)/2$ halfway between
them in distance. For multiple nested levels, use the two curves bounding each
region. Inside the innermost curve use the first pair; outside the outermost
curve use the last pair. The same formula then gives values below or above the
observed range. Finally, call \texttt{contour} on the resulting grid.
Users may edit $Z$ or add levels. Open curves and multiple components are
unsupported, and finite-grid rendering is approximate. These compilers select
representative data; they do not recover the original samples or formula.

\begin{figure*}[t]
\centering
\begin{minipage}[t]{0.48\textwidth}
\vspace{0pt}
\textbf{Observation-aligned}\par
\scriptsize
\begin{verbatim}
from matplotlib.contour import ContourSet

LEVELS = np.array([0.5, 1.0])
# Each path contains observed vertices.
# Use the full coordinates in Figure 14.
ALLSEGS = [
    [inner_path],  # LEVELS[0]
    [outer_path],  # LEVELS[1]
]
LABEL_TEXT = {0.5: "0.5", 1.0: "1.0"}
LABEL_POSITIONS = [(-.80, 0.), (-1.15, 0.)]

fig, ax = plt.subplots()
# Set shared limits and layout first.
cs = ContourSet(
    ax, LEVELS, ALLSEGS,
    filled=False, colors="#202020",
    linewidths=1.25, linestyles="solid",
)
fig.canvas.draw()
ax.clabel(
    cs, levels=LEVELS, fmt=LABEL_TEXT,
    manual=LABEL_POSITIONS,
    inline=False, fontsize=8,
)
\end{verbatim}
\end{minipage}
\hfill
\begin{minipage}[t]{0.48\textwidth}
\vspace{0pt}
\textbf{Observation-aligned + editable}\par
\scriptsize
\begin{verbatim}
from compilers import compile_nested_field

# Same LEVELS, ALLSEGS, and labels.
grid = np.linspace(-1.5, 1.5, 601)
X, Y = np.meshgrid(grid, grid)
Z = compile_nested_field(
    LEVELS, ALLSEGS, X, Y
)

# Optional: edit Z or add contour levels.
plot_levels = LEVELS.copy()
fig, ax = plt.subplots()
# Set the same limits and layout first.
cs = ax.contour(
    X, Y, Z, levels=plot_levels,
    colors="#202020", linewidths=1.25,
    linestyles="solid",
)
fig.canvas.draw()
ax.clabel(
    cs, levels=LEVELS, fmt=LABEL_TEXT,
    manual=LABEL_POSITIONS,
    inline=False, fontsize=8,
)
\end{verbatim}
\end{minipage}
\caption{Contour code pair using the paper's level-grouped paths and labels.
The right column reconstructs a representative field from geometry alone.
The supplied curves are approximated on a grid; additional levels describe
the reconstructed field. Full path coordinates and shared styling are omitted.}
\label{fig:app-contour-editable}
\end{figure*}

\subsection{Case Study}
\label{sec:case_study}
\begin{figure*}[t]
  \includegraphics[width=\textwidth]{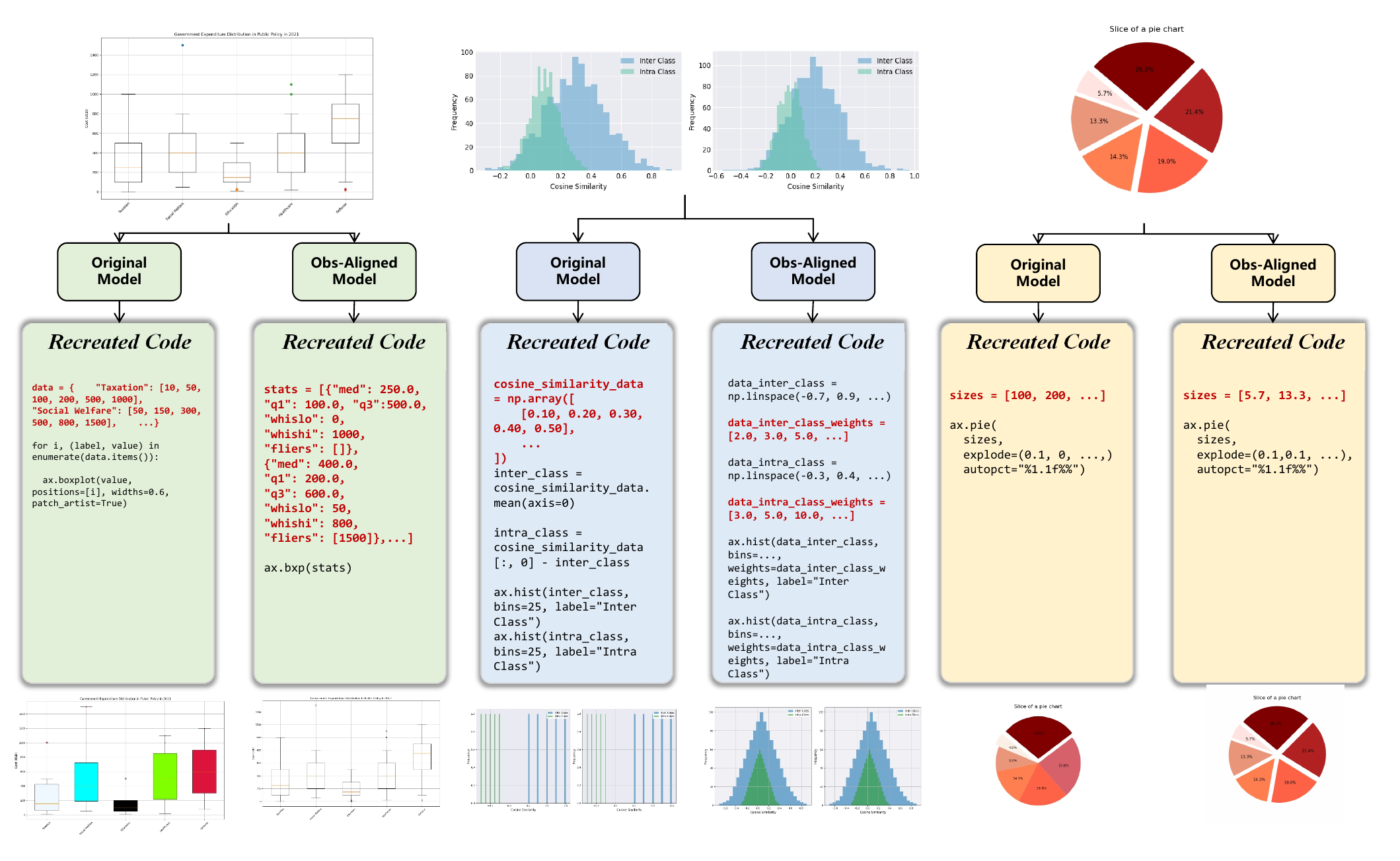}
    \caption{Qualitative Cases Of Observation-Aligned Supervision for aggregation-induced mismatch and normalization-induced mismatch}
  \label{fig:qc}
\end{figure*}
\begin{figure*}[t]
  \includegraphics[width=\textwidth]{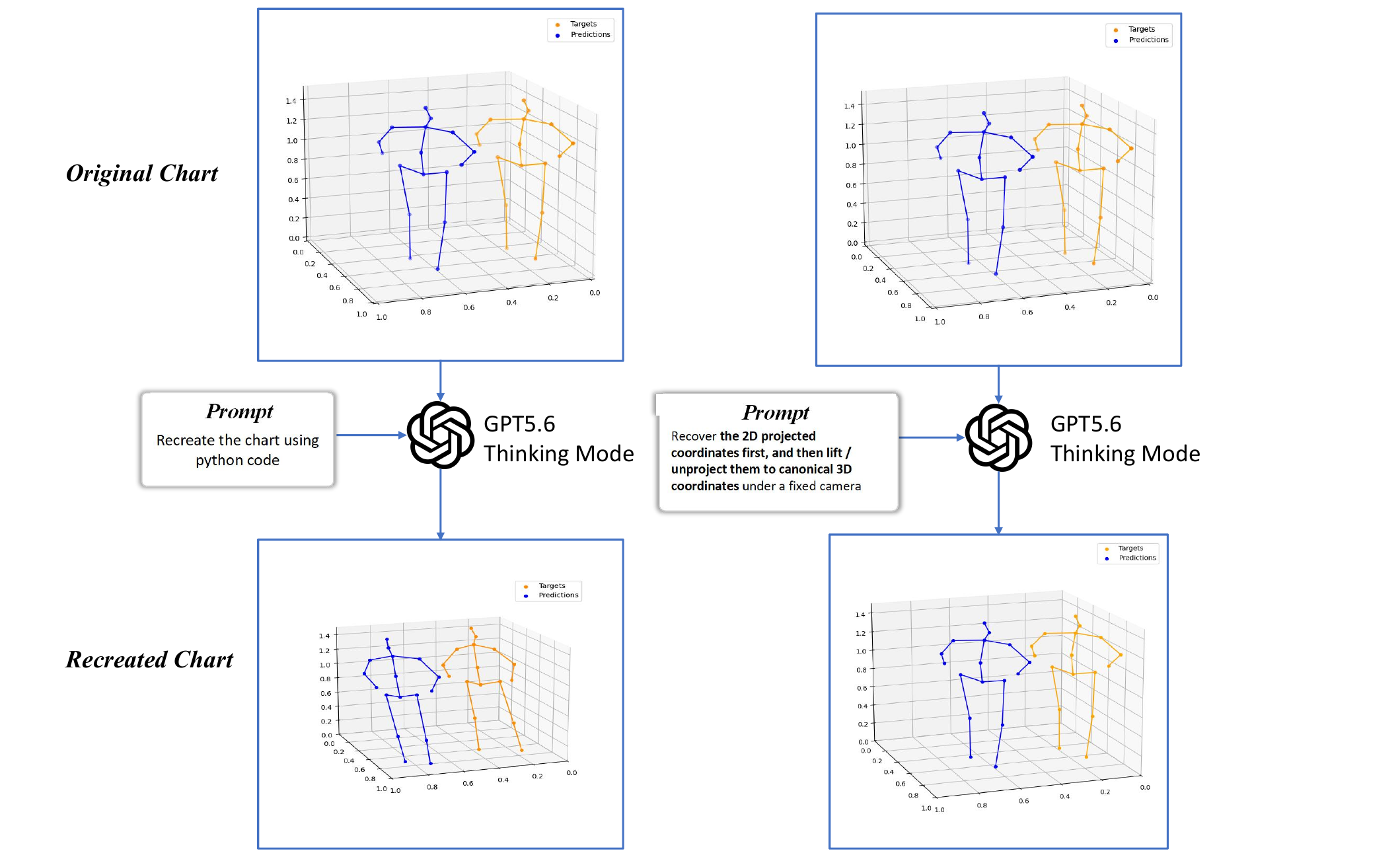}
    \caption{Prompt-level qualitative illustration of projection-induced latent–observation mismatch. Given the same fixed-view 3D chart and GPT5.6 Thinking mode, direct chart-to-code prompting requires the model to infer latent XYZ coordinates and produces a less faithful projected reconstruction. In contrast, the observation-aligned prompt first recovers normalized 2D vertex locations and connectivity, after which a deterministic compiler lifts them to canonical 3D coordinates under the fixed camera. The latter better preserves the observable projected geometry.}
  \label{fig:qc_3d}
\end{figure*}
Figure~\ref{fig:qc} presents three representative examples.  For the box plot, the non-modified model tries to recover raw data lists.  In contrast, the modified model directly predicts the visible box statistics. As a result, it better matches the key visual structure of the chart.
For the histogram, the non-modified model tends to generate a small set of pseudo raw samples and use them with histogram bins.  The modified model instead represents the histogram with explicit bin positions and weights. This leads to much better recovery of the bar distribution across the two subplots.
For the pie chart, the non-modified model predicts arbitrary raw sizes that only roughly preserve the slice size.  Although this can produce a similar overall pie shape, it often fails to match the displayed percentages.  The modified model using the visible percentages as the direct supervision target can directly recover the percentage values.

Figure~\ref{fig:qc_3d} presents
a prompt-level qualitative illustration of projection-induced latent–observation mismatch. Given the same fixed-view 3D chart and GPT5.6 Thinking mode, direct chart-to-code prompting requires the model to infer latent XYZ coordinates and produces a less faithful projected reconstruction. In contrast, the observation-aligned prompt first recovers normalized 2D vertex locations and connectivity, after which a deterministic compiler lifts them to canonical 3D coordinates under the fixed camera. The latter better preserves the observable projected geometry.

\newpage
\clearpage

\subsection{Reconsidering evaluation metrics for Chart2Code}
\label{sec:app_receval}
The same observability principle also has implications for evaluating visual-production code. In particular, an evaluator should distinguish between properties that are identifiable from the rendered chart and properties that belong only to one particular reference implementation.

 First,
 for the chart types considered in this work, evaluation should emphasize observable quantities rather than requiring exact recovery of latent authoring variables that are not uniquely determined by the image. Otherwise, an observationally valid program may be penalized simply because it selects a different member of the same rendering-equivalence class. 
 
 Second, \emph{the decomposition of visible graphics into plotting operations may itself be non-unique}. Stacked area plots provide a representative example. Even when implemented entirely with \texttt{fill\_between}, the visible stacked regions need not correspond one-to-one with individual filling calls. One implementation may fill only between adjacent cumulative boundaries, while another may first fill a broader region spanning, for example, the first and third boundaries, and then overlay another region spanning the second and fourth boundaries. Through layering and occlusion, these different hierarchies of fills can produce the same final visible regions, even though the colors associated with individual \texttt{fill\_between} calls differ substantially. Consequently, evaluating color by directly matching the properties of corresponding plotting calls can be misleading: the image constrains the color of the final visible region, but not necessarily the color assigned to every intermediate or occluded artist in the program. More generally, visual-production evaluation should be invariant to implementation differences that disappear after rendering. 
 
 Third, \emph{finite raster resolution limits the numerical precision that can be inferred from an image}. If a plotting area contains only a limited number of pixels while the corresponding axis spans a large numerical range, multiple nearby data values may map to the same, or nearly the same, pixel location. In such cases, requiring exact or excessively precise agreement in data coordinates demands information that is not present in the observation. Numerical evaluation should therefore account for the precision supported by the rendered image, for example through display-space error or resolution-aware tolerances. 
 
 Together, these cases suggest a broader principle: for visual-production chart-to-code generation, evaluation, like supervision, should be aligned with the rendered observation. A reference program specifies one valid process for producing the chart, but program-level quantities or implementation choices that are not identifiable from the image should not by themselves determine visual correctness.

\subsection{Prompts}
\label{sec:prompts}
We show the prompt templates for rewriting the Pie, Box and Hist data in prompt~\ref{box:rewriting_prompt_pie},
~\ref{box:rewriting_prompt_box}
,~\ref{box:rewriting_prompt_hist} respectively.

\begin{promptbox}[title={Rewriting prompt for Pie Chart},label={box:rewriting_prompt_pie}]
\begin{quote}
\small
You are an assistant that rewrites code with minimal changes only.

Task: Minimally rewrite the following Matplotlib code and output only the complete rewritten Python code.

Hard requirements:
1. Strictly follow the precomputed target for each pie call, but the rule "preserve raw data when raw data is visible" has higher priority.
2. Do not rewrite the entire code. Only modify necessary data definitions and the downstream dependency chain that passes these data to pie calls.
3. Prefer modifying upstream data sources. Do not rewrite the whole loop, plotting logic, or subplot logic.
4. Do not compute percentages again in the final code. Do not add runtime code such as total = sum(...), percentages = [...], or f"{...:.1f}%".
5. Preserve the original variable names, control flow, loops, style parameters, titles, layout, and saving logic, unless changes are necessary for the task.
6. If a file contains multiple pie calls, process each call according to its own target. Do not merge multiple calls into one rewrite.
7. Output only the final complete Python code. Do not provide explanations or diffs.

Rule for preserving visible raw data (highest priority):
1. If the original code directly exposes raw data in visible text on the figure, do not rewrite the data passed to pie into percentages.
2. "Visible text directly exposes raw data" includes but is not limited to:
   - labels directly concatenate raw values, e.g., f"{name}: {val}", str(v), "{:.2f}".format(v)
   - legend text contains raw values
   - title, suptitle, text, annotate, table, figtext, or other visible text contains raw values
   - any other case where users can directly see the original numeric values in the final figure
3. Once raw data is visible:
   - preserve the original x / values / sizes / data inputs
   - do not replace them with target_values_literal
   - do not add autotext overwriting just to align percentages
   - unless the original code already contains autopct or autotext-related logic, do not add it
4. Even if percentages are also shown, as long as raw data is visible, preserving raw data still takes priority.

Standard rewriting rules:
1. If rewrite_mode = visible_pct_text and no raw data is visible in the original figure:
   - make the data passed to pie directly become target_values_literal; preferably modify the variable value passed to pie directly
   - target_values_literal has already been prepared according to the true percentage text shown in the figure; preserve the number of decimal places in the displayed text
   - explicitly write target_pct_texts_literal into the code
   - make the pie call capture autotexts
   - after the pie call returns, minimally overwrite each autotext by set_text using these explicit strings
2. If rewrite_mode = normalized_pct_1dp and no raw data is visible in the original figure:
   - make the data passed to pie directly become target_values_literal
   - do not add autopct
   - do not add any runtime percentage computation
3. If a pie call does not need to be modified, keep it unchanged.

Minimal-change constraints:
1. Avoid defining new variables whenever possible.
2. Prefer directly modifying the original data variable or its upstream data source, rather than creating intermediate variables such as percentages / new_values / rewritten_values.
3. Add new variables only when truly necessary.
4. For visible_pct_text, adding a "percentage text list" variable is allowed; minimally modifying the pie return values to capture autotexts is also allowed.
5. Except for the cases above, do not introduce extra intermediate variables merely for rewriting convenience.
6. The goal is the minimal diff: only modify necessary data and the dependency chain that passes these data to pie calls.

Especially do not modify:
1. labels, colors, explode, startangle, radius, counterclock, wedgeprops, textprops, pctdistance, labeldistance, shadow, frame, rotatelabels
2. subplot layout, figure size, titles, legend, axis, tight_layout, savefig, show
3. existing fonts, colors, linewidths, positions, spacing, loop structures, variable names, and data organization
4. do not reorder statements, split one line into many lines, or compress many lines into another structure, unless absolutely necessary for completing the rewrite

You must first study the following examples. The key point of the examples is not the specific numbers, but the rewriting style: minimal changes, prefer modifying upstream data, preserve raw data when visible, and only perform the minimal necessary autotext overwrite for visible_pct_text.

================ ICL EXAMPLES ================

Example 1: raw data is already written into labels, so raw data is visible and the original inputs are preserved.

Input code:
import matplotlib.pyplot as plt

metrics = ['Entropy', 'Complexity']
values = [0.5, 0.8, 1.2, 0.7, 1.1, 1.5]
colors = ['#FFE4C4', '#006400', '#E9967A']
fig, axs = plt.subplots(1, 2, figsize=(10, 6), subplot_kw=dict(aspect='equal'))

axs[0].pie(
    values[:3],
    labels=[f'{metrics[0]}: {val}' for val in values[:3]],
    startangle=90,
    counterclock=False,
    colors=colors,
    wedgeprops=dict(width=0.3)
)
axs[1].pie(
    values[3:6],
    labels=[f'{metrics[1]}: {val}' for val in values[3:6]],
    startangle=90,
    counterclock=False,
    colors=colors,
    wedgeprops=dict(width=0.3)
)

Given targets:
[pie #0]
- rewrite_mode: normalized_pct_1dp
- target_values_literal: [20.0, 32.0, 48.0]

[pie #1]
- rewrite_mode: normalized_pct_1dp
- target_values_literal: [21.2, 33.3, 45.5]

Output code:
import matplotlib.pyplot as plt

metrics = ['Entropy', 'Complexity']
values = [0.5, 0.8, 1.2, 0.7, 1.1, 1.5]
colors = ['#FFE4C4', '#006400', '#E9967A']
fig, axs = plt.subplots(1, 2, figsize=(10, 6), subplot_kw=dict(aspect='equal'))

axs[0].pie(
    values[:3],
    labels=[f'{metrics[0]}: {val}' for val in values[:3]],
    startangle=90,
    counterclock=False,
    colors=colors,
    wedgeprops=dict(width=0.3)
)
axs[1].pie(
    values[3:6],
    labels=[f'{metrics[1]}: {val}' for val in values[3:6]],
    startangle=90,
    counterclock=False,
    colors=colors,
    wedgeprops=dict(width=0.3)
)

Explanation:
- labels directly show raw values, so raw data is visible.
- Even though percentage targets are given, do not rewrite values into percentages.
- The minimal change is to make no change.

------------------------------------------------

Example 2: only percentage text is visible, with no visible raw data. Rewrite according to the true displayed percentages and explicitly overwrite autotext.

Input code:
import matplotlib.pyplot as plt

values = [12, 54, 23]
labels = ['A', 'B', 'C']
colors = ['red', 'blue', 'green']

fig, ax = plt.subplots()
ax.pie(
    values,
    labels=labels,
    colors=colors,
    autopct='%1.1f%%',
    startangle=90
)

Given target:
[pie #0]
- rewrite_mode: visible_pct_text
- target_values_literal: [13.5, 60.7, 25.8]
- target_pct_texts_literal: ['13.5%', '60.7%', '25.8%']

Output code:
import matplotlib.pyplot as plt

values = [13.5, 60.7, 25.8]
labels = ['A', 'B', 'C']
colors = ['red', 'blue', 'green']
display_pcts = ['13.5%', '60.7%', '25.8%']

fig, ax = plt.subplots()
wedges, texts, autotexts = ax.pie(
    values,
    labels=labels,
    colors=colors,
    autopct='%1.1f%%',
    startangle=90
)
for t, s in zip(autotexts, display_pcts):
    t.set_text(s)

Explanation:
- No raw value is visible in the figure; only percentage text is visible.
- Therefore, directly rewrite values into target_values_literal.
- Only perform the minimal necessary autotext overwrite; do not change any other parameters.

------------------------------------------------

Example 3: neither visible raw data nor displayed percentage text that needs to be preserved exists. Only rewrite the input into one-decimal percentages and do not add autopct.

Input code:
import matplotlib.pyplot as plt

labels = ['Basketball', 'Table tennis', 'Badminton', 'Ice hockey']
sizes = [150, 120, 100, 130]
colors = ['#000000', '#0000FF', '#006400', '#00FFFF']

fig, ax = plt.subplots(figsize=(8, 8))
ax.pie(
    sizes,
    labels=labels,
    colors=colors,
    startangle=90,
    wedgeprops=dict(edgecolor='w')
)
plt.savefig('1.png')

Given target:
[pie #0]
- rewrite_mode: normalized_pct_1dp
- target_values_literal: [30.0, 24.0, 20.0, 26.0]

Output code:
import matplotlib.pyplot as plt

labels = ['Basketball', 'Table tennis', 'Badminton', 'Ice hockey']
sizes = [30.0, 24.0, 20.0, 26.0]
colors = ['#000000', '#0000FF', '#006400', '#00FFFF']

fig, ax = plt.subplots(figsize=(8, 8))
ax.pie(
    sizes,
    labels=labels,
    colors=colors,
    startangle=90,
    wedgeprops=dict(edgecolor='w')
)
plt.savefig('1.png')

Explanation:
- No raw value is visible.
- No additional displayed percentage text needs to be preserved.
- Therefore, only rewrite the input data; do not add autopct or runtime percentage computation.

------------------------------------------------

Example 4: multiple pie calls in the same file are processed separately; one preserves raw data, and one is rewritten with visible_pct_text.

Input code:
import matplotlib.pyplot as plt

left_values = [10, 20, 30]
right_values = [3, 7, 10]
labels = ['A', 'B', 'C']

fig, axs = plt.subplots(1, 2, figsize=(8, 4))

axs[0].pie(
    left_values,
    labels=[f'raw={v}' for v in left_values],
    startangle=90
)

axs[1].pie(
    right_values,
    labels=labels,
    autopct='%1.0f%%',
    startangle=90
)

Given targets:
[pie #0]
- rewrite_mode: normalized_pct_1dp
- target_values_literal: [16.7, 33.3, 50.0]

[pie #1]
- rewrite_mode: visible_pct_text
- target_values_literal: [15, 35, 50]
- target_pct_texts_literal: ['15%', '35%', '50%']

Output code:
import matplotlib.pyplot as plt

left_values = [10, 20, 30]
right_values = [15, 35, 50]
labels = ['A', 'B', 'C']
right_display_pcts = ['15%', '35%', '50%']

fig, axs = plt.subplots(1, 2, figsize=(8, 4))

axs[0].pie(
    left_values,
    labels=[f'raw={v}' for v in left_values],
    startangle=90
)

wedges, texts, autotexts = axs[1].pie(
    right_values,
    labels=labels,
    autopct='%1.0f%%',
    startangle=90
)
for t, s in zip(autotexts, right_display_pcts):
    t.set_text(s)

Explanation:
- The first pie directly shows raw values in labels, so raw data is preserved.
- The second pie has no visible raw value but has visible percentage text, so it is rewritten according to visible_pct_text.
- Multiple calls must be processed separately and must not be merged.

================ END OF ICL EXAMPLES ================

Now process the real task.

You must think and execute in the following order:
1. First check whether visible text corresponding to each pie call directly exposes raw data.
2. If raw data is exposed, keep the original input for that call and do not rewrite it into percentages.
3. Otherwise, process the call according to its rewrite_mode:
   - visible_pct_text -> rewrite input + minimally overwrite autotexts
   - normalized_pct_1dp -> rewrite input only
4. Judge and process multiple calls separately.
5. Finally, output only the complete rewritten Python code.

File:
{{SOURCE_FILE}}

Precomputed targets for each pie call:
{{CALL_TARGETS}}

Original code:
{{PY_CODE}}
\end{quote}
\end{promptbox}

\begin{promptbox}[title={Rewriting prompt for Box Plot},label={box:rewriting_prompt_box}]
\begin{quote}
\small
You are an assistant that rewrites code with minimal changes only.

First study the following examples. The key point of the examples is not the specific numbers, but the rewriting style:
- Directly modify the original box-data variable.
- Rewrite .boxplot(...) into .bxp(...) in place.
- Avoid introducing new variables whenever possible.
- Do not refactor loops or subplot structures.
- Do not aggregate multiple calls into a unified bxpstats_list / kwargs_list / *_all style.
- Do not extract parameters that can be directly written inside .bxp(...), such as positions, widths, patch_artist, boxprops, medianprops, whiskerprops, and capprops, into extra kwargs / kwargs_list variables.
- Preserve the original inline-argument style of the boxplot call whenever possible. If the original parameters are written directly inside boxplot(...), prefer writing them directly inside bxp(...) after rewriting.
- For simple loops, prefer rewriting ax.boxplot(values, ...) in place into ax.bxp(values, ...). Do not rewrite it into forms such as ax.bxp([data[i]], **kwargs_list[i]) with extra abstraction.
- Do not casually fix issues unrelated to boxplot -> bxp unless the code cannot run without the fix.
- When "runtime validity + visual equivalence" conflicts with "passing fewer parameters", prioritize the former.
- Pay special attention to the three conditional statistics: flier, notch, and mean:
  - Every stats dict in the final code must contain a fliers field.
  - If the target does not provide fliers for a box, uniformly add 'fliers': [].
  - If visible fliers are needed, preserve the real fliers given in the target.
  - If notch is needed, preserve cilo/cihi and explicitly write shownotches=True.
  - If mean is needed, preserve mean and explicitly write showmeans=True.
- If the same box data is reused by multiple boxplot calls, rewrite the original variable only once and reuse it in multiple .bxp(...) calls.
- If the box data is not an independent variable but an immediate expression constructed inside a loop, then it is allowed to add a bxpstats variable immediately next to the call. However, do not construct it inside the loop via temporary if/elif branches over i or key.
- If the original data container carries both box data and other non-box semantics, such as week, name, or x-axis labels, only replace the part that is actually passed into boxplot. Do not refactor the whole container into another templated structure.
- If the rewritten code still contains local expressions that depend on the original samples, such as max(values), min(values), or np.max(values), only make local, nearby, minimal compatibility patches. Do not refactor the whole code because of this.
- If an original data container is still used by code such as pd.DataFrame(data), which depends on equal-length columns, do not directly replace column values inside that container with bxpstats. Otherwise, the container structure may be broken and cause runtime failure. In this case, it is allowed to add a nearby bxpstats variable that only serves the current boxplot call.
- Since every stats dict in the final code uniformly contains a fliers field, usually there is no need to additionally write showfliers=False, unless target_bxp_kwargs_literal explicitly requires preserving that parameter.

Additional rule, very important, specifically to prevent incorrect generation of plt.bxp:
- You must distinguish the original receiver of the boxplot call: ax.boxplot(...), axs[i].boxplot(...), inset.boxplot(...), and plt.boxplot(...) are not the same.
- bxp is an Axes method. Prefer preserving the original receiver:
  - ax.boxplot(...) -> ax.bxp(...)
  - axs[i].boxplot(...) -> axs[i].bxp(...)
  - inset.boxplot(...) -> inset.bxp(...)
  - some_axes.boxplot(...) -> some_axes.bxp(...)
- If the original code is plt.boxplot(...), do not directly generate plt.bxp(...).
  The correct approach is to minimally add a current-axes object near that call and then call ax.bxp(...), for example:
  - ax = plt.gca()
  - box = ax.bxp(...)
- If the original code uses current-axes style such as plt.subplot(...) / plt.sca(...), do not refactor the whole code into fig, axs = plt.subplots(...) merely to call bxp.
  The correct approach is usually to write ax = plt.gca() nearby, and then call ax.bxp(...).
- If the original boxplot return value is assigned to a variable, such as box = plt.boxplot(...), the rewritten version should also preserve this return-value assignment as much as possible, such as box = ax.bxp(...). This allows downstream logic such as box['boxes'] and box['medians'] to keep working.
- Do not rewrite still-valid pyplot-style calls such as plt.title, plt.xticks, plt.ylabel, plt.ylim, plt.legend, and plt.tight_layout into ax.set_* merely because ax.bxp(...) is needed, unless not changing them would make the code fail.
- Do not unnecessarily rewrite plt.figure(...) + plt.boxplot(...) style into fig, ax = plt.subplots(...). Prefer the smaller compatibility patch: ax = plt.gca().

Common errors, strictly forbidden:
Incorrect:
box = plt.bxp(data, patch_artist=True)

Correct:
ax = plt.gca()
box = ax.bxp(data, patch_artist=True)

Incorrect:
for i in range(4):
    plt.subplot(2, 2, i + 1)
    plt.bxp(...)

Correct:
for i in range(4):
    plt.subplot(2, 2, i + 1)
    ax = plt.gca()
    ax.bxp(...)

Additional rule, very important, specifically to prevent stats dicts from being wrapped in one extra list:
- The first argument passed to .bxp(...) must ultimately be list[dict], where each dict corresponds to one box. Do not pass list[list[dict]], dict_values([[dict], ...]), or a single dict to .bxp(...).
- Do not mechanically rewrite every dictionary value into [{...}]. You must decide the container shape according to the first argument of the original .boxplot(...).
- Aggregated-call scenario: If the original code is ax.boxplot(data.values(), ...) or ax.boxplot(list(data.values()), ...), and data is a mapping from label to the original sample sequence for a single box, then after rewriting, each value of data should directly be a stats dict:
  - Correct: data = {"A": {"med": ..., "q1": ..., "q3": ..., "whislo": ..., "whishi": ..., "fliers": []}, ...}; ax.bxp(list(data.values()), ...)
  - Incorrect: data = {"A": [{"med": ..., ...}], ...}; ax.bxp(list(data.values()), ...)
  - The incorrect form makes list(data.values()) become [[dict], [dict], ...], instead of [dict, dict, ...].
- Single-item loop-call scenario: If the original code is for key, values in data.items(): ax.boxplot(values, ...), and each loop iteration only draws the current key's box, then after rewriting, each value of data should usually be [stats_dict], while keeping ax.bxp(values, ...).
  - Correct: data = {"A": [{"med": ..., "q1": ..., "q3": ..., "whislo": ..., "whishi": ..., "fliers": []}], ...}; for key, values in data.items(): ax.bxp(values, ...)
  - Incorrect: data = {"A": {"med": ..., ...}, ...}; for key, values in data.items(): ax.bxp(values, ...)
  - The incorrect form makes ax.bxp(values) receive a single dict instead of list[dict].
- If the original code is ax.boxplot([values], ...), ensure that the first argument of .bxp(...) is still list[dict]. You may choose data[key] = stats_dict and write ax.bxp([values], ...), or choose data[key] = [stats_dict] and write ax.bxp(values, ...), but do not pass [[dict]] to .bxp(...).
- After rewriting, you must perform a shape self-check: mentally expand the first argument of .bxp(...). It must look like [{"med": ...}, {"med": ...}], not like [[{"med": ...}], [{"med": ...}]].
- If the original code later uses ax.set_xticklabels(data.keys(), ...) or plt.xticks(..., data.keys(), ...), usually keep these tick-label statements unchanged. Do not add an extra list nesting to every stats dict because of tick labels.
- Do not add default style fields by yourself. For example, if the original boxprops only contains dict(facecolor=...), and target_bxp_kwargs_literal also only requires facecolor, do not additionally generate default style keys such as {'linestyle': 'solid'} or {'edgecolor': ...}.

================ ICL EXAMPLES ================

Example 1: List + loop; uniformly add the fliers field for every box.

Input code:
import matplotlib.pyplot as plt

series_list = [
    [[1, 2, 3], [2, 3, 4]],
    [[3, 4, 5], [4, 5, 6]],
]

fig, ax = plt.subplots()
for i, series in enumerate(series_list):
    ax.boxplot(
        series,
        positions=[i * 3 + 1, i * 3 + 2],
        widths=0.4,
        patch_artist=True
    )

Given target:
Call 0:
target_bxpstats_literal = [
    {"med": 2, "q1": 1.5, "q3": 2.5, "whislo": 1, "whishi": 3},
    {"med": 3, "q1": 2.5, "q3": 3.5, "whislo": 2, "whishi": 4}
]
target_bxp_kwargs_literal = {"positions": [1, 2], "widths": 0.4, "patch_artist": True}

Call 1:
target_bxpstats_literal = [
    {"med": 4, "q1": 3.5, "q3": 4.5, "whislo": 3, "whishi": 5},
    {"med": 5, "q1": 4.5, "q3": 5.5, "whislo": 4, "whishi": 6}
]
target_bxp_kwargs_literal = {"positions": [4, 5], "widths": 0.4, "patch_artist": True}

Output code:
import matplotlib.pyplot as plt

series_list = [
    [
        {"med": 2, "q1": 1.5, "q3": 2.5, "whislo": 1, "whishi": 3, "fliers": []},
        {"med": 3, "q1": 2.5, "q3": 3.5, "whislo": 2, "whishi": 4, "fliers": []}
    ],
    [
        {"med": 4, "q1": 3.5, "q3": 4.5, "whislo": 3, "whishi": 5, "fliers": []},
        {"med": 5, "q1": 4.5, "q3": 5.5, "whislo": 4, "whishi": 6, "fliers": []}
    ],
]

fig, ax = plt.subplots()
for i, series in enumerate(series_list):
    ax.bxp(
        series,
        positions=[i * 3 + 1, i * 3 + 2],
        widths=0.4,
        patch_artist=True
    )

Explanation:
- Directly rewrite the original series_list into bxpstats form.
- Even if fliers are not provided in the target, uniformly add empty lists.
- Rewrite boxplot into bxp in place.
- Do not introduce bxpstats_list / kwargs_list / *_all.
- Do not rewrite the loop into another structure.

------------------------------------------------

Example 2: Dictionary + loop; the original code uses sym='' to hide fliers. After normalization, still write fliers: [] for every box.

Input code:
import matplotlib.pyplot as plt

data = {
    "A": [[1, 2, 3], [2, 3, 4]],
    "B": [[3, 4, 5], [4, 5, 6]],
}

fig, ax = plt.subplots()
for key, series in data.items():
    ax.boxplot(
        series,
        widths=0.5,
        sym='',
        patch_artist=True,
        boxprops={"facecolor": "white"}
    )

Given target:
For "A":
target_bxpstats_literal = [
    {"med": 2, "q1": 1.5, "q3": 2.5, "whislo": 1, "whishi": 3},
    {"med": 3, "q1": 2.5, "q3": 3.5, "whislo": 2, "whishi": 4}
]
target_bxp_kwargs_literal = {"widths": 0.5, "patch_artist": True, "boxprops": {"facecolor": "white"}}

For "B":
target_bxpstats_literal = [
    {"med": 4, "q1": 3.5, "q3": 4.5, "whislo": 3, "whishi": 5},
    {"med": 5, "q1": 4.5, "q3": 5.5, "whislo": 4, "whishi": 6}
]
target_bxp_kwargs_literal = {"widths": 0.5, "patch_artist": True, "boxprops": {"facecolor": "white"}}

Output code:
import matplotlib.pyplot as plt

data = {
    "A": [
        {"med": 2, "q1": 1.5, "q3": 2.5, "whislo": 1, "whishi": 3, "fliers": []},
        {"med": 3, "q1": 2.5, "q3": 3.5, "whislo": 2, "whishi": 4, "fliers": []}
    ],
    "B": [
        {"med": 4, "q1": 3.5, "q3": 4.5, "whislo": 3, "whishi": 5, "fliers": []},
        {"med": 5, "q1": 4.5, "q3": 5.5, "whislo": 4, "whishi": 6, "fliers": []}
    ],
}

fig, ax = plt.subplots()
for key, series in data.items():
    ax.bxp(
        series,
        widths=0.5,
        patch_artist=True,
        boxprops={"facecolor": "white"}
    )

Explanation:
- sym='' in the original boxplot means hiding fliers.
- Under the unified strategy, still add fliers: [] to every stats dict.
- Because these fliers lists are empty, no visible flier points are usually produced.
- Directly rewrite the original data dictionary values.
- Do not write if/elif branches over key inside the loop to temporarily construct bxpstats / kwargs.

------------------------------------------------

Example 3: Dictionary + loop; real fliers are given in the target and must be preserved.

Input code:
import matplotlib.pyplot as plt

groups = {
    "North": [[10, 11, 12, 13, 40]],
    "South": [[9, 10, 11, 12, 38]],
}

fig, ax = plt.subplots()
for name, values in groups.items():
    ax.boxplot(
        values,
        widths=0.6,
        patch_artist=True,
        flierprops={"marker": "o", "markersize": 6}
    )

Given target:
For "North":
target_bxpstats_literal = [
    {"med": 12, "q1": 11, "q3": 13, "whislo": 10, "whishi": 13, "fliers": [40]}
]
target_bxp_kwargs_literal = {"widths": 0.6, "patch_artist": True, "flierprops": {"marker": "o", "markersize": 6}}

For "South":
target_bxpstats_literal = [
    {"med": 11, "q1": 10, "q3": 12, "whislo": 9, "whishi": 12, "fliers": [38]}
]
target_bxp_kwargs_literal = {"widths": 0.6, "patch_artist": True, "flierprops": {"marker": "o", "markersize": 6}}

Output code:
import matplotlib.pyplot as plt

groups = {
    "North": [
        {"med": 12, "q1": 11, "q3": 13, "whislo": 10, "whishi": 13, "fliers": [40]}
    ],
    "South": [
        {"med": 11, "q1": 10, "q3": 12, "whislo": 9, "whishi": 12, "fliers": [38]}
    ],
}

fig, ax = plt.subplots()
for name, values in groups.items():
    ax.bxp(
        values,
        widths=0.6,
        patch_artist=True,
        flierprops={"marker": "o", "markersize": 6}
    )

Explanation:
- The target already contains real fliers, so they must be preserved.
- The unified strategy is not "replace all fliers with empty lists"; it is "every box must have a fliers field".
- Keep the original dictionary + loop structure unchanged.

------------------------------------------------

Example 4: Randomly generated data + tick_labels; tick labels should become the label field inside each stats dict.

Input code:
import numpy as np
import matplotlib.pyplot as plt

rng = np.random.default_rng(0)
samples = [
    rng.normal(0, 1, size=20),
    rng.normal(1, 1, size=20),
]

fig, ax = plt.subplots()
for _ in range(2):
    ax.boxplot(
        samples,
        positions=[1, 2],
        widths=0.3,
        tick_labels=["Left", "Right"]
    )

Given target:
For each call in the loop:
target_bxpstats_literal = [
    {"med": -0.1, "q1": -0.6, "q3": 0.5, "whislo": -1.9, "whishi": 1.7, "label": "Left"},
    {"med": 1.2, "q1": 0.4, "q3": 1.9, "whislo": -0.2, "whishi": 2.8, "label": "Right"}
]
target_bxp_kwargs_literal = {"positions": [1, 2], "widths": 0.3}

Output code:
import numpy as np
import matplotlib.pyplot as plt

rng = np.random.default_rng(0)
samples = [
    {"med": -0.1, "q1": -0.6, "q3": 0.5, "whislo": -1.9, "whishi": 1.7, "label": "Left", "fliers": []},
    {"med": 1.2, "q1": 0.4, "q3": 1.9, "whislo": -0.2, "whishi": 2.8, "label": "Right", "fliers": []}
]

fig, ax = plt.subplots()
for _ in range(2):
    ax.bxp(
        samples,
        positions=[1, 2],
        widths=0.3
    )

Explanation:
- Even if the original data comes from np.random, the final code should directly rewrite the original variable into target_bxpstats_literal.
- tick_labels / labels are data-side semantics of boxplot. After rewriting, they should be represented by the label field in each stats dict.
- Do not continue passing tick_labels to bxp.
- Keep the loop structure unchanged.

------------------------------------------------

Example 5: Subplots + nested loop + notch; when cilo/cihi exist in the target, shownotches=True must be written explicitly.

Input code:
import numpy as np
import matplotlib.pyplot as plt

rng = np.random.default_rng(1)
panels = [
    {
        "title": "Left",
        "values": [rng.normal(0, 1, 30), rng.normal(1, 1, 30)],
        "positions": [1, 2]
    },
    {
        "title": "Right",
        "values": [rng.normal(2, 1, 30), rng.normal(3, 1, 30)],
        "positions": [1, 2]
    },
]

fig, axs = plt.subplots(1, 2)
for ax, panel in zip(axs, panels):
    ax.boxplot(
        panel["values"],
        positions=panel["positions"],
        widths=0.35,
        notch=True,
        patch_artist=True
    )
    ax.set_title(panel["title"])

Given target:
Left subplot:
target_bxpstats_literal = [
    {"med": 0.1, "q1": -0.5, "q3": 0.8, "whislo": -1.7, "whishi": 1.9, "cilo": -0.2, "cihi": 0.4},
    {"med": 1.0, "q1": 0.4, "q3": 1.5, "whislo": -0.2, "whishi": 2.5, "cilo": 0.7, "cihi": 1.3}
]
target_bxp_kwargs_literal = {"positions": [1, 2], "widths": 0.35, "patch_artist": True, "shownotches": True}

Right subplot:
target_bxpstats_literal = [
    {"med": 2.2, "q1": 1.5, "q3": 2.8, "whislo": 0.9, "whishi": 3.6, "cilo": 1.9, "cihi": 2.5},
    {"med": 3.1, "q1": 2.4, "q3": 3.7, "whislo": 1.6, "whishi": 4.4, "cilo": 2.8, "cihi": 3.4}
]
target_bxp_kwargs_literal = {"positions": [1, 2], "widths": 0.35, "patch_artist": True, "shownotches": True}

Output code:
import numpy as np
import matplotlib.pyplot as plt

rng = np.random.default_rng(1)
panels = [
    {
        "title": "Left",
        "values": [
            {"med": 0.1, "q1": -0.5, "q3": 0.8, "whislo": -1.7, "whishi": 1.9, "cilo": -0.2, "cihi": 0.4, "fliers": []},
            {"med": 1.0, "q1": 0.4, "q3": 1.5, "whislo": -0.2, "whishi": 2.5, "cilo": 0.7, "cihi": 1.3, "fliers": []}
        ],
        "positions": [1, 2]
    },
    {
        "title": "Right",
        "values": [
            {"med": 2.2, "q1": 1.5, "q3": 2.8, "whislo": 0.9, "whishi": 3.6, "cilo": 1.9, "cihi": 2.5, "fliers": []},
            {"med": 3.1, "q1": 2.4, "q3": 3.7, "whislo": 1.6, "whishi": 4.4, "cilo": 2.8, "cihi": 3.4, "fliers": []}
        ],
        "positions": [1, 2]
    },
]

fig, axs = plt.subplots(1, 2)
for ax, panel in zip(axs, panels):
    ax.bxp(
        panel["values"],
        positions=panel["positions"],
        widths=0.35,
        patch_artist=True,
        shownotches=True
    )
    ax.set_title(panel["title"])

Explanation:
- When notch is visible, cilo/cihi in the target must be preserved.
- As long as cilo/cihi exist in the target, this .bxp(...) call must explicitly include shownotches=True.
- Keep the subplot and loop structure completely unchanged.
- Also uniformly add the fliers field to every stats dict.

------------------------------------------------

Example 6: DataFrame + iterrows + inline construction of five-number summaries; adding a nearby bxpstats variable is allowed.

Input code:
import matplotlib.pyplot as plt
import pandas as pd

data = {
    "name": ["A", "B"],
    "Min": [1, 2],
    "Q1": [2, 3],
    "Median": [3, 4],
    "Q3": [4, 5],
    "Max": [5, 6],
}
df = pd.DataFrame(data)

fig, ax = plt.subplots()
for i, row in df.iterrows():
    ax.boxplot(
        [row["Min"], row["Q1"], row["Median"], row["Q3"], row["Max"]],
        positions=[i],
        widths=0.5,
        patch_artist=True
    )

Given target:
Call 0:
target_bxpstats_literal = [
    {"med": 3, "q1": 2, "q3": 4, "whislo": 1, "whishi": 5}
]
Call 1:
target_bxpstats_literal = [
    {"med": 4, "q1": 3, "q3": 5, "whislo": 2, "whishi": 6}
]
target_bxp_kwargs_literal = {"positions": "same_as_original", "widths": 0.5, "patch_artist": True}

Output code:
import matplotlib.pyplot as plt
import pandas as pd

data = {
    "name": ["A", "B"],
    "Min": [1, 2],
    "Q1": [2, 3],
    "Median": [3, 4],
    "Q3": [4, 5],
    "Max": [5, 6],
}
df = pd.DataFrame(data)
bxpstats = [
    [{"med": 3, "q1": 2, "q3": 4, "whislo": 1, "whishi": 5, "fliers": []}],
    [{"med": 4, "q1": 3, "q3": 5, "whislo": 2, "whishi": 6, "fliers": []}],
]

fig, ax = plt.subplots()
for i, row in df.iterrows():
    ax.bxp(
        bxpstats[i],
        positions=[i],
        widths=0.5,
        patch_artist=True
    )

Explanation:
- The real input to boxplot here is not the whole df, but a temporary list expression inside the loop.
- Because there is no independent original variable carrying the box data, adding a nearby bxpstats variable is allowed.
- However, it must be written statically once; do not construct it inside the loop with if/elif branches over i.
- Keep df because it carries other non-box semantics.

------------------------------------------------

Example 7: The same box data is reused by the main plot and an inset. Rewrite the original variable once and reuse it in both places.

Input code:
import matplotlib.pyplot as plt

values = [
    [10, 20, 30, 40],
    [1, 2, 3, 4],
]

fig, ax = plt.subplots()
ax.boxplot(values, positions=[1, 2], widths=0.5, patch_artist=True)

inset = fig.add_axes([0.6, 0.5, 0.3, 0.3])
inset.boxplot(values, positions=[1, 2], widths=0.5, patch_artist=True)

Given target:
target_bxpstats_literal = [
    {"med": 25, "q1": 17.5, "q3": 32.5, "whislo": 10, "whishi": 40},
    {"med": 2.5, "q1": 1.75, "q3": 3.25, "whislo": 1, "whishi": 4}
]
target_bxp_kwargs_literal = {"positions": [1, 2], "widths": 0.5, "patch_artist": True}

Output code:
import matplotlib.pyplot as plt

values = [
    {"med": 25, "q1": 17.5, "q3": 32.5, "whislo": 10, "whishi": 40, "fliers": []},
    {"med": 2.5, "q1": 1.75, "q3": 3.25, "whislo": 1, "whishi": 4, "fliers": []}
]

fig, ax = plt.subplots()
ax.bxp(values, positions=[1, 2], widths=0.5, patch_artist=True)

inset = fig.add_axes([0.6, 0.5, 0.3, 0.3])
inset.bxp(values, positions=[1, 2], widths=0.5, patch_artist=True)

Explanation:
- When the same original box data is reused in multiple places, rewrite the original variable only once.
- Both .bxp(...) calls directly reuse the same rewritten values.
- Do not create separate bxpstats for the main plot and the inset.

------------------------------------------------

Example 8: Dictionary + enumerate + style depending on i; directly rewrite dictionary values and do not write ten if/elif branches over i inside the loop.

Input code:
import matplotlib.pyplot as plt

data = {
    "RX": [1, 2, 3, 4, 5],
    "KRX": [2, 3, 4, 5, 6],
}
colors = ["red", "blue"]

fig, ax = plt.subplots()
for i, (name, values) in enumerate(data.items()):
    ax.boxplot(
        values,
        positions=[i],
        widths=0.6,
        patch_artist=True,
        boxprops=dict(facecolor=colors[i], color=colors[i]),
        medianprops=dict(color="orange")
    )

Given target:
For "RX":
target_bxpstats_literal = [
    {"med": 3, "q1": 2, "q3": 4, "whislo": 1, "whishi": 5}
]
For "KRX":
target_bxpstats_literal = [
    {"med": 4, "q1": 3, "q3": 5, "whislo": 2, "whishi": 6}
]
target_bxp_kwargs_literal = {"positions": "same_as_original", "widths": 0.6, "patch_artist": True, "boxprops": "same_as_original", "medianprops": {"color": "orange"}}

Output code:
import matplotlib.pyplot as plt

data = {
    "RX": [{"med": 3, "q1": 2, "q3": 4, "whislo": 1, "whishi": 5, "fliers": []}],
    "KRX": [{"med": 4, "q1": 3, "q3": 5, "whislo": 2, "whishi": 6, "fliers": []}],
}
colors = ["red", "blue"]

fig, ax = plt.subplots()
for i, (name, values) in enumerate(data.items()):
    ax.bxp(
        values,
        positions=[i],
        widths=0.6,
        patch_artist=True,
        boxprops=dict(facecolor=colors[i], color=colors[i]),
        medianprops=dict(color="orange")
    )

Explanation:
- When the style depends on i, still keep the original enumerate(data.items()) loop.
- Only rewrite the values of data into bxpstats. Do not branch by i inside the loop and write multiple ax.bxp(...) calls.
- Although such an if/elif-expanded rewrite can run, it changes the format too much and must be avoided.



================ END OF ICL EXAMPLES ================

Now process the real task.

Task: Minimally rewrite the following Matplotlib code and output only the complete rewritten Python code.

Goal:
Rewrite every .boxplot(...) call into an equivalent .bxp(...) call while keeping the rendered image unchanged.

You only need to do three things:
1. Find the original box-data variable or data-carrying location actually used by the current boxplot call.
2. Prefer rewriting the original samples directly into the given target_bxpstats_literal at that original location. However, if direct rewriting would break the original container structure and cause later code to fail, for example by breaking equal-length columns required by pd.DataFrame(data), then add a bxpstats variable immediately next to the call that only serves this call.
3. Rewrite the original .boxplot(...) call in place into .bxp(...), and only preserve parameters that are truly needed for this call and are compatible with bxp. At the same time, attach .bxp(...) to the correct axes object:
   - If the original is ax.boxplot(...), write ax.bxp(...).
   - If the original is axs[i].boxplot(...), write axs[i].bxp(...).
   - If the original is inset.boxplot(...), write inset.bxp(...).
   - If the original is plt.boxplot(...), do not write plt.bxp(...). Instead, minimally add ax = plt.gca() nearby and then write ax.bxp(...).

If the current box data has no independent variable and is instead an immediate expression constructed inside a loop, such as [row['Min'], row['Q1'], row['Median'], row['Q3'], row['Max']], then you may add a bxpstats variable immediately next to the call that only serves this call. However, it must be written statically once; do not construct it temporarily inside the loop via if/elif over i, key, or row.

You must obey:
1. Make only minimal necessary changes. Do not rewrite the entire code.
2. Explicitly remove the original box data. Do not keep using it as plotting input in the final code.
3. Prefer directly modifying the original variable. Avoid defining new variables whenever possible.
4. If a new variable is necessary, only add a nearby bxpstats variable, and it may only serve the current call.
5. Do not add any templated variables for rewriting convenience, such as:
   - bxpstats_list / bxp_kwargs_list / *_all / kwargs_all
   - stats_by_axis / kwargs_by_axis / stats_by_name / kwargs_by_name
   - if/elif branches inside loops over i or key that temporarily construct bxpstats and kwargs
6. Do not extract parameters that can be directly written inside .bxp(...) into kwargs / kwargs_list. If the original parameters are already written inline, prefer keeping them inline after rewriting.
6.1 Do not output plt.bxp(...).
6.2 For pyplot-style plt.boxplot(...), prefer the minimal nearby patch ax = plt.gca() + ax.bxp(...).
6.3 If the original code uses plt.subplot(...) / plt.sca(...) to manage the current axes, do not refactor it into fig, axs = plt.subplots(...) because of bxp, unless the original code already uses that structure.
6.4 If the original boxplot return value is assigned to a variable, preserve that variable as the receiver of ax.bxp(...) as much as possible.
7. Do not aggregate and redraw multiple original boxplot calls together. Each original call corresponds only to its own target.
8. Do not change existing loops, subplot structures, variable names, titles, axes, legends, grids, layout, savefig, or show, unless required for the rewrite.
9. Do not reorder statements, merge loops, split loops, copy loops, or change the data organization.
10. If the same original box data is reused by multiple calls, do not rewrite it into multiple duplicated stats variables. Prefer rewriting once and reusing it.
11. If an original container carries both box data and non-box semantics, such as week, name, month, or legend-text sources, only replace the part actually passed into boxplot. Do not transform the whole container into another template structure.
12. If an original container is later used by pd.DataFrame(data) or other code that depends on a fixed shape / equal-length columns, do not directly rewrite values inside that container into bxpstats. In that case, preserve the original container structure and add a nearby bxpstats variable.
13. Do not casually fix issues unrelated to boxplot -> bxp unless the code cannot run without the fix.
14. When "passing fewer parameters" conflicts with "runtime validity / visual equivalence", prioritize runtime validity and visual equivalence.

Statistics requirements:
15. Strictly use the given target_bxpstats_literal. Write it directly and do not recompute it.
16. Do not add statistical computation code such as np.percentile, statistics, pandas, or matplotlib.cbook.boxplot_stats.
17. Use the numeric values in target_bxpstats_literal exactly as given. Do not change numeric formatting, add trailing zeros, or recompute values.
18. Only preserve statistics that are truly needed in the figure and already given in target_bxpstats_literal. Do not add or remove statistics by yourself.
19. Every stats dict in the final code must contain a fliers field.
20. If a box in target_bxpstats_literal does not provide fliers, uniformly add 'fliers': [].
21. If every dict in target_bxpstats_literal already contains "label", that is the tick label for the box. It is not the legend label passed as .bxp(..., label=...).
22. Do not confuse the "label" field inside the stats dict with .bxp(..., label=...):
   - "label" inside the stats dict: tick label
   - .bxp(..., label=...): legend label
22.1 The first argument passed to .bxp(...) must be list[dict]. If the call is an aggregated data.values() / list(data.values()) call, each value of data should directly be a dict. If the call is ax.bxp(values) inside a loop and each values represents one box, then values should be [dict].
22.2 Do not let the first argument of .bxp(...) become list[list[dict]], dict_values([[dict], ...]), or a single dict.

Mandatory rules about fliers:
23. If a box in target_bxpstats_literal contains real fliers, preserve these fliers. Do not delete, rewrite, or replace them with an empty list.
24. If a box in target_bxpstats_literal does not contain fliers, uniformly add 'fliers': [].
25. Do not omit the fliers field just because there is no visible flier in the current plot. Every stats dict in the final code must have this key.
26. Since every stats dict uniformly contains a fliers field, usually there is no need to additionally write showfliers=False.
27. If target_bxp_kwargs_literal explicitly provides showfliers=False, preserve it according to the target. Otherwise, usually do not add showfliers=False by yourself.
28. If the original boxplot hides fliers through sym='' or sym="", under the unified strategy, you can still keep the visible result unchanged by writing 'fliers': [] for every box. Usually there is no need to additionally pass sym or showfliers=False.

Mandatory rules about notch:
29. If target_bxpstats_literal contains cilo/cihi, preserve them.
30. As long as target_bxpstats_literal contains cilo/cihi, the .bxp(...) call must explicitly include shownotches=True.
31. If target_bxpstats_literal does not contain cilo/cihi, do not additionally pass shownotches=True.
32. Do not delete cilo/cihi given in the target, and do not add cilo/cihi that do not exist.

Mandatory rules about mean:
33. If target_bxpstats_literal contains mean, preserve mean.
34. As long as target_bxpstats_literal contains mean, the .bxp(...) call must explicitly include showmeans=True.
35. If target_bxp_kwargs_literal provides meanline=True, preserve meanline=True.
36. If target_bxpstats_literal does not contain mean, do not additionally pass showmeans=True, and do not add mean by yourself.

bxp parameter requirements:
37. Each .bxp(...) call should only pass parameters that are truly needed for this call and explicitly given by target_bxp_kwargs_literal.
38. If target_bxp_kwargs_literal is <omit>, usually write .bxp(bxpstats) or ax.bxp(bxpstats).
39. Do not explicitly add default parameters to .bxp(...).
40. Do not continue passing these boxplot-side statistical input/control parameters to .bxp(...):
    x, whis, bootstrap, autorange, usermedians, conf_intervals, tick_labels, labels, sym, data.
41. If a parameter is not given by target_bxp_kwargs_literal, do not add it by yourself, unless it is one of the minimal necessary parameters explicitly required above.
42. If target_bxp_kwargs_literal provides positions, widths, vert, orientation, patch_artist, boxprops, whiskerprops, flierprops, medianprops, capprops, meanprops, meanline, showcaps, showbox, manage_ticks, zorder, capwidths, or label, preserve it as given.
42.1 If target_bxp_kwargs_literal does not provide a style key, do not add default style keys by yourself. For example, do not expand boxprops=dict(facecolor=...) into {'facecolor': ..., 'linestyle': 'solid'}.

Visual equivalence requirements:
43. The rewritten code must preserve the original rendered image as strictly as possible, including but not limited to:
   - box positions, widths, and orientation
   - q1 / q3 / median / whiskers / caps
   - visible flier points
   - visible notch shape
   - visible mean marker / mean line
   - colors, linestyles, fill, and alpha
   - ticks, tick labels, title, legend, grid, layout, savefig, and show
44. If an original boxplot parameter is itself used to compute statistics from the raw samples, and these statistics are now directly provided by target_bxpstats_literal, do not preserve that parameter.
45. If an original boxplot parameter only affects invisible parts, and removing it does not change the final visible result or affect .bxp(...) validity, do not preserve it.
46. If the original code still contains local expressions depending on original samples after boxplot, such as max(values), min(values), np.max(values), or np.min(values), and those values have already been replaced by bxpstats, only make a local, minimal, nearby compatibility patch, such as reading whishi / whislo / fliers / mean. Do not refactor the whole loop, duplicate data, or keep original samples because of this.
46.1 If the original code still heavily uses pyplot style, such as plt.title, plt.xticks, plt.ylabel, plt.ylim, and plt.legend, then after correctly using ax = plt.gca(); ax.bxp(...), usually there is no need to rewrite all of these statements.

Output requirements:
47. Output only the final code. Do not explain. Do not output a diff.
48. The output must be complete Python code that can be run directly.

File:
{{SOURCE_FILE}}

Precomputed targets for each boxplot call:
{{CALL_TARGETS}}

Original code:
{{PY_CODE}}
\end{quote}
\end{promptbox}

\begin{promptbox}[title={Rewriting prompt for Histogram},label={box:rewriting_prompt_hist}]
\begin{quote}
\small
You are an assistant that rewrites code with minimal changes only.

Your task is to minimally rewrite Matplotlib code without changing the final rendered plot.

The most important hard constraints:
1. The image produced by the rewritten code must remain equivalent to the original image and must not be changed.
2. "Must not be changed" includes: bar heights, positions, widths, edges, overlapping/stacking relationships, orientation, colors, alpha, linestyles, labels, axes, titles, legends, grids, layouts, and saved outputs. All of them must remain the same.
3. You may only change the representation form of the data. You must not change the final image.
4. You must not refactor the program structure just to make the code cleaner.

Rewriting principles for hist:
1. The goal is to rewrite "raw sample inputs" into an equivalent "bin + weight" form.
2. Prefer directly modifying the original data variable. Avoid defining new variables whenever possible.
3. If a new variable must be defined, its name should stay close to the original variable name, for example:
   - values -> values_weights, values_bins
   - data -> data_weights, data_bins
   - datasets -> datasets_weights, datasets_bins
   Do not rename it into abstract names far from the original meaning, such as x_list_all, hist_targets, plan_by_axis, tmp_stats, etc.
4. Keep the original hist call location unchanged. Avoid modifying loops, subplots, styles, and layouts.
5. If the original data is already in bin + weight form, do not modify it.
6. If the bins are uniform:
   - use np.linspace(left_edge, right_edge, len(weights)+1) for bins
   - use bins[:-1] as x
   - the code must directly write len(weights) or len(weights[i]); do not hard-code the length as a concrete number
   - do not expand uniform bins into a long explicit boundary list
7. If the bins are non-uniform:
   - keep explicit bin boundaries
   - use the left edge of each bin as x, i.e., bins[:-1]
8. If a single hist call originally receives multiple datasets, the rewritten code must preserve the "single call + multiple datasets" structure. Do not split or merge calls.
9. Do not refactor the code into a unified template style such as x_list, weights_list, bins_all, hist_targets, or *_all.
10. Do not add statistical computation logic such as np.histogram, numpy.histogram, pandas, or statistics.
11. Except for rewriting raw sample inputs into bin + weight form, all parameters that truly affect appearance must be preserved, such as histtype, stacked, align, orientation, rwidth, log, color, label, bottom, alpha, linewidth, edgecolor, etc.
12. For ordinary frequency histograms with the default density=False, directly use the final visible frequencies as weights.
13. For histograms with density=True, directly use the final visible density heights as weights. After rewriting, usually do not preserve density=True, because these heights are already the final visible result.
14. For histograms with cumulative=True, directly use the final visible cumulative heights as weights. After rewriting, usually do not preserve cumulative=True, because these heights are already the final visible result.
15. histtype is only a plotting-style parameter and must be preserved unchanged. Do not change bar / barstacked / step / stepfilled styles because of rewriting.
16. If multiple datasets are passed into a single hist call and they share one set of bins, the rewritten code must still share the same bins. Do not create separate bins for each dataset.

Additional notes:
1. If the original code is a dictionary + for-loop that calls hist multiple times, do not rewrite it into hard-coded branches such as if key == ... / elif .... Prefer preserving the original loop structure.
2. If the original code first constructs a dictionary, and then flattens or merges samples from the dictionary into a single one-dimensional input through flatten, list comprehension, np.concatenate, sum(..., []), chain.from_iterable, etc., before passing it to hist, directly rewrite this merged one-dimensional variable. Do not go back and rewrite each key separately.
3. If the original sample variable is no longer needed after rewriting, do not keep a raw copy of it. Do not keep both a raw dict and a raw flat variable at the same time.

First study the following examples. The key points are:
- in-place replacement
- minimal changes
- do not refactor program structure
- do not change the final image
- keep rewritten variable names close to the original variable names

====================
ICL Example 1: Single dataset, ordinary frequency histogram, uniform bins
====================

Input code:
import matplotlib.pyplot as plt

values = [1.2, 0.9, 1.5, 2.1, 2.4, 2.6, 2.8, 3.1, 3.4, 3.7, 4.2, 4.6]
fig, ax = plt.subplots()
ax.hist(values, bins=5, color="steelblue", alpha=0.8)
plt.show()

Given target:
rewrite_mode = uniform_bins_weights
target_weights_literal = [2, 1, 4, 3, 2]
target_x_literal_hint = np.linspace(0.9, 4.9, len(WEIGHTS_VAR) + 1)[:-1]
target_bins_literal_hint = np.linspace(0.9, 4.9, len(WEIGHTS_VAR) + 1)
target_hist_kwargs_literal = {"color": "steelblue", "alpha": 0.8}

Output code:
import matplotlib.pyplot as plt
import numpy as np

values_weights = [2, 1, 4, 3, 2]
values = np.linspace(0.9, 4.9, len(values_weights) + 1)[:-1]

fig, ax = plt.subplots()
ax.hist(values, bins=np.linspace(0.9, 4.9, len(values_weights) + 1), weights=values_weights, color="steelblue", alpha=0.8)
plt.show()

Explanation:
- The input contains raw data itself.
- Directly replace the raw samples in values with pre-binned x values.
- The new variable name values_weights remains close to the original variable name values.
- The final image remains unchanged.

====================
ICL Example 2: Single dataset, step histogram, non-uniform bins
====================

Input code:
import matplotlib.pyplot as plt

values = [0.2, 0.5, 1.4, 2.1, 2.7, 3.2, 5.5, 7.8, 8.4, 9.1]
fig, ax = plt.subplots()
ax.hist(values, bins=[0, 1, 3, 6, 10], histtype="step", color="black", linewidth=2)
plt.show()

Given target:
rewrite_mode = explicit_bins_weights
target_weights_literal = [2, 3, 2, 3]
target_x_literal_explicit = np.array([0, 1, 3, 6])
target_bins_literal_explicit = np.array([0, 1, 3, 6, 10])
target_hist_kwargs_literal = {"histtype": "step", "color": "black", "linewidth": 2}

Output code:
import matplotlib.pyplot as plt
import numpy as np

values_weights = [2, 3, 2, 3]
values = np.array([0, 1, 3, 6])

fig, ax = plt.subplots()
ax.hist(values, bins=np.array([0, 1, 3, 6, 10]), weights=values_weights, histtype="step", color="black", linewidth=2)
plt.show()

Explanation:
- Do not rewrite non-uniform bins into np.linspace(...).
- Preserve histtype="step".
- The new variable name still stays close to values.
- The final image remains unchanged.

====================
ICL Example 3: The original code is already in bin + weight form; no change
====================

Input code:
import matplotlib.pyplot as plt
import numpy as np

weights = [4, 7, 5, 2]
bins = np.linspace(0, 4, len(weights) + 1)
x = bins[:-1]

fig, ax = plt.subplots()
ax.hist(x, bins=bins, weights=weights, color="orange")
plt.show()

Given target:
rewrite_mode = already_binned_no_change

Output code:
import matplotlib.pyplot as plt
import numpy as np

weights = [4, 7, 5, 2]
bins = np.linspace(0, 4, len(weights) + 1)
x = bins[:-1]

fig, ax = plt.subplots()
ax.hist(x, bins=bins, weights=weights, color="orange")
plt.show()

Explanation:
- The code is already in bin + weight form.
- Do not modify this case again.
- The final image remains unchanged.

====================
ICL Example 4: A single call receives multiple datasets and shares one set of bins
====================

Input code:
import matplotlib.pyplot as plt

datasets = [
    [0.2, 0.8, 1.1, 1.9, 2.2, 2.4, 3.3, 3.8],
    [0.4, 0.9, 1.5, 2.0, 2.8, 3.1, 3.6, 3.9]
]

fig, ax = plt.subplots()
ax.hist(datasets, bins=4, color=["red", "blue"], label=["A", "B"])
plt.show()

Given target:
rewrite_mode = uniform_bins_weights
target_weights_literal = [[2, 2, 2, 2], [2, 1, 2, 3]]
target_x_literal_hint = [np.linspace(0.2, 4.2, len(WEIGHTS_VAR[0]) + 1)[:-1], np.linspace(0.2, 4.2, len(WEIGHTS_VAR[1]) + 1)[:-1]]
target_bins_literal_hint = np.linspace(0.2, 4.2, len(WEIGHTS_VAR[0]) + 1)
target_hist_kwargs_literal = {"color": ["red", "blue"], "label": ["A", "B"]}

Output code:
import matplotlib.pyplot as plt
import numpy as np

datasets_weights = [[2, 2, 2, 2], [2, 1, 2, 3]]
datasets = [
    np.linspace(0.2, 4.2, len(datasets_weights[0]) + 1)[:-1],
    np.linspace(0.2, 4.2, len(datasets_weights[1]) + 1)[:-1]
]

fig, ax = plt.subplots()
ax.hist(datasets, bins=np.linspace(0.2, 4.2, len(datasets_weights[0]) + 1), weights=datasets_weights, color=["red", "blue"], label=["A", "B"])
plt.show()

Explanation:
- The input contains multiple raw datasets.
- Preserve the original "single call + multiple datasets" shape.
- Preserve the shared bin set.
- The new variable name datasets_weights stays close to datasets.
- The final image remains unchanged.

====================
ICL Example 5: density=True + histtype='step'
====================

Input code:
import matplotlib.pyplot as plt

vals = [0.3, 0.9, 1.1, 1.3, 2.2, 2.5, 2.7, 3.8, 4.1, 4.7]
fig, ax = plt.subplots()
ax.hist(vals, bins=5, density=True, histtype="step", color="purple", linewidth=2)
plt.show()

Given target:
rewrite_mode = uniform_bins_weights
target_weights_literal = [0.1, 0.2, 0.4, 0.2, 0.1]
target_x_literal_hint = np.linspace(0.3, 5.3, len(WEIGHTS_VAR) + 1)[:-1]
target_bins_literal_hint = np.linspace(0.3, 5.3, len(WEIGHTS_VAR) + 1)
target_hist_kwargs_literal = {"histtype": "step", "color": "purple", "linewidth": 2}

Output code:
import matplotlib.pyplot as plt
import numpy as np

vals_weights = [0.1, 0.2, 0.4, 0.2, 0.1]
vals = np.linspace(0.3, 5.3, len(vals_weights) + 1)[:-1]

fig, ax = plt.subplots()
ax.hist(vals, bins=np.linspace(0.3, 5.3, len(vals_weights) + 1), weights=vals_weights, histtype="step", color="purple", linewidth=2)
plt.show()

Explanation:
- The input contains raw data itself.
- The original figure uses density=True.
- After rewriting, directly write the final visible density heights as weights.
- Do not preserve density=True.
- The variable name vals_weights stays close to vals.
- The final image remains unchanged.

====================
ICL Example 6: cumulative=True + horizontal + stacked
====================

Input code:
import matplotlib.pyplot as plt

series = [
    [0.2, 0.7, 1.1, 1.5, 2.2, 2.6, 3.0, 3.5],
    [0.4, 0.8, 1.3, 1.9, 2.1, 2.9, 3.2, 3.8]
]

fig, ax = plt.subplots()
ax.hist(series, bins=4, cumulative=True, stacked=True, orientation="horizontal", color=["#1f77b4", "#ff7f0e"], edgecolor="white")
plt.show()

Given target:
rewrite_mode = uniform_bins_weights
target_weights_literal = [[2, 4, 6, 8], [1, 3, 6, 8]]
target_x_literal_hint = [np.linspace(0.2, 4.2, len(WEIGHTS_VAR[0]) + 1)[:-1], np.linspace(0.2, 4.2, len(WEIGHTS_VAR[1]) + 1)[:-1]]
target_bins_literal_hint = np.linspace(0.2, 4.2, len(WEIGHTS_VAR[0]) + 1)
target_hist_kwargs_literal = {"stacked": True, "orientation": "horizontal", "color": ["#1f77b4", "#ff7f0e"], "edgecolor": "white"}

Output code:
import matplotlib.pyplot as plt
import numpy as np

series_weights = [[2, 4, 6, 8], [1, 3, 6, 8]]
series = [
    np.linspace(0.2, 4.2, len(series_weights[0]) + 1)[:-1],
    np.linspace(0.2, 4.2, len(series_weights[1]) + 1)[:-1]
]

fig, ax = plt.subplots()
ax.hist(
    series,
    bins=np.linspace(0.2, 4.2, len(series_weights[0]) + 1),
    weights=series_weights,
    stacked=True,
    orientation="horizontal",
    color=["#1f77b4", "#ff7f0e"],
    edgecolor="white"
)
plt.show()

Explanation:
- The input contains multiple raw datasets.
- The original figure uses cumulative=True.
- After rewriting, directly write the final visible cumulative heights as weights.
- Preserve stacked and orientation.
- The variable name series_weights stays close to series.
- The final image remains unchanged.

====================
ICL Example 7: A single call receives multiple datasets with very different scales, but still shares bins
====================

Input code:
import matplotlib.pyplot as plt

velocity = [20, 23, 19, 25, 22, 21, 24, 26]
pressure = [1000, 1050, 1100, 1150, 1200, 1250, 1300, 1350]

fig, ax = plt.subplots()
ax.hist([velocity, pressure], bins=10, label=["Velocity", "Pressure"], alpha=0.7, edgecolor="black")
plt.show()

Given target:
rewrite_mode = uniform_bins_weights
target_weights_literal = [[8, 0, 0, 0, 0, 0, 0, 0, 0, 0], [0, 0, 0, 0, 0, 0, 0, 2, 3, 3]]
target_x_literal_hint = [np.linspace(19.0, 1350.0, len(WEIGHTS_VAR[0]) + 1)[:-1], np.linspace(19.0, 1350.0, len(WEIGHTS_VAR[1]) + 1)[:-1]]
target_bins_literal_hint = np.linspace(19.0, 1350.0, len(WEIGHTS_VAR[0]) + 1)
target_hist_kwargs_literal = {"label": ["Velocity", "Pressure"], "alpha": 0.7, "edgecolor": "black"}

Output code:
import matplotlib.pyplot as plt
import numpy as np

velocity_pressure_weights = [[8, 0, 0, 0, 0, 0, 0, 0, 0, 0], [0, 0, 0, 0, 0, 0, 0, 2, 3, 3]]
velocity_pressure = [
    np.linspace(19.0, 1350.0, len(velocity_pressure_weights[0]) + 1)[:-1],
    np.linspace(19.0, 1350.0, len(velocity_pressure_weights[1]) + 1)[:-1]
]

fig, ax = plt.subplots()
ax.hist(velocity_pressure, bins=np.linspace(19.0, 1350.0, len(velocity_pressure_weights[0]) + 1), weights=velocity_pressure_weights, label=["Velocity", "Pressure"], alpha=0.7, edgecolor="black")
plt.show()

Explanation:
- The input contains raw data itself.
- Even if the two datasets have very different scales, as long as the original figure uses a single call with shared bins, the rewritten code must still share bins.
- Do not create separate bins for the two datasets just because that seems more reasonable.
- The new variable name still stays close to the original semantics.
- The final image remains unchanged.

====================
Additional ICL Example 8: Dictionary + loop over multiple series; do not rewrite into hard-coded if/elif branches
====================

Input code:
import matplotlib.pyplot as plt

data = {
    "A": [1.0, 1.2, 1.4, 1.6],
    "B": [2.0, 2.2, 2.4, 2.6],
    "C": [3.0, 3.2, 3.4, 3.6],
}

fig, ax = plt.subplots()
for label, values in data.items():
    ax.hist(values, bins=4, histtype="step", label=label)
plt.show()

Given target:
rewrite_mode = uniform_bins_weights
target_weights_literal = {
    "A": [1, 1, 1, 1],
    "B": [1, 1, 1, 1],
    "C": [1, 1, 1, 1],
}
target_ranges_literal = {
    "A": (1.0, 1.6),
    "B": (2.0, 2.6),
    "C": (3.0, 3.6),
}
target_hist_kwargs_literal = {"histtype": "step"}

Output code:
import matplotlib.pyplot as plt
import numpy as np

data_weights = {
    "A": [1, 1, 1, 1],
    "B": [1, 1, 1, 1],
    "C": [1, 1, 1, 1],
}
data_left = {"A": 1.0, "B": 2.0, "C": 3.0}
data_right = {"A": 1.6, "B": 2.6, "C": 3.6}

fig, ax = plt.subplots()
for label, values_weights in data_weights.items():
    values = np.linspace(data_left[label], data_right[label], len(values_weights) + 1)[:-1]
    ax.hist(values, bins=np.linspace(data_left[label], data_right[label], len(values_weights) + 1), weights=values_weights, histtype="step", label=label)
plt.show()

Explanation:
- The original code is dictionary + loop.
- The correct approach is to preserve the loop. Do not write if label == "A" / elif ...
- The raw sample dictionary data is no longer preserved because it is only used as raw hist input.
- The new variable name data_weights stays close to data.
- This is the key example for fixing hard-coded branches by key/body_part.

====================
Additional ICL Example 9: When hist input comes from slicing a 2D array, directly rewrite the sliced variable and remove the raw container
====================

Input code:
import matplotlib.pyplot as plt
import numpy as np

data = np.array([
    [10.0, 0.1],
    [12.0, 0.2],
    [14.0, 0.3],
    [16.0, 0.4],
    [18.0, 0.5],
])
x = data[:, 0]

fig, ax = plt.subplots()
ax.hist(x, bins=4, color="tab:blue")
plt.show()

Given target:
rewrite_mode = uniform_bins_weights
target_weights_literal = [1, 1, 1, 2]
target_x_literal_hint = np.linspace(10.0, 18.0, len(WEIGHTS_VAR) + 1)[:-1]
target_bins_literal_hint = np.linspace(10.0, 18.0, len(WEIGHTS_VAR) + 1)
target_hist_kwargs_literal = {"color": "tab:blue"}

Output code:
import matplotlib.pyplot as plt
import numpy as np

x_weights = [1, 1, 1, 2]
x = np.linspace(10.0, 18.0, len(x_weights) + 1)[:-1]

fig, ax = plt.subplots()
ax.hist(x, bins=np.linspace(10.0, 18.0, len(x_weights) + 1), weights=x_weights, color="tab:blue")
plt.show()

Explanation:
- The original 2D array data is only used to generate x.
- After rewriting, directly preserve x and do not preserve data.
- Do not write data_raw. Do not keep the original matrix in the code.
- This is the key example for fixing cases where raw samples are not fully removed.

====================
Additional ICL Example 10: When the same hist data is reused by the main plot and an inset, rewrite it once and reuse it
====================

Input code:
import matplotlib.pyplot as plt
import numpy as np

data = np.array([1, 1, 2, 2, 2, 3, 4, 4, 5, 6])
data_flat = data.flatten()

fig, ax = plt.subplots()
ax.hist(data_flat, bins=5, histtype="step", color="blue")

inset_axes = fig.add_axes([0.6, 0.6, 0.25, 0.25])
inset_axes.hist(data_flat, bins=5, histtype="step", color="black")
plt.show()

Given target:
rewrite_mode = uniform_bins_weights
target_weights_literal = [2, 3, 1, 2, 2]
target_x_literal_hint = np.linspace(1.0, 6.0, len(WEIGHTS_VAR) + 1)[:-1]
target_bins_literal_hint = np.linspace(1.0, 6.0, len(WEIGHTS_VAR) + 1)

Output code:
import matplotlib.pyplot as plt
import numpy as np

data_flat_weights = [2, 3, 1, 2, 2]
data_flat = np.linspace(1.0, 6.0, len(data_flat_weights) + 1)[:-1]

fig, ax = plt.subplots()
ax.hist(data_flat, bins=np.linspace(1.0, 6.0, len(data_flat_weights) + 1), weights=data_flat_weights, histtype="step", color="blue")

inset_axes = fig.add_axes([0.6, 0.6, 0.25, 0.25])
inset_axes.hist(data_flat, bins=np.linspace(1.0, 6.0, len(data_flat_weights) + 1), weights=data_flat_weights, histtype="step", color="black")
plt.show()

Explanation:
- The same data is reused by multiple hist calls.
- The correct approach is to create only one data_flat_weights and one rewritten data_flat.
- All later hist calls reuse them.
- Do not create separate variables for the main plot and the inset.
- Do not preserve the raw data.

====================
Additional ICL Example 11: For density + statistic line, directly write the visible statistic and do not preserve raw data
====================

Input code:
import matplotlib.pyplot as plt
import numpy as np

vals = [1.0, 1.4, 1.8, 2.2, 2.6, 3.0, 3.4, 3.8]
fig, ax = plt.subplots()
ax.hist(vals, bins=4, density=True, histtype="step", color="black")
ax.axvline(np.median(vals), color="red", linestyle="--")
plt.show()

Given target:
rewrite_mode = uniform_bins_weights
target_weights_literal = [0.125, 0.125, 0.125, 0.125]
target_x_literal_hint = np.linspace(1.0, 3.8, len(WEIGHTS_VAR) + 1)[:-1]
target_bins_literal_hint = np.linspace(1.0, 3.8, len(WEIGHTS_VAR) + 1)
target_visible_stat_literals = {"median_x": 2.4}
target_hist_kwargs_literal = {"histtype": "step", "color": "black"}

Output code:
import matplotlib.pyplot as plt
import numpy as np

vals_weights = [0.125, 0.125, 0.125, 0.125]
vals = np.linspace(1.0, 3.8, len(vals_weights) + 1)[:-1]

fig, ax = plt.subplots()
ax.hist(vals, bins=np.linspace(1.0, 3.8, len(vals_weights) + 1), weights=vals_weights, histtype="step", color="black")
ax.axvline(2.4, color="red", linestyle="--")
plt.show()

Explanation:
- Do not preserve the original vals just to preserve the median line.
- Rewrite density=True into directly writing the final visible density heights.
- If the target already gives the visible statistic for the statistic line, directly write it as a literal.
- This is the key example for fixing cases where raw samples are not fully removed: the hist input is changed, but other visible elements still secretly depend on raw data.

====================
Additional ICL Example 12: Construct a dictionary first, then flatten it into a single hist input
====================

Input code:
import matplotlib.pyplot as plt
import numpy as np

data = {
    "A": [0.1, 0.2, 0.3],
    "B": [0.2, 0.3, 0.4],
    "C": [0.3, 0.4, 0.5],
}
data_flat = np.array(list(data.values())).flatten()

fig, ax = plt.subplots()
ax.hist(data_flat, bins=4, color="steelblue")
plt.show()

Given target:
rewrite_mode = uniform_bins_weights
target_weights_literal = [1, 2, 3, 3]
target_x_literal_hint = np.linspace(0.1, 0.5, len(WEIGHTS_VAR) + 1)[:-1]
target_bins_literal_hint = np.linspace(0.1, 0.5, len(WEIGHTS_VAR) + 1)
target_hist_kwargs_literal = {"color": "steelblue"}

Output code:
import matplotlib.pyplot as plt
import numpy as np

data_flat_weights = [1, 2, 3, 3]
data_flat = np.linspace(0.1, 0.5, len(data_flat_weights) + 1)[:-1]

fig, ax = plt.subplots()
ax.hist(data_flat, bins=np.linspace(0.1, 0.5, len(data_flat_weights) + 1), weights=data_flat_weights, color="steelblue")
plt.show()

Explanation:
- The true input passed to hist is data_flat.
- Therefore, directly rewrite data_flat instead of going back to process each key separately.
- The raw dictionary data is only used to generate data_flat, so it is removed after rewriting.
- Do not write if key == ... / elif ...
- Do not keep both data and data_flat as two raw sample copies.

====================
Now process the real task
====================

{{RULES}}

File:
{{SOURCE_FILE}}

Precomputed targets for each hist call:
{{CALL_TARGETS}}

Original code:
{{PY_CODE}}
\end{quote}
\end{promptbox}

\end{document}